\documentclass[lettersize,journal]{IEEEtran}

\usepackage{amsmath,amssymb,amsfonts}

\usepackage{soul,color}
\usepackage{booktabs} 

\usepackage{graphicx}
\usepackage{textcomp}
\usepackage{float}

\usepackage{amssymb}
\usepackage{enumitem}
\usepackage{pifont}
\usepackage[linesnumbered,ruled,vlined]{algorithm2e}

\usepackage{pifont}

\usepackage[utf8]{inputenc}
\usepackage[T1]{fontenc}
\usepackage{textcomp}
\usepackage{multirow}

\usepackage{pbox}
 
\usepackage{makecell}
\usepackage{tabularray}

\SetCommentSty{mycommfont}
\usepackage{bbding}

\SetKwInput{KwInput}{Input}                
\SetKwInput{KwOutput}{Output}              
\SetKwFunction{NewFunction}{NewFunction}   

\usepackage{amsmath,amsfonts}

\usepackage{algorithmic}
\usepackage{array}

\usepackage[caption=false,font=normalsize,labelfont=sf,textfont=sf]{subfig}

\usepackage{textcomp}

\usepackage{verbatim}
\usepackage{graphicx}
\usepackage{cite}
\begin{document}

\title{IMOVNO+: A Regional Partitioning and
	Meta-Heuristic Ensemble Framework for Imbalanced
	Multi-Class Learning}

\author{
    \scriptsize
    \begin{minipage}{0.20\textwidth}
        \centering
        \IEEEauthorblockN{Soufiane Bacha $^{1}$} \\
        \IEEEauthorblockA{School of Computer and \\Communication Engineering,\\
        University of Science and Technology Beijing\\
        Beijing 100083, China. \\ 
     Key Laboratory of Xinjiang Coal Resources Green Mining, Ministry of Education\\ 
     Urumqi,830023,China\\ 
      Email: d202361018@xs.ustb.edu.cn}
    \end{minipage}
    \hfill
    \begin{minipage}{0.19\textwidth}
        \centering
        \IEEEauthorblockN{Laouni Djafri $^{2}$} \\
        \IEEEauthorblockA{Department of Mathematics,  Ibn Khaldoun University,Tiaret, 14000,Algeria\\
        LIM Laboratory of Informatics and Mathematics,  Ibn Khaldoun University,Tiaret,Algeria\\
        Email:laouni.djafri@univ-tiaret.dz}
    \end{minipage}
    \hfill
    \begin{minipage}{0.20\textwidth}
        \centering
        \IEEEauthorblockN{Sahraoui Dhelim $^{3}$} \\
        \IEEEauthorblockA{School of Computing, \\ Dublin City University, Ireland \\ 
        Email: sahraoui.dhelim@dcu.ie}
    \end{minipage}
    \hfill
    \begin{minipage}{0.20\textwidth}
        \centering
        \IEEEauthorblockN{Huansheng Ning $^{4}$} \\
        \IEEEauthorblockA{School of Computer and \\Communication Engineering,\\
        University of Science and Technology Beijing\\
        Beijing 100083, China. \\ 
        Email: ninghuansheng@ustb.edu.cn}
    \end{minipage}
    
}

\maketitle

\begin{abstract}
At the data level, class imbalance, overlap, and noise degrade data quality, reduce model reliability, and limit generalization. While these problems have been extensively studied in binary classification, multi-class problems remain largely underexplored. In binary scenarios, the minority–majority class relationship is well defined, which simplifies management. However, multi-class problems involve complex inter-class relationships, where conventional clustering often fails to capture distribution shape. Furthermore, approaches that rely solely on geometric distances risk removing informative samples and producing low-quality synthetic data. Similarly, traditional binarization addresses imbalance locally, ignoring global inter-class dependencies. At the algorithmic level, ensemble methods encounter challenges of integrating weak classifiers, often resulting in limited robustness and suboptimal predictive performance. To address these limitations, this paper proposes IMOVNO+ (IMbalance–OVerlap–NOise+ Algorithm-Level Optimization), a two-level framework designed to jointly enhance data quality and algorithmic robustness for both binary and multi-class classification tasks. At the data level, first, conditional probability is used to quantify the informativeness of each sample. Second, the dataset is partitioned into core, overlapping, and noisy regions, enabling region-specific processing and reducing computational search across the entire dataset. Third, an overlapping-cleaning algorithm is introduced that combines Z-score metrics with a big-jump gap distance. Fourth, a smart oversampling algorithm based on multi-regularization controls synthetic sample proximity, preventing new overlaps. At the algorithmic level, a meta-heuristic prunes the ensemble classifiers, mitigating weak-learner influence and improving performance. The proposed framework was evaluated on 35 datasets, including 13 multi-class and 22 binary datasets. Experimental findings demonstrate that the proposed framework outperforms state-of-the-art methods, approaching 100\% in several cases. For multi-class datasets, IMOVNO+ achieves gains of 37.06–57.27\% in G-mean, 25.13–43.93\% in F1-score, 24.58–39.10\% in precision, and 25.62–42.54\% in recall. In binary classification, the framework consistently attains near-perfect performance across all evaluation metrics, with improvements ranging from 13.77\% to 39.22\%. Practically, the proposed framework provides a robust solution for handling data scarcity and imbalance arising from data collection difficulties and privacy constraints.
\end{abstract}


\section{Introduction}

Data is regarded as the most important component of artificial models, allowing intelligent algorithms to extract hidden patterns, detect correlations, and build predictive assumptions. The quality of the supplied data is even more important for building robust, reliable, and realistic models. However, in real-world conditions, such as collection difficulties, privacy constraints, input errors, and other anomalies, data quality is often compromised, negatively affecting the learning process. Imbalanced class distributions, class overlap, and noise are intrinsic characteristics of real-world data that degrade data quality, reduce model performance, lower reliability, and lead to poor generalization.  Class imbalance is defined as a situation in which the observations in one class (the majority class) significantly outnumber those in other classes (the minority classes) \cite{Introdcution2}. Under the condition of class imbalance, the minority class participates less in the final prediction, which leads the model to overfit toward the majority class, causing it to predict all instances as the majority class and achieve a higher accuracy value. However, it is meaningless to predict healthy patients accurately while ignoring unhealthy patients. In practical applications, the data used is often imbalanced.  For example, in software defect detection, datasets naturally exhibit class imbalance between defective and non-defective classes  \cite{Introdcution1}. In network intrusion detection, normal network traffic data is much larger than intrusion traffic  \cite{Introdcution3}; and in the context of medical data, diagnostic datasets are often imbalanced \cite{Introdcution4}, etc. Studies \cite{Introdcution5}\cite{Introdcution6} have pointed out that class overlap, where the classes intersect and increase the uncertainty in distinguishing between them, can be similarly or even more challenging than class imbalance, negatively impacting the performance of learners. Meanwhile, noise is regarded as another major challenge that affects classification performance. Studies \cite{Introdcution5}\cite{Introdcution6} have shown that noisy data is considered one of the most damaging factors to classification accuracy, followed by class overlap and class imbalance.

For binary classification, the class distribution is straightforward, and the inter-class decision boundary between the majority and minority classes is easier to identify. Over the past decade, various binary class imbalance approaches have been developed. Broadly, these approaches can be divided into three families: data-level, ensemble-based, and algorithm-level approaches \cite{Introdcution7}. In recent years, extensive research has focused on developing numerous solutions for handling binary imbalanced datasets. However, few of them pay attention to multi-class imbalance challenges. Nevertheless, multi-class imbalance classification is common in many fields, such as medical disease diagnosis \cite{Introdcution8}, fault diagnosis \cite{Introdcution9}, intrusion detection systems \cite{Introdcution10}, etc. In a multi-class scenario, the relationships between classes are more complex. For instance, in multi-class imbalance, the dataset may contain a majority class and many minority classes, a minority class and many majority classes, or multiple minority and majority classes \cite{Introdcution11}.  In addition, overlapping and borderline samples can appear across more than two decision boundary regions \cite{Introdcution2}, which increases class intersection and the difficulty of distinguishing between classes. Unfortunately, due to this complexity, the approaches used for binary class imbalances are invalid for applying directly to multi-class imbalanced data.

The current most popular methods for addressing multi-class imbalance are binary solutions, which can be grouped into two families: binarization strategies and ad hoc approaches. The binarization approach decomposes a multi-class problem into subproblems of two classes, employing one-versus-one (OVO) or one-versus-all (OVA) techniques \cite{Introdcution5}. However, treating a multi-class problem as a set of several binary classification problems causes the loss of inter-class relationship information. Additionally, this approach transfers a multi-class problem to binary subproblems without addressing issues such as class overlap and noisy data. For instance, the OVO method MC-CCR, proposed in \cite{Introdcution15}, uses an energy-based approach to model the appropriate region for oversampling, followed by a cleaning process. However, pushing the majority samples outside leads to the creation of overlapping regions and crowded areas with other classes, resulting in a complex data distribution. Ad hoc is an alternative solution to binarization, which aims to design methods that directly handle the original multi-class imbalanced problem. This approach has shown promising results, as it can maintain class associations and improve predictive performance.  Earlier, a data-level method of binary classification had been extended to address a multi-class problem based on SMOTE oversampling methods. For instance, studies \cite{Introdcution13}\cite{Introdcution14} used oversampling methods to address multi-class imbalance and noisy data. Although they showed promising results, they generated new samples in the vicinity of the current minority class, followed by cleaning, which resulted in unclear decision boundaries. Other improvements to oversampling multi-class, proposed in \cite{Introdcution16}, use the AMDO method to capture the structure of the minority class and generate data to balance classes. However, as the author states, this method overlooks the integration of ensemble learning techniques, which decreases classifier diversity. Nevertheless, the development of a comprehensive approach that simultaneously tackles class imbalance, class overlap, and data noise remains an open research problem. Another work \cite{Introdcution17} proposed the MDOBoost method for multi-class imbalance, which combines MDO with the AdaBoost.M2 ensemble. However, it does not address class overlap or noisy data. Recently, studies \cite{Introdcution18} proposed the E-EVRS method based on ensemble learning. Meanwhile,\cite{Introdcution19} proposed the SAMME.C2 method, which integrates SAMME AdaBoost with Ada.C2 cost-sensitive learning. However, despite their effectiveness, these methods do not address overlapping or noisy data. Additionally, there is no control over the placement of synthetic data, nor any metaheuristic for pruning the ensemble of classifiers.

In summary, the major shortcomings of the aforementioned multi-class imbalanced approaches are: (1) Most previous studies on multi-class imbalance, which employ geometrical computations to identify noisy data or reduce imbalance and overlap, overlook any assessment of the information content within each sample (e.g., probability, entropy). (2) There is a lack of regularization penalty over the location of new synthetic data to avoid creating new overlapping issues. (2) Clustering or distance-based membership methods often label the farthest samples (isolated points) as noisy, overlooking the evaluation of their contribution, which results in the loss of important samples. (3) The use of the Jaya algorithm for ensemble pruning, to reduce the size of the ensemble while maintaining performance, remains largely unexplored. (4) None of the previous studies has applied the Big Jump concept together with a Z-score metric to reduce class overlap. Instead, they rely on radius- or sphere-based criteria, which are highly sensitive to the choice of the radius parameter. These drawbacks motivate the development of a unified framework to address the popular intrinsic characteristics of data, including class imbalance, class overlapping, and noisy data for both binary and multi-class contexts.

To tackle the aforementioned limitations, this paper aims to develop a novel unified framework called IMOVNO+. The objectives of the proposed framework are to effectively address the combined challenges of class imbalance and overlapping data, reduce the loss of informative and high-contributing samples, minimize noisy data, improve the quality of synthetic samples, and enhance the clarity of inter-class decision boundaries in both multi-class and binary classification. Additionally, pruning the ensemble of weak learners. The proposed framework employs conditional probability to quantify the informativeness and learning contribution of each sample, rather than relying solely on numerical size differences used in prior studies, to balance the data and identify overlapping and noisy samples. The conditional probability divides the dataset into three main regions: the core region, the overlapping region, and the noisy region. Core-region samples are those with high certainty of belonging to their class. A low probability indicates samples with low informativeness, which are discarded, while overlapping samples are those whose probability of belonging to another class is higher than that of their class. Secondly, the overlapping-cleaning process is applied using the Z-score and the big-jump distance, which addresses the limitations of previous studies regarding the difficulty of determining appropriate sphere and radius parameters. The core-region samples and the non-overlapping samples returned from the cleaning step are then used to generate new synthetic data for balancing the dataset. Thirdly, the SMOTE-based oversampling process is applied to generate high-quality synthetic samples composed of highly confident instances. To further avoid creating new overlapping samples, a multi-regularization penalty is introduced to ensure that the synthetic samples are generated near or within each minority class and remain distant from majority-class samples, thereby forming a compact and well-separated minority-class distribution. Fourth, the balanced data obtained from these steps is assigned to an ensemble learning process to benefit from the strengths of multiple classifier decisions. Additionally, an ensemble-pruning strategy is proposed using a modified Jaya algorithm to reduce the ensemble size while maintaining overall performance. The experimental findings demonstrate the effectiveness of the proposed framework, showing significant improvements. For multi-class datasets, IMOVNO+ achieves gains of 37.06–57.27\% in G-mean, 25.13–43.93\% in F1-score, 24.58–39.10\% in precision, and 25.62–42.54\% in recall. In binary classification, the framework consistently attains near-perfect performance across all evaluation metrics, with improvements ranging from 13.77\% to 39.22\%.

The main contributions of the proposed framework can be summarized as follows:

\begin{itemize}

\item The proposed framework, IMOVNO+, directly addresses class imbalance, class overlap, and noisy data problems in both multi-class and binary classification contexts, rather than decomposing the problem into binary subproblems, thereby preserving inter-class relationships.

\item 	IMOVNO+ addresses the aforementioned problems based on the information contribution of samples and degree of belonging rather than numerical distance, minimizing the loss of highly contributing data.

\item 	The framework partitions the dataset into core, overlapping, and noisy regions, enabling separate treatment of each and reducing overall complexity according to the difficulty level of each region.

\item The discrete Jaya method has been adopted for ensemble pruning, reducing the ensemble size, selecting high-contributing classifiers, and preserving overall performance.

\item Extensive experimental testing was conducted on various imbalanced datasets in both multi-class and binary classification contexts. 

\end{itemize}

The rest of this paper is organized as follows. Section 2 summarizes existing methods addressing binary and multi-class imbalanced data, as well as class overlap and noisy data. Section 3 presents the proposed framework in detail for handling data quality problems. Section 4 presents the experimental setup. Section 5 reports comprehensive experimental evaluations, followed by a thorough discussion and analysis of the results. Finally, the conclusion and future scope are presented in Section 6.

\section{Related work}
In this section, we discuss the challenges of real-world data, including imbalanced classes, class overlap, and noise, as well as existing approaches to address them in the contexts of binary and multi-class classification, binarization studies, and meta-heuristic strategies, where these challenges remain insufficiently explored in multi-class tasks. 

\subsection{Challenges of class imbalance and class overlap in binary classification}

Numerous prior studies have proposed solutions for imbalanced binary problems, broadly categorized into data-level, algorithm-level, and ensemble-based approaches. Data-level methods rebalance training data through oversampling, undersampling, or a combination of both. While few studies consider the contents of samples during resampling using active learning \cite{RelatedWork1}, most existing approaches remove data based on numerical differences across classes, risking the loss of informative data. On the other hand, the Synthetic Minority Oversampling Technique (SMOTE) is a popular oversampling approach \cite{RelatedWork2}. Although effective, basic SMOTE assumes that the minority class is homogeneous and blindly interpolates samples. However, real-world data are often overlapping and crowded, which can increase class overlap  \cite{RelatedWork3} and generate low-quality, noisy data.  The literature has proposed numerous enhancements to basic SMOTE, addressing three key aspects. First, this work proposes a modified SMOTE solution that generates new samples far from the decision boundary \cite{RelatedWork4} \cite{RelatedWork5}. Although this approach preserves a clear boundary and reduces overlap, generating synthetic samples distant from the boundary leads to limited diversity of minority samples in these regions and overlooks important information required to modify the decision boundary  \cite{RelatedWork6}. Conversely, the second aspect focuses on generating samples near the decision boundary to capture its characteristics \cite{RelatedWork7}; however, it may produce overlapping samples due to the lack of a regularization penalty for controlling synthetic data. The third aspect is based on clustering minority classes, followed by generating synthetic data within each cluster \cite{RelatedWork8}\cite{RelatedWork9}.

Algorithm-level approaches are the second most common strategy for class imbalance, requiring a profound understanding of the model to adapt to imbalanced data \cite{RelatedWork10}. Two popular classifiers have integrated an algorithm-level perspective into their structure. According to earlier studies, the decision tree classifier has replaced the conventional splitting criteria, entropy, and Gini, with the Hellinger Distance Decision Tree (HDDT). Similarly, instance weighting has been applied to adapt the support vector machine (SVM) classifier. These approaches can be categorized into two groups: one-class learning (OCC) and cost-sensitive learning (CSL). OCC focuses on one class as the target class while treating others as noisy data \cite{RelatedWork11}. CSL balances the data by assigning different cost values to misclassified samples based on their importance using a cost matrix.  Zhang\cite{RelatedWork12} developed two CSL methods named Distance-CS-KNN and Direct-CS-KNN to handle class imbalance. Other research has also integrated CSL logic with boosting approaches, such as AdaCost, CSB1, CSB2, RareBoost, and AdaC1 \cite{RelatedWork13}. However, this approach faces the challenge of determining appropriate cost values.

Ensemble-based learning is a third alternative solution for imbalanced classes, combining one of the previous approaches with popular ensemble strategies such as bagging or boosting. SMOTEBoost is considered an older ensemble-based learning technique that integrates SMOTE with boosting techniques developed by Chawla et al. \cite{RelatedWork14}. Although efficient, this method can result in additional computational costs and may blindly generate data that overlaps and contains noise. Recently, various modified versions have been proposed. Pangastuti et al. \cite{RelatedWork15} applied data mining to the Bidikmisi scholarship dataset using a random forest in combination with boosting and bagging strategies. The proposed framework achieved an accuracy score of over 90\%. Li et al. \cite{RelatedWork16} proposed a Center Jumping Boosting Machine (CJBM) method, which clusters the minority class and then oversamples at the cluster centers.

\subsection{Challenges of class imbalance and class overlap in multiclass classification}

Classification with imbalanced data presents more challenges than the binary scenario due to complex associations across classes \cite{RelatedWork17} . The separation between classes is more crowded, especially in the presence of overlapping issues. In multi-class classification scenarios, one class can play the role of a majority class for some classes and a minority class for others, resulting in a multi-imbalance situation. This phenomenon differs from binary classification, where imbalance is restricted to two classes. Additionally, a multiclass scenario can result in simultaneous overlapping, noisy data, and outlier issues, making the learning task complex for any classifier \cite{Introdcution2}. Two popular taxonomies have been proposed in the literature to address multiclass classification tasks: (1) binarization strategy and (2) ad hoc approaches.

\subsubsection{Binarization strategy}

This solution is a popular approach to handling multiclass imbalanced data, which aims to transform a multiclass problem into a conventional binary problem. Two popular techniques proposed in this research direction include one-versus-one (OVO), formulated as $C(C-1)/2$ ($C$ representing the dataset classes), and one-versus-all (OVA) \cite{Introdcution5}. For instance, Noise-robust Oversampling Algorithm (NROMM) \cite{Introdcution14} was proposed to remove noisy data within clusters for multiclass imbalanced datasets. It integrates both the original and newly generated data from each class using a one-versus-ensemble (OVE) strategy, which resembles the traditional OVO decomposition. NROMM employs advanced clustering techniques to capture the structural distribution of both majority and minority classes, followed by an adaptive embedding algorithm to generate safe samples and clearer class boundaries, thereby increasing the separation between classes. Experimental results demonstrated that NROMM outperforms comparison methods by approximately 10\% across various benchmark datasets.

The authors in \cite{Introdcution13} introduced a novel oversampling method for multiclass classification called Multiclass Radial-Based Oversampling (MC-RBO). Unlike SMOTE-based algorithms, MC-RBO utilizes the nearest neighbors of minority samples while incorporating data from all classes. The method achieved promising results with statistical significance. However, this method generates new instances only in the vicinity of the current minority class and relies on a subsequent cleaning process, which often leads to unclear decision boundaries between majority and minority classes \cite{Introdcution2}. Another OVO approach present in another work \cite{Introdcution15} proposed the MC-CCR method, which uses an energy-based approach to model the appropriate region for oversampling, followed by a cleaning process. The process starts by rounding a sphere around the minority samples and gradually expanding it until it covers some majority samples. These majority samples are then pushed outside the sphere. However, pushing the majority of samples outside leads to the creation of overlapping regions and crowded areas with other classes, resulting in complex data distribution.  Moreover, most of these methods, which use geometrical computations for identifying noisy data or reducing imbalance and overlap, overlook any quantification process of the information content within each sample. For instance, considering isolated samples as noisy data due to their distance from others and removing them may lead to the loss of informative instances, which is crucial in sensitive domains like medical scenarios.

In contrast, our method directly evaluates the informativeness and contribution of each sample, ensuring the preservation of high-quality samples. Noisy data, on the other hand, are those with a low degree of class membership, regardless of whether they are isolated or surrounded by other classes.  Broadly speaking, OVO is a temporary solution for handling multiclass problems; however, it cannot be fully relied upon. OVO treats a multiclass task as a set of binary subproblems, preserving the relationship between only two classes at a time while losing the remaining information from the other classes.

\subsubsection{Ad hoc  approaches}

This is an alternative solution to the binarization strategy for multiclass imbalanced data, which aims to design methods to directly handle the original multiclass scenario without subcategorizing into paired classes. This approach showed promising results, as it could both preserve the relationship between classes and achieve higher performance. Various ad hoc methods have been developed in the literature, offering both advantages and drawbacks. Earlier, a data-level approach for binary classification was extended to tackle multi-class problems based on basic SMOTE methods.  \cite{Introdcution11} have divided the multi-class dataset into four key regions: safe, borderline, rare, and outliers, to obtain the characteristics of each class, followed by a synthetic procedure according to the importance of each region. This approach provides a deeper understanding of the nature of the multi-class imbalanced problem.  Alternative data-level approaches use Mahalanobis-distance techniques to generate new data. For instance, reference \cite{Introdcution16} proposed Adaptive Mahalanobis Distance-based Over-sampling (AMDO). The purpose of this strategy is to capture the covariance structure of the minority class and generate samples along the contours learned by the algorithm. However, as the authors stated, this method overlooks the integration of ensemble techniques, which reduces classifier diversity. It is also notable that there is no noise-filtering mechanism. Another work \cite{Introdcution17}  introduced a Mahalanobis distance-based oversampling technique combined with boosting (MDOBoost), which integrates MDO for oversampling and the AdaBoost.M2 ensemble. Although it is effective due to the benefits of sampling and reweighting, it does not address class overlap or noisy data. It is worth noting that both methods, MDO and MDOBoost, treat samples numerically (distance-based) rather than quantifying their contribution or informativeness during learning. Algorithm-level solutions have also been extended to multi-class imbalanced data scenarios. For instance, Hoens et al. \cite{RelatedWork26} proposed multi-class HDDTs (MC-HDDTs) to address class imbalance. This method applies the Hellinger distance as the splitting criterion for decision trees, evaluating all possible choices of positive and negative classes to determine the best split. However, Hellinger distance performs poorly in multi-class scenarios and often places splits inaccurately. Finally, cost-sensitive and ensemble methods have also been adopted for multi-class classification tasks. For instance, the work in \cite{RelatedWork27} proposed the E-EVRS method employing belief function theory and ensemble learning to rebalance multi-class imbalanced data. It selects ambiguous majority class data for undersampling, followed by oversampling minority class samples in the borderline region. Finally, these procedures are integrated into ensemble learning. The SAMME.C2 method \cite{RelatedWork28} proposed, combining SAMME AdaBoost and Ada.C2 cost-sensitive. Despite its effectiveness, it does not address overlapping and noisy data issues, and there is no control for new synthetic data (absence of a regularization penalty), which leads to more overlapping and crowded samples. Moreover, it is worth noting that this method does not apply any heuristic algorithm for classifier selection or parameter optimization. Additionally, defining a suitable cost value is considered a challenging task.

\subsection{Meta-heuristic for ensemble learning pruning}
The ensemble approach is a machine learning technique that combines several weak base classifiers and aggregates their outputs (e.g., through averaging) to produce a final prediction. This strategy improves both prediction accuracy and generalization performance. However, a major challenge is that including too many classifiers can be ineffective, especially when some of them contribute little or nothing to the final decision. Ensemble pruning (selective ensemble) addresses this issue by selecting only a subset of classifiers that can match or even outperform the full ensemble. A well-known example supporting this idea is presented in \cite{RelatedWork29}, which shows that achieving strong performance does not require hundreds of classifiers. The main objective, therefore, is to identify the best combination of classifiers that reduces ensemble size while maintaining, or even improving, generalization performance compared with using all classifiers. Three popular pruning approaches have been proposed in the literature: clustering-based, ordering-based, and optimization-based techniques. Since our research focuses on metaheuristic methods, we review some examples of applying metaheuristics to ensemble pruning.

Metaheuristics have shown promising results in searching for an optimal combination of classifiers \cite{RelatedWork30}. Generally, metaheuristic-based ensemble pruning methods can be categorized into two main techniques: trajectory-based and population-based techniques. According to [31], a trajectory-based method starts with an initial solution and iteratively replaces it with a new one by applying the modification operators of the algorithm, improving the current solution until it reaches a local optimum (where no further improvement is possible). In contrast, population-based metaheuristics operate on a group of search agents (potential solutions) rather than a single solution. Each iteration replaces some existing solutions with new ones, selecting the best-performing candidates. A population-based technique typically involves a set of algorithms inspired by natural phenomena. These techniques mainly include evolutionary computation (e.g., genetic algorithm, differential evolution, and gene expression programming), swarm intelligence (e.g., colony optimization, firefly, and particle swarm optimization), biology-based, chemistry-based, nature-inspired, and physics-based algorithms \cite{RelatedWork31}\cite{RelatedWork32}.

Previous studies \cite{RelatedWork33} employed human-like foresight and hill-climbing optimization to select a subset of base classifiers across ten datasets, showing that pruning a heterogeneous ensemble improves accuracy. In \cite{RelatedWork34}, a genetic algorithm was applied to prune an ensemble of classifiers in a multi-class imbalanced scenario. More recently, a study \cite{RelatedWork35} proposed a novel self-adaptive stacking ensemble model (SSEM) that prunes a set of base classifiers using a genetic algorithm. The results demonstrated that the optimal subset outperformed eleven state-of-the-art classifiers. Another work \cite{RelatedWork36}  proposed EvoBagging, which iteratively enhances the diversity of classifiers in the bagging ensemble. The results showed that this method outperformed conventional bagging and random forest approaches in both binary and multiclass settings, under balanced and imbalanced scenarios. More recent works include a hybrid dynamic ensemble pruning framework for time series prediction \cite{RelatedWork37}, which applies Genetic Algorithm (GA), Particle Swarm Optimization (PSO), and Artificial Fish Swarm Algorithm (AFSA) for ensemble optimization. Another study \cite{RelatedWork38} used a perturbation-based binary Salp Swarm Algorithm to identify the optimal subset of Extreme Learning Machines (ELMs). The Jaya algorithm is a recent and efficient metaheuristic. Compared to previous methods, it is parameterless and faster at obtaining solutions. Although effective, it is worth noting that the Jaya algorithm has been widely used for feature pruning \cite{RelatedWork39} and hyperparameter optimization \cite{Methodology8}, as reported in several studies. However, its application to ensemble pruning, especially in multi-class imbalanced scenarios, remains largely unexplored, which we address in this paper.

\section{The Proposed Framework}

\begin{figure*}[!t]
\centering
\includegraphics[width=18cm,height=9cm]{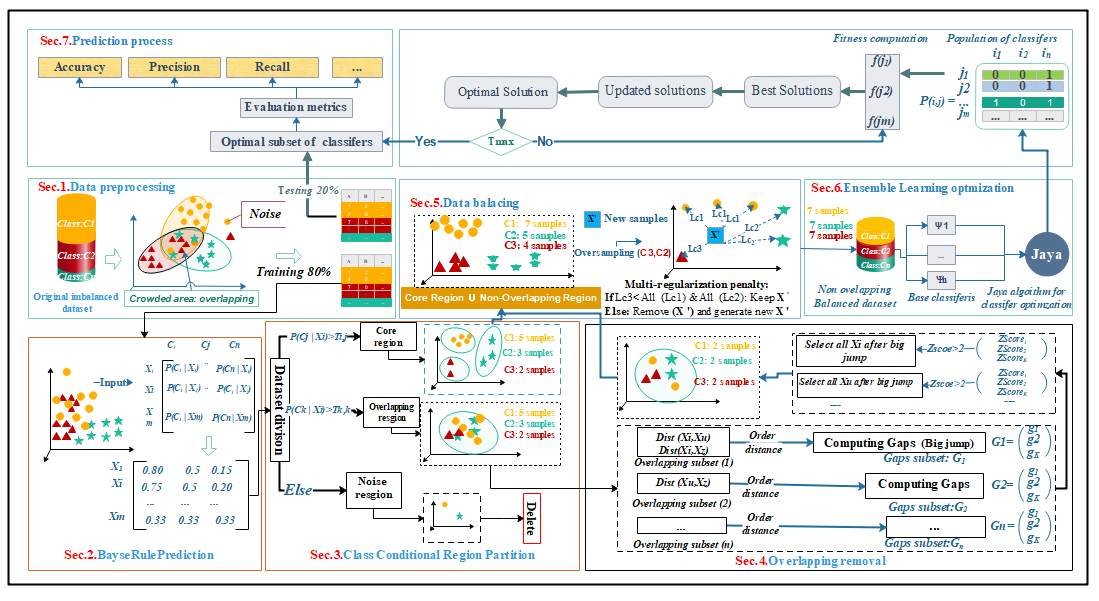}
\caption{Flowchart of the proposed framework}
\label{fig:Proposedframework}
\end{figure*}

The full architecture of the proposed framework is illustrated in Fig.~\ref{fig:Proposedframework}. The proposed framework is designed to handle the main data-quality challenges, including imbalanced data, overlapping samples, and noisy data in both multi-class and binary classification. The proposed solution consists of a series of coordinated and controlled phases that incorporate probabilistic computation, meta-heuristic techniques, and machine learning algorithms. These successive phases mainly consist of (1) Initial imbalanced raw data, (2) Region partition: the dataset is divided into three regions, including a core region, an overlapping region, and noisy data, to facilitate and reduce the processing of each region separately using conditional probability as the membership degree for each data instance. (3) Overlap cleaning: aiming to clean crowded areas between classes and maximize inter-class separation to optimize learning. (4) Data balancing: aiming to handle the imbalanced data challenge by oversampling the classes, (5) Oversampling multi-regularization penalty, which applies modified SMOTE with the proposed multi-regularization penalty to oversample classes and control the position of new samples to prevent creating new overlapping issues between classes, (6) Finally, optimized ensemble methods are applied, which use the balanced and cleaned overlapping data from former steps, followed by a pool of weak classifiers optimized by a meta-heuristic algorithm to select high-contributing classifiers and neglect those low-performing ones. These phases, along with their corresponding algorithms, are described in detail in the following subsections.

\subsection{Class-conditional region partition and noise removal}

Unlike the binary classification scenarios, handling imbalanced, overlapping, and noisy data is a more challenging task in multi-class scenarios due to the complex relationships between classes \cite{RelatedWork17}. To address these challenges in both multi-class and binary classification scenarios, we propose a new ad hoc framework, IMOVNO+. The proposed solution consists of a series of successive algorithms applied to the initial data containing the aforementioned anomalies, producing cleaned, structured, and high-quality data. The first phase of this framework is mainly based on region partitioning and noise cleaning using conditional probability measurements.
In this phase, conditional probability is employed as a measure to divide the initial input data into three main regions: the core region, the overlapping region, and the noisy region. This decision aims to process each region separately and reduce processing time and searching across the entire dataset. Unlike the existing solutions for dividing datasets \cite{Introdcution5}, which employ distance-based membership degrees, the existing approaches mainly focus on numerical computation and overlook any measurement of the information contribution of the samples. Additionally, membership distance relies on determining the data center; however, this becomes more difficult when the shape of the distribution is not uniform. Moreover, an inappropriate choice of the fuzzy degree value in the FCM algorithm can lead to inappropriate clustering. For this reason, we propose a conditional probability to divide the dataset. Our framework aims to quantify the degree of contribution of each sample in order to identify core regions, overlapping areas, and noise, which provides an advantage over distance-based membership and other probabilistic methods. Conditional probability does not require training a model to obtain the probability of sample membership, thus avoiding the issue of training on imbalanced data, where models built on such data are weak at detecting positive cases  \cite{Methodology13}. For distance-based membership methods, noisy samples are typically identified by their low membership values, which are often associated with being the farthest samples from their classes. However, this approach overlooks their information content, meaning that isolated samples with high informative value may be incorrectly removed. In contrast, the proposed framework detects noisy samples based on low class-membership contribution, enabling noise removal across the entire dataset rather than only at the farthest points, thereby preserving informative samples. The conditional probability is described in Algorithm~\ref{alg:SMP} with the following main steps.

Conditional probability refers to the likelihood of an event occurring based on the occurrence of a previous event (or outcome). Bayes’ rule forms the core of conditional probability. Let $X = \{ x_1, x_2, \ldots, x_z \}$ represent a set of attribute vectors, and $C = \{ c_1, c_2, \ldots, c_n \}$
represent the $n$ classes of $C$.  Bayes’ rule can be defined in  Eq.~(\ref{eq:probability1}).

\begin{equation}
P(C_i/X) = \frac{p(C_i/ X )\, P(C_i)}{\sum_{i=1}^{n} P(X / C_i)\, P(C_i)}
\label{eq:probability1}
\end{equation}

Let $X_i$ be the i-th attribute, which can take various values $x_d$. The Naive Bayes classifier assumes that the attributes are independent of each other, so the likelihood of an instance $X_i$ given a class $C_j$  can be defined as in Eq.~(\ref{eq:probability2}).

\begin{equation}
P(C_i / X)= \prod_{d=1}^{z} p(X = x_d /C_j)
\label{eq:probability2}
\end{equation}

In ordinary classification tasks, Eq.~(\ref{eq:probability1})  may be insufficient to estimate the most probable class given the input data. Therefore, the posterior probabilities in  Eq.~(\ref{eq:probability3}) can be obtained based on the prior probability  $ P(C_i) $ of each class and the conditional probability of the data from Eq.~(\ref{eq:probability1}) and (\ref{eq:probability2}).

\begin{equation}
\frac{
    P(C_i) \prod_{d=1}^{z} p(X = x_d \mid C_j)
}{
    \sum_{j=1}^{n} P(C_j) \prod_{d=1}^{z} p(X = x_d \mid C_j)
}
\label{eq:probability3}
\end{equation}

In the next step, the probability results of each class returned from Algorithm~\ref{alg:SMP} are applied to divide the dataset into three regions: core, overlapping, and noisy, based on the degree of membership of samples to their respective classes, as outlined in Algorithm~\ref{alg:CCRP}. To identify these different regions, a moderated ratio (threshold) of each class distribution in the data must to be established. To accomplish this, we define $T_{(i,j)}$ as the threshold for the current class $ j$, while $T_{(k,k)}$ represents the threshold of the other classes excluding the current class $(k \neq j)$, as expressed by the following equations.

\begin{equation}
T_{i,j} = \overline{\overline{k_j} + \mathrm{MCP}_j}
\end{equation}

\begin{equation}
T_{k,k} = \overline{\overline{k_k} + \mathrm{MCP}_k}
\end{equation}
$T_{i,j}$: Minimum threshold for the class $j$.\\
$T_{k,k}$: Minimum threshold for the other classes $k$ ($k \neq j$).\\
$\overline{k_j}$: Average conditional probability of belonging to the class $j$.\\
$\overline{k_k}$: Average conditional probability of belonging to the class $k$.\\
$\mathrm{MCP}_j$: Maximum conditional probability of the class $j$ in the real samples.\\
$\mathrm{MCP}_k$: Maximum conditional probability of the class $k$ in the real samples.\\

According to the threshold value, the data can be grouped into three regions. Samples with probabilities higher than the threshold of their original class are more likely to belong to that class, which defines the core region. Samples with probabilities lower than their class's threshold but higher than other classes' thresholds are more likely to belong to multiple classes. This condition is a difficult situation where the classes intersect and overlap, which makes it challenging for the classifier to distinguish between them. This situation defines the overlapping region.  Finally, samples whose probabilities are lower than the threshold of both their original class and the thresholds of other classes form noisy data, where the degree of belonging to any class is too low due to the low information content of those samples.  Unlike distance-based membership methods, this definition is not limited to samples closest to their original class; rather, it extends to any sample in the dataset, regardless of its location.

In multi-class imbalanced datasets, important minority instances tend to have fewer samples. If classes are not clearly distinguished, minority instances are likely to be incorrectly labeled. Suppose there is a dataset S, with  $m$ samples $X = \{ X_1, X_2, \ldots, X_m \}$ and $n$ classes $C = \{ c_1, c_j, c_n \}$  in the dataset, where the number of instances in each class is $|c_1|, |c_j|, \ldots, |c_n|$, and the number of instances follows the order $|c_1| > |c_j| > |c_n|$. Mathematically, the previous regions can be modeled through the following definition.

\textbf{Definition 1 (Core region)}
Sample $X_i$ belongs to class $j$ as its true label. When the probability of membership of $X_i$ is greater than the average probability of all class j samples belonging to class j, it is denoted as $T_{(i,j)}$.  It indicates a higher certainty that this sample belongs to the core region of its class. Mathematically, it is defined as in Eq.~(\ref{eq:Coreregion}).
\begin{equation}
\text{Core\_region} = \{ X_i \in S \mid P(X_i) > T_{i,j} \} \label{eq:Coreregion}
\end{equation}

\textbf{Definition 2 (Overlapping  region)}
The sample $X_i$ belongs to class $j$ as its true label, and $k$ indicates other classes except class $j(k \neq j) $. When the probability of membership of $X_i$ is greater than the average probability of all other class $k$ samples belonging to class $k$, it is denoted as $T_{(k_i,k)}$. It indicates a higher certainty that this sample belongs to the overlapping region. Mathematically, it is defined as in Eq.~(\ref{eq:Overlappingregion}).
\begin{equation}
\text{Overlapping\_region} = \{ X_i \in S \mid P(X_i) > T_{k,k} \} \label{eq:Overlappingregion}
\end{equation}

\textbf{Definition 3 (Noisy region)}
Sample $X_i$ belongs to class $j$ as its true label, and $k$ indicates other classes except class $j(k \neq j)$. When the probability of membership of $X_i$ is strictly lower than both the average probability of all class $j$ samples belonging to class $j$, and the average probability of all other class $k$ samples belonging to class $k$. It indicates a higher certainty that these samples are noisy, meaning their information content and contribution to learning are low, and they are automatically removed from the data in this step. Mathematically, it is defined as in Eq.~(\ref{eq:Noisyregion}).
\begin{equation}
\text{Noisy\_region} = \{ X_i \in S \mid P(X_i) < T_{i,j} \ \&\ T_{k,k} \} \label{eq:Noisyregion}
\end{equation}

\begin{algorithm}[t]
\small 
\SetAlgoVlined
\SetAlgoNlRelativeSize{-1}

\KwInput{
\quad - Attribute vector: $X = (x_1, \dots, x_z)$\\
\quad - Set of class labels: $C = \{C_1, \dots, C_n\}$\\
\quad - Prior probabilities: $P(C_j)$ for all $C_j \in C$\\
\quad - Likelihoods: $P(X_i \mid C_j)$ 
}

\KwOutput{
\quad - $P(C_j \mid X_i), j = 1, \dots, n, i = 1, \dots, m$\\
\quad PosteriorProbability[$i$][$j$] = 
\scalebox{0.6}{$
\begin{bmatrix}
P(C_1 \mid X_1) & \dots & P(C_n \mid X_1) \\
\vdots & \ddots & \vdots \\
P(C_1 \mid X_m) & \dots & P(C_n \mid X_m)
\end{bmatrix}
$}
}

\SetKwFunction{FMain}{SMP}
\SetKwProg{Fn}{Function}{:}{}
\Fn{\FMain{$X_i$}}{

    \ForEach{$C_j \in \{C_1, \dots, C_n\}$}{
        Compute prior: $P(C_j)$\;
        Compute likelihood: $P(X_i \mid C_j) = \prod_{d=1}^{z} P(X_i=x_d\mid C_j)$\;
        Compute unnormalized posterior: $\text{Numerator}_i = P(C_j) \cdot \prod_{d=1}^{z} P(X_i = x_d \mid C_j)$\;
    }

    Compute normalization constant:\\
    $\text{Denominator} = \sum_{j=1}^{n} P(C_j) \cdot \prod_{d=1}^{z} P(X_i = x_d \mid C_j)$\;

    \ForEach{$C_j \in \{C_1, \dots, C_n\}$}{
        Compute posterior probability:\\
        $P(C_j \mid X_i) = \frac{\text{Numerator}_i}{\text{Denominator}}$\;
    }

    \textbf{Return} all $P(C_j \mid X_i)$
}
\caption{Sample Membership Probability (SMP)}

\label{alg:SMP}
\end{algorithm}

\begin{algorithm}[!ht]
\KwInput{\\
 $S$: Training set\\
$X_i$: A dataset sample\\
 $S_{Core\_region} = S_{Overlapping\_region} =  \varnothing$ \tcp*{Initialization of different regions}
}

\KwOutput{
$S_{Core\_region}, S_{Overlapping\_region}$
} 

\SetKwFunction{FMain}{CCRP}
\SetKwProg{Fn}{Function}{:}{}
\Fn{\FMain{$S$}}{

\vspace{0.2cm}

\ForEach{$j \in \{1,\dots,C_n\}$}{

  \ForEach{$X_i \in S_j$}{

\vspace{0.1cm}

     $\textit{PosteriorProbability} \longleftarrow \textbf{SMP}(X_i)$ 

\vspace{0.1cm}
    \uIf{\scalebox{0.80}{$\text{PosteriorProbability}[i][j] > T_{i,j}$}}{
      $S^{j}_{Core\_region} \longleftarrow S^{j}_{Core\_region} \cup \{X_i\}$   \tcp*{Core region}
    }
    
    \uElseIf{ \scalebox{0.70}{$\exists\ k \ne j,\ k \in \{1, \dots, C_n\},\ \text{where}\ \text{PosteriorProbability}[i][k] > T_{k,k}$}
 }{
        $S^{j}_{Overlapping\_region} \longleftarrow S^{j}_{Overlapping\_region} \cup \{X_i\}$\tcp*{Overlapping region}
      }
    
    \Else{
      \textbf{Delete($X_i$)} \tcp*{ \tcp*{Noisy sample}(very low confidence)}
    }

  }

}

\vspace{0.3cm}

\textbf{Return} $S^{j}_{Core\_region}, S^{j}_{Overlapping\_region}$
}

\caption{Class-Conditional Region Partition (CCRP)}
\label{alg:CCRP}
\end{algorithm}

\subsection{Overlapping  cleaning }
Various scholars have paid attention to class imbalance and its impact on the learning process. However, recent research has shown that the effect of class imbalance on learning performance remains limited if the overlapping issue is not addressed, a phenomenon known as the joint effect of class overlap and class imbalance \cite{Methodology1}.  Class overlapping refers to the situation where minority class samples share the same region as majority class samples, creating ambiguity and making crowded areas difficult to distinguish accurately between classes, which negatively impacts classifier performance even more than class imbalance itself \cite{Introdcution5}.

Motivated by the problem of class overlap and the limitations of previous methods, a new approach for addressing class overlap in multi-class scenarios is proposed in Algorithm~\ref{alg:SOR}. The proposed Sample Overlapping Removal (SOR) method has the advantage of restricting the search and ignoring overlapping samples from overlapping areas without affecting the core areas. This process reduces the overall time required to identify overlapping samples across the dataset. To detect the overlapping samples for a particular class, a set of phases has been applied as follows:

\begin{itemize}
    \item \textbf{Phase (1): Sample mapping and distance arranging} \\
    In this phase, two steps are performed to compute the distance values for overlapping samples and to obtain sequential distance vectors for each overlapping region of a particular class.

    \textbf{Step (1):} \\
    A Euclidean distance, given in Eq.~(\ref{eq:EcludianDistance}), is used as a distance mapping to calculate the distance value of each instance $x_i$ in the overlapping region of a particular class to all the overlapping and all core region samples of other classes.
    \begin{equation}
        D(x_i, y_j) = \sqrt{\sum_{k=1}^{m} (x_{ik} - y_{jk})^2}
        \label{eq:EcludianDistance}
    \end{equation}
    where $x_i = (x_{i,1}, x_{i,2}, \dots, x_{i,m})$ represents the feature vector of the $i$-th overlapping sample of a particular class, $y_j = (y_{j,1}, y_{j,2}, \dots, y_{j,m})$ represents the feature vector of the $j$-th sample of another region (overlapping or core region samples of other classes), and $m$ represents the number of features of a sample. The distance mapping for each overlapping sample of a particular class can be formulated by a function $f(x)$ as in Eq.~(\ref{eq:Median}).
  \begin{equation}
  	\begin{split}
  		f(x_i) = \{ d(x_i, y_j) \mid y_j \in 
  		Core\_region_j \\
  		\cup Overlapping\_region_j \}
  	\end{split}
  	\label{eq:Median}
  \end{equation}
    The result of this process is a set of large distances, including outliers (the farthest distances) and extreme values. To reduce their effect, the median of these distances is taken to represent the central tendency. Finally, the set of median values is returned for each overlapping sample of a particular class, as represented by the function $g(x_i)$.
    \begin{equation}
        g(x_i) = Median(f(x_i))
        \label{eq:11}
    \end{equation}

    \textbf{Step (2):} \\
   After obtaining the median values of all overlapping instances of a class, we sort them in ascending order to create the sequential median distance vector $\Omega$, as in Eq.~(\ref{eq:SortMedina}). To preserve the correspondence between each instance and its distance value, a hash table is used, where the key indicates the overlapping sample and the value represents the median distance of that overlapping sample.
    \begin{equation}
        \Omega = Sort(x_i, g(x_i))
        \label{eq:SortMedina}
    \end{equation}

    \item \textbf{Phase (2): Big jump distance and Z-score for overlapping detection} \\
    In this phase, the vector of instances and their corresponding median distances, $\Omega$, are used to detect big jumps. Typically, samples before a big jump are very close to each other, indicating overlapping samples, while samples after the jump are relatively far apart, corresponding to non-overlapping or safe samples. Unlike using the ordered distances directly, which do not reflect the exact boundary between overlapping and non-overlapping samples, analyzing the jump in distances allows identification of the start and end of the overlapping region. A large jump is detected by computing the difference between consecutive distances in $\Omega$, which identifies the sample where the significant change occurs, as shown in Eq.~(\ref{eq:BigJump}).
    \begin{equation}
        \Delta \Omega(x_i) = \Omega_{i+1} - \Omega_i
        \label{eq:BigJump}
    \end{equation}
    where $\Delta \Omega(x_i)$ represents the change in distance between two successive sample distances.  

    Unlike using a fixed threshold for the distance jump to determine overlapping and non-overlapping samples, a Z-score metric is applied as a dynamic threshold for each class, corresponding to the values in its distance vector. The Z-score in Eq.~(\ref{eq:Zscore}) represents the standardized distance of an observation from its mean and is used to detect outliers, i.e., data points that differ significantly from the majority of observations \cite{Methodology12}.
    \begin{equation}
        Z\text{-score} = \frac{x - \mu}{\sigma}
        \label{eq:Zscore}
    \end{equation}
    where $x$ is the value of a data point, and $\mu$ and $\sigma$ represent the mean and standard deviation, respectively. In the proposed scenario, the data point $x$ corresponds to the change in distance $\Delta \Omega(x_i)$, with $\mu$ and $\sigma$ calculated over all $\Delta \Omega(x_i)$ values.

   \item \textbf{phase 3: Collecting the non-overlapping samples}
After calculating the Z-score for each sample in the overlapping region of a particular class, the non-overlapping samples are selected from this region based on a predefined Z-score threshold.

\textbf{Special Case: absence of a big jump or very small number of overlapping samples}

If the Z-scores of all samples are below the predefined threshold (e.g., < 2), a predefined proportion of the samples (e.g., max(1, int(fraction × number of samples))) is selected as non-overlapping. The selection starts from the farthest samples. This ensures that, even when the proportion is very small (less than 1), at least one non-overlapping sample is retained, as shown in Algorithm~\ref{alg:SOR},( $\#$line 40  ).

\end{itemize}

\begin{algorithm}[t]
\small 
\KwIn{\\
$S$: Training set \\
$S_{Non\_overlapping\_region} = \varnothing$ \\
Hash $= \{ \text{Sample} \mapsto \text{distance} \} = \varnothing$ \\
$\alpha$: percentage of selected non-overlapping samples (e.g., 30\%)
}
\KwOut{
$S_{Non\_overlapping\_region}$
}

\SetKwFunction{FMain}{SOR}
\SetKwProg{Fn}{Function}{:}{}
\Fn{\FMain{}}{

$S^{all}_{Overlapping\_region} \gets \textbf{CCRP}(S)$ \;
$S^{all}_{CORE\_region} \gets \textbf{CCRP}(S)$ \;

\ForEach{$j \in \{1,\dots, C_{m}\}$}{

  \ForEach{$u\in S^{all}_{Overlapping\_region}$}{

    $\forall X^{j}_i \in S^{all}_{CORE\_region},\ \forall X^{u}_i \in S^{all}_{Overlapping\_region}:$ \;

    $\mathcal{D}(X^{u}_i) =
    \{\, \text{Dist}(X^{u}_i, X^{k}_r)
    \mid 
    X^{k}_r \in S^{all}_{CORE\_region}(k)\ 
    \text{or}\ 
    X^{k}_r \in S^{all}_{Overlapping\_region}(k),\ 
    k \neq j
    \,\}$ \;

    $\tilde{d}(X^{u}_i) = \text{median}(\mathcal{D}(X^{u}_i))$ \;

    $\text{Hash} \gets \text{Hash} \cup \{ X^{u}_i \mapsto \tilde{d}(X^{u}_i) \}$ \;
  }

  $\text{SortedPairs} \gets \text{Sort\_by\_Value}(\text{Hash})$ \;
  $\text{OrderedSamples} \gets \{ k \mid \exists v,\ (k, v) \in \text{SortedPairs} \}$ \;
  $\text{OrderingDistances} \gets \{ v \mid \exists k,\ (k, v) \in \text{SortedPairs} \}$ \;

  \ForEach{$f \in \{1,\dots, \text{length}(\text{OrderingDistances})-1\}$}{
      $g \gets \text{OrderingDistances}[f + 1] - \text{OrderingDistances}[f]$ \;
      $Gaps.\text{append}(g)$ \;
  }

  $\mu \gets \text{mean}(Gaps)$ \;
  $\sigma \gets \text{std}(Gaps)$ \;

  $f \gets 1$ \;
  jumpDetected $\gets$ \textbf{false} \;

  \While{$f < \text{length}(Gaps)$}{
      $ZScore_f \gets \frac{Gaps[f] - \mu}{\sigma}$ \;

      \If{$ZScore_f \geq 2$}{
          jumpDetected $\gets$ \textbf{true} \;
          \textbf{break} \;
      }

      $S_{Non\_overlapping\_region} \gets S_{Non\_overlapping\_region} 
      \cup \{ \text{OrderedSamples}[f] \}$ \;

      $f \gets f + 1$ \;
  }

  \If{\textbf{not} jumpDetected}{
    $ q \gets \max \big( 1, \lfloor \alpha \cdot |\text{OrderedSamples}| \rfloor \big)$

      $S_{Non\_overlapping\_region} \gets
      S_{Non\_overlapping\_region} \cup 
      \{\text{last } q \text{ samples of OrderedSamples}\}$ \;
  }

}

\Return{$S_{Non\_overlapping\_region}$}

}

\caption{Samples Overlap Removal (SOR)}
\label{alg:SOR}
\end{algorithm}

\subsection{Data  balancing }

Imbalanced data remains a major challenge, especially in multi-class scenarios where the relationships between classes are more complex \cite{RelatedWork17}. In multi-class classification, the minority classes are not fixed; one class can play the role of the majority class for another class, while it can itself be the minority class for yet another class. To address these issues and prevent the degradation of classifier performance due to the scarcity of the minority class, a data balancing procedure is proposed in Algorithm~\ref{alg:databalacing}, based on the class with the highest number of samples (maximum class) in the dataset, obtained from the previous phases (After overlapping cleaning). The core region and non-overlapping region are combined as the basic data to generate new synthetic samples and augment each minority class to reach the maximum class using the proposed modified SMOTE method. Using the core and non-overlapping regions ensures that the new synthetic data are high-quality samples with higher information content (the sample highly belongs to its class), thus contributing more effectively to learning. The number of new synthetic samples for a particular class is defined as the remaining samples $(k_{remaining})$ of the class with the highest number of samples ($ C^{max}$, maximum class) minus the number of samples of class $j$, i.e. $ k_{remaining} =C^j-C^{max}$, as shown in Algorithm~\ref{alg:databalacing}. The required number of synthetic samples and supplied data are transferred to the proposed modified SMOTE method (OMRP) to create new data for each minority class.
\begin{algorithm}[t]

\small
\LinesNotNumbered

\KwInput{
$S^{all (j)}_{Non\_overlapping\_region} = \varnothing$ \tcc{Non-overlapping samples of class $j$}

$S^{all (j)}_{Core\_region} = \varnothing$ \tcc{Core region samples of class $j$}
}

$S^{C(j)}_{New} = \varnothing$ \tcc{Overall dataset with core and non-overlapping samples}
$k^{j}_{Remaining}$ \tcc{Required balancing number to reach the max class}

\KwOutput{
$S^{*}_{BalancedSet}$ \tcc{Return balanced dataset}
}

\SetKwFunction{FMain}{Data\_balancing}
\SetKwProg{Fn}{Function}{:}{}

\Fn{\FMain{$C_j,x_i$}}{
  \For{$j = 1 \dots C_n$}{
    $S^{all (j)}_{Non\_overlapping\_region} \gets \textbf{SOR()}$\\
    $S^{all (j)}_{Core\_region} \gets \textbf{CCRP()}$\\
    $S^{C(j)}_{New} \gets S^{all (j)}_{Non\_overlapping\_region} \cup S^{all (j)}_{Core\_region}$
  }

  $C^{j}_{Sorted} \gets \textbf{Sorting}(\lvert S^{C_1}_{New}\rvert, \dots, \lvert S^{C_n}_{New}\rvert)$

  \For{$j \in C^{Sorted}$}{
    $k^{j}_{Remaining} \gets |C^{j}_{Sorted} - C^{Max}_{Sorted}|$\\
    $S^{C^{*}_j}_{BalancedClass} \gets \textbf{OMRP}(k^{j}_{Remaining}, S^{C_j}_{New})$\\
    $S^{*}_{BalancedSet} \gets S^{*}_{BalancedSet} \cup S^{C^{*}_j}_{BalancedClass}$
  }

  \Return{$S^{*}_{BalancedSet}$}
}

\caption{Data\_balancing}
\label{alg:databalacing}
\end{algorithm}

\subsection{Oversampling multi-regularization penalty}

Data augmentation of minority classes is an effective way to balance the data and force the classifier to recognize these classes. The SMOTE  algorithm \cite{Methodology2} is a widely used oversampling method based on linear interpolation between k-nearest neighbor minority samples. For each minority class instance $x_i^j $ of a given class $j$, a sample  $x_i$ is selected from its k-nearest neighbors, and a new minority class sample $x^{'}$ is synthesized by linearly interpolating between them using Eq.~(\ref{eq:SmoteEquqtion}).

\begin{equation}
x' = x_i + \text{rand}(0,1) \cdot (x_j - x_i) \label{eq:SmoteEquqtion}
\end{equation}

where $rand(\cdot)$ represents a random number between (0,1) and $x^{'}$ is a new synthesized sample.

Although SMOTE is an effective and popular oversampling technique, it blindly generates synthetic samples through linear interpolation between k-nearest neighbor minority samples \cite{Methodology3}, regardless of whether these samples are located within the minority or the majority class. In this situation, excessive oversampling can lead to the creation of samples that are close to other classes, which increases the complexity of the decision boundary and reduces the clarity of inter-class boundaries, thereby introducing new local overlapping issues between classes. To address this scenario, a modified SMOTE method, OMRP, has been proposed to balance the data, ensuring clear inter-class boundaries and reducing 
the creation of new overlapping areas, as shown in Algorithm~\ref{alg:OMRP}. We propose a multi-regularization penalty to control the position of newly generated synthetic samples. Typically, the regularization penalty aims to push synthetic samples of each particular minority class far away from other classes (where $k \neq j$) and closer to their own class. This control of position can be modeled as the distance between the synthetic sample of a particular minority class, its own class, and the positions $L_i(x_j, x_{syn}x^{'})$, which is used to compute the distance between each synthetic sample of a particular minority class $j$ and all samples of the same class $j$. 
Similarly, the Euclidean distance norm $L_j(x_k, x_{syn}x^{'})$ is used to compute the distance between each synthetic sample of a particular minority class j and all samples of each other class k. The multi-regularization penalty control, $L_i$  and $L_k$, is defined as in Eq.~(\ref{eq:EcludianNorm1}) and (\ref{eq:EcludianNorm2}).

\begin{equation}
L_i(x_j, x_{\text{syn}}') = \lVert x_{\text{syn}}' - x_j \rVert
\label{eq:EcludianNorm1}
\end{equation}

\begin{equation}
L_k(x_k, x_{syn}') = \| x_{syn}' - x_k \| \label{eq:EcludianNorm2}
\end{equation}

In this context, $x_j$, $x_k$, and $x_{syn}^{'} $ represent the sample of the particular class j, the sample of another class k, and the synthetic sample, respectively. According to the definitions of $L_j$ and $L_k$, the feature distance between the synthetic samples of the minority class and both their own class and other classes can be modeled as an optimization problem. The objective is to improve the location of the new synthetic samples in the overlapping region between classes by maximizing their distance from other classes while minimizing their distance to their own class. The mathematical formulation is given by Eq.~(\ref{eq:EcludianNorm3}) and (\ref{eq:EcludianNorm4}).

\begin{equation}
L_{\text{Min}}(x_j, x_{syn}') = \min\big( Dist(x_j, x_{syn}') \big) \label{eq:EcludianNorm3}
\end{equation}

\begin{equation}
L_{\text{Max}}(x_k, x_{syn}') = \max\big(Dist(x_k, x_{syn}') \big) \label{eq:EcludianNorm4}
\end{equation}

Where $Dist$ represents the Euclidean distance norm. From Eq.~(\ref{eq:EcludianNorm3}) and (\ref{eq:EcludianNorm4}), the optimization problem of the multi-regularization penalty can be written as in Eq.~(\ref{eq:EcludianNorm5}). For every new synthetic sample that does not satisfy this condition, it is discarded, and only those samples that meet the condition are retained until the classes are balanced.

\begin{equation}
L_{\text{Min}}(x_j, x_{syn}') < L_{\text{Max}}(x_k, x_{syn}') \label{eq:EcludianNorm5}
\end{equation}

\begin{algorithm}[t]

\LinesNotNumbered
\small

\KwInput{
$S_{Syn} = \varnothing$ \\
$Dist^{C_z}_{i} := \min_{x \in \mathcal{C}_z} \text{distance}(x_y, x)$ \tcp{Minimum distance between the synthetic sample and all samples in other classes} \\
$Dist^{j}_{i} := \min_{x \in j} \text{distance}(x_y, x)$ \tcp{Minimum distance between the synthetic sample and samples in current class $j$}
}

\KwOutput{
$S^{*}_{BalancedClass}$ \tcp{Return the balanced set for class $j$ including generated synthetic samples}
}

\setcounter{AlgoLine}{0} 
\LinesNumbered

\SetKwFunction{FMain}{OMRP}
\SetKwProg{Fn}{Function}{:}{}

\Fn{\FMain{$k, S^{C_j}_{New}$}}{
\Indp

\nl \While{$x_i \in S^{C_j}_{New}$ \textbf{and} $j < K$}{

    \nl $\alpha = \text{rand}(0,1)$ \tcp{Random interpolation factor $\alpha$ in $[0,1]$}
    \nl neighbors = find\_k\_nearest\_neighbors($x_i$) \tcp{Find k-nearest neighbors of $x_i$}
    \nl $\bar{x}_i =$ random\_choice(neighbors) \tcp{Select one neighbor randomly}
    \nl $\hat{x}_i = x_i + \alpha \cdot (\bar{x}_i - x_i)$ \tcp{Generate synthetic sample by interpolation}

    \textbf{Multi-Regularization penalty:}

    \nl $Dist^{C_z}_{i} \gets L(\hat{x}_i, x_y), \forall x_y \in S^{C_z}_{New}, z \neq j$ \tcp{Compute penalty to all other classes}
    \nl $Dist^{j}_{i} \gets L(\hat{x}_i, x_i)$ \tcp{Compute penalty for current class $j$}

    \nl \If{$Dist^{j}_{i} \le Dist^{C_z}_{i}$}{
        $S_{Syn} \gets S_{Syn} \cup \{\hat{x}_i\}$ \tcp{Keep sample if distance to current class <= distance to other classes}
        $S^{C^{*}_j}_{BalancedClass} \gets S^{C^{*}_j}_{BalancedClass} \cup S_{Syn}$
    }
}

\nl \Return{$S^{*}_{BalancedClass} \gets S^{*}_{BalancedClass} \cup S^{C^{*}_j}_{BalancedClass}$} \tcp{Return balanced set for class $j$}

\Indm
}

\caption{Oversampling Multi-Regularization Penalty (OMRP)}

\label{alg:OMRP}
\end{algorithm}

\subsection{Ensemble learning pruning }

Ensemble learning is a powerful machine-learning technique. It transforms a single weak classifier (e.g., a decision stump) into a strong classifier by combining multiple weak learners, reducing generalization error and improving prediction performance \cite{Methodology4}. However, training all base classifiers is pricier, and increasing model complexity grows at least linearly with ensemble size, which increases computational time and limits practical applicability \cite{Methodology5}. Additionally, model diversity can impact ensemble performance. Ensemble learning employs aggregation strategies (e.g., hard majority voting) that treat all base classifiers equally, which can allow less-contributing (redundant) classifiers to negatively affect overall performance \cite{Methodology6}. To address this, ensemble learning pruning has been used in the literature to select the optimal subset of base classifiers with higher contribution. In the proposed framework, we introduce a new ensemble pruning approach. Unlike previous studies  \cite{Methodology8} that employ the Jaya algorithm for optimizing meta-learning parameters or feature selection, or other common metaheuristic methods (e.g., genetic algorithms, swarm intelligence) for pruning \cite{Methodology7}, the use of Jaya for ensemble pruning remains largely unexplored.We adapt the Jaya method from numerical optimization to categorical selection in order to choose the optimal set of base classifiers, defining its objective function as modeled in Algorithm~\ref{alg:OEL}.

\subsubsection{ Jaya optimization}

Jaya is a recent stochastic population-based metaheuristic optimization algorithm \cite{Methodology8}. Unlike existing population-based heuristic algorithms (e.g., evolutionary algorithms, swarm intelligence), Jaya has several advantages. It is a parameterless algorithm, requiring only two parameters: population size and number of iterations. Additionally, it is a faster metaheuristic in terms of convergence efficiency compared to other algorithms, such as Differential Evolution (DE), allowing it to reach optimal solutions in less time \cite{Methodology9}. This algorithm employs a best-and-worst solution strategy: at each iteration, it attempts to move towards the optimal solution while avoiding the worst one. The Jaya algorithm can be applied to a wide range of optimization problems, including discrete (categorical) ones, and its efficient search strategy enables it to find optimal solutions while effectively avoiding local minima compared to many other algorithms  \cite{Methodology10}. In this context, the Jaya algorithm has been adapted for ensemble pruning to select highly contributing base classifiers while disregarding less effective ones, as described in Algorithm~\ref{alg:OEL}, with the following corresponding steps.

Let $\Psi$ denote the set of base classifiers. In our work, each chromosome encodes a candidate solution, where each gene represents a classifier: a value of 1 indicates that the classifier is selected, whereas 0 indicates that it is excluded from the ensemble. Thus, every solution can be represented as a binary vector, e.g.,$\{0,1,0,0,0\}$.

The $i^{th}$ solution in generation tis denoted by $\Psi_i^{(t)}$, and the  $j^{th}$ position of the  $i^{th}$ solution in generation trepresents the corresponding gene (classifier), denoted by $\Psi_{(i,j)}^{((t))}$. The steps for applying the Jaya algorithm to optimize the set of potential classifiers are as follows.

\begin{itemize}
    \item \textbf{Step (1). Population initialization and digitization}  Initially, random values in the range $[0,1]$ are generated to form a solution matrix of population size $N$ (i.e., $i=1,2,\dots,N$), with the number of decision variables corresponding to the number of base classifiers (i.e., $j=1,2,\dots,m$).
    
A digitization process is then applied to convert the continuous Jaya values into binary form, producing a binary vector indicating selected and unselected classifiers, using Eq.~(\ref{eq:Digitilization}).

\begin{equation}
\Psi_{i,j}^{(t)} =
\begin{cases}
1, & \Psi_{i,j}^{(t)} > 0.5 \\
0, & \text{otherwise}
\end{cases}
\label{eq:Digitilization}
\end{equation}

In this context, classifiers with values above 0.5 are set to 1, and those below to 0. Unlike studies using a dynamic threshold, we use a fixed 0.5, as most weak classifiers perform slightly better than random, allowing broader selection and avoiding cases where the threshold exceeds all values.

 \item \textbf{Step (2).  Fitness function}
A fitness (objective) function is a vital criterion for evaluating the performace of a candidate solution (chromosome). The fitness function of a chromosome in a population of size Nwith mdecision variables (i.e., a set of classifiers) can be described as shown in Eq.~(\ref{eq:fitnessfunction}).

\begin{equation}
f(i) = f\big(\Psi_i^{(t)}\big) \label{eq:fitnessfunction}
\end{equation}

In this context, the F-score was used as the fitness function, following the approach in \cite{Methodology11}, where it was applied in the context of a genetic algorithm.

 \item \textbf{Step (3). Updating population }
 
The worst and best candidate solutions in generation $t$ are used to generate the new population. These solutions are defined as those with the lowest and highest fitness values (F-score), denoted by $\Psi_{\text{best},j}^{(t)}$ and $\Psi_{\text{worst},j}^{(t)}$, respectively.  
Considering the best and worst solutions of generation $t$ and the $j$-th position of the old candidate solution $\Psi_{i,j}^{(t)}$, the updated candidate solution at generation $t+1$ is $\Psi_{i,j}^{(t+1)}$, as described in Eq.~(\ref{eq:Jayamodification}).

\begin{equation}
\Psi_{i,j}^{(t+1)} = \Psi_{i,j}^{(t)} + r_1 \, \bigl| \Psi_{\text{best},j}^{(t)} - \Psi_{i,j}^{(t)} \bigr| - r_2 \, \bigl| \Psi_{\text{worst},j}^{(t)} - \Psi_{i,j}^{(t)} \bigr|
\label{eq:Jayamodification}
\end{equation}

Where  $r_1 $ and  $r_2 $ are two random number between 0 and 1.

 \item \textbf{Step (4).  Termination conditions and digitization}
The Jaya algorithm is an iterative process that continues until the optimal solution is reached. The search can be terminated either by setting the maximum number of iterations or by reaching the maximum value of the objective function. Since the proposed pruning ensemble evaluates the heterogeneity of the base classifiers, the maximum value of the objective function cannot be determined before the algorithm is executed. Therefore, using the maximum objective function as a stopping criterion may result in prolonged execution times, which is impractical in practice. In this paper, we terminate the pruning search process based on the maximum number of iterations., $t_{max}$.
Finally, the best candidate solution is obtained at the end of the iterations. This solution represents a pruned ensemble containing a smaller set of classifiers that contribute most significantly. However, the final vector consists of continuous numerical values, which require digitization. To achieve this, Eq.~(\ref{eq:Digitilization}) is applied to convert the continuous values into categorical form, where 1 represents a selected classifier and 0 represents an unselected classifier. The optimal pruned ensemble is thus a vector containing only the selected classifiers.

\end{itemize}

\begin{algorithm}[t]

\LinesNotNumbered

\KwInput{\\
$S_{BalancedSet}$ \tcp*{A balanced dataset}

$\mathcal{L}_i^{Init} $ \tcp*{Initial pool of base classifiers (e,g, DT. KNN)}
}

\KwOutput{
$\mathcal{L}_{i}^{Optimal}$ \tcp*{A set of optimal base classifiers selected using Jaya optimization}
}

\SetKwFunction{FMain}{OEL\_Jaya}
\SetKwProg{Fn}{Function}{:}{}

\Fn{\FMain{$C_{j}, x_i$}}{
\Indp

$S_{BalancedSet} \gets \textbf{Data\_balancing}$ 

Construct initial pool of base classifiers: $\mathcal{L}_i^{Init} = \{\Psi_1, \Psi_2, \dots, \Psi_n\}$\;

Initialize population $P_{i,j}, \quad i = 1, \dots, N, \quad j = 1, \dots, m$ \tcp*{Each solution encodes a candidate ensemble}

\ForEach{$t = 1$ \KwTo $T_{max}$}{
    
    \tcp{Step 1: Evaluate fitness of each candidate solution}
    Compute fitness $f(i) = f(\Psi_i^{(t)})$ for all $i$, e.g., using F-score \tcp*{Equation (22)}
    
    Identify best and worst solutions in current generation: 
    $\Psi_{\text{best}}^{(t)}$, $\Psi_{\text{worst}}^{(t)}$\;
    
    \tcp{Step 2: Update population using Jaya rule}
    \ForEach{$i = 1$ \KwTo $N$}{
        \ForEach{$j = 1$ \KwTo $m$}{
            $r_1, r_2 \gets$ random numbers in $[0,1]$\;
            $\Psi_{i,j}^{(t+1)} \gets \Psi_{i,j}^{(t)} + 
            r_1 \left| \Psi_{\text{best},j}^{(t)} - \Psi_{i,j}^{(t)} \right| -
            r_2 \left| \Psi_{\text{worst},j}^{(t)} - \Psi_{i,j}^{(t)} \right|$ \tcp*{Equation (23)}
        }
    }
    
 \tcp{Step 3: Final selection}
Pick the optimal solution with highest fitness: $j^* = \arg\max_i f(\Psi_i^{(T_{max})})$\;
Return corresponding subset of classifiers:
\[
\mathcal{L}_i^{Optimal} = \mathcal{L}_i^{Init}(j^*)
\]

}

  \tcp{Step 4: Digitization}
    Convert continuous values to binary using threshold $0.5$:
    \[
    \Psi_{i,j}^{(t+1)} = 
    \begin{cases} 
    1 & \text{if } \Psi_{i,j}^{(t+1)} > 0.5 \\
    0 & \text{otherwise}
    \end{cases}
    \] 

\Indm
}

\caption{Optimized Ensemble Learning (OEL)}
\label{alg:OEL}
\end{algorithm}

\section{Experimetation setup}
In this section, a series of experiments is conducted to evaluate the effectiveness of the proposed framework. At the data level, the experiments aim to improve data quality by addressing class imbalance,  minimizing class overlap, and cleaning noisy data. At the algorithmic level, they assess the improvement in performance and reliability of both multi-class and binary classification models, in accordance with the objectives outlined earlier. The following subsections describe the datasets, evaluation metrics, comparative methods, and other components of the experimental setup.

\subsection{Imbalanced dataset}
Several binary and multi-class imbalanced datasets are used to evaluate the proposed method. All datasets are publicly available from the KEEL(https://sci2s.ugr.es/keel/datasets.php) and UCI (https://archive.ics.uci.edu) repositories. These datasets are characterized by their names, number of instances, number of features, and imbalance ratio (IR). The IR measures the degree of skewness (imbalance) between classes. 

For binary classification, the IR is defined as the ratio between the number of samples in the majority class and the number of samples in the minority class (e.g., $N_{\text{majority}} / N_{\text{minority}}$), while in multi-class imbalanced datasets, the IR is expressed as the ratio between the number of samples in the largest class and the smallest class (e.g., $C_{\text{max}} / C_{\text{min}}$) \cite{Methodology11}. Tables~\ref{tab:MultiClassDataset} and ~\ref{tab:BinaryClassDataset} summarize the datasets and their characteristics.
\begin{table}[H]  
\centering
\caption{Description of multi-class imbalanced datasets.}
\begin{tabular}{l c c c c}
\hline
\textbf{Datasets} & \textbf{Features} & \textbf{Instances} & \textbf{IR} & \textbf{No. of classes} \\
\hline
Contraceptive & 9  & 1473 & 1.89  & 3  \\
Dermatology   & 34 & 366  & 5.55  & 6  \\
Balance       & 4  & 625  & 5.88  & 3  \\
New-thyroid   & 5  & 215  & 5     & 3  \\
Zoo           & 16 & 101  & 10.25 & 7  \\
Yeast         & 8  & 1484 & 92.6  & 10 \\
Pageblocks    & 10 & 548  & 92.6  & 5  \\
Vehicle       & 18 & 846  & 1.09  & 4  \\
Glass         & 9  & 214  & 8.44  & 6  \\
Heart-disease Hungarian & 13 & 294 & 12.53 & 5  \\
Winequality-red & 11 & 1599 & 68.1 & 6  \\
Bioconcentration & 10 & 729 & 7.18 & 3  \\
Vertebral     & 6  & 310  & 2.5   & 3  \\
\hline
\end{tabular}

\label{tab:MultiClassDataset}
\end{table}

\begin{table}[H]  
\centering
\caption{Description of binary-class imbalanced datasets.}
\begin{tabular}{l c c c}
\hline
\textbf{Datasets} & \textbf{Instances} & \textbf{Features} & \textbf{IR} \\
\hline
Yeast1                       & 1484 & 8  & 2.46  \\
Ecoli1                        & 336  & 8  & 3.36  \\
Yeast-1\_vs\_7                & 459  & 7  & 14.30 \\
Yeast2vs8                     & 482  & 9  & 23.10 \\
Glass2                        & 214  & 9  & 11.59 \\
Abalone9vs18                  & 731  & 9  & 16.40 \\
Pima                          & 768  & 8  & 1.87  \\
Yeast4                        & 1484 & 9  & 28.10 \\
Yeast6                        & 1484 & 9  & 41.40 \\
Messidor\_features             & 1151 & 20 & 1.13  \\
Biodeg                        & 1055 & 42 & 1.96  \\
Vehicle3                       & 846  & 18 & 2.99  \\
New-thyroid1                   & 215  & 5  & 5.14  \\
Segment0                       & 2308 & 19 & 6.02  \\
Yeast3                         & 1484 & 8  & 8.10  \\
Ecoli3                         & 336  & 7  & 8.60  \\
Yeast-2\_vs\_4                 & 514  & 8  & 9.08  \\
Ecoli-0-1-4-7\_vs\_5-6         & 332  & 6  & 12.28 \\
Ecoli4                         & 336  & 7  & 15.80 \\
Wilt                           & 4839 & 5  & 17.54 \\
Ecoli-0-1-3-7\_vs\_2-6         & 281  & 7  & 39.14 \\
Abalone-20\_vs\_8-9-10         & 1916 & 10 & 72.69 \\
\hline
\end{tabular}

\label{tab:BinaryClassDataset}
\end{table}

\subsection{Assessment metrics}
To evaluate the performance of the proposed framework, a set of popular metrics was used, including G-Mean, F-score, recall, and precision \cite{Introdcution6} \cite{Introdcution18}. The G-Mean metric uses information from both classes (majority and minority), which balances classification performance, as shown in Eq.~(\ref{eq:G-means}). The F-score is the harmonic mean of precision and recall, focusing on the positive class, as shown in Eq.~(\ref{eq:f-score}). Precision is the proportion of true positives among all predicted positive samples, as shown in Eq.~(\ref{eq:Precision}). Recall, known as the sensitivity metric, represents the fraction of all positive samples that are correctly predicted as positive, as shown in Eq.~(\ref{eq:Recall}).
\begin{equation}
\text{G-means} = \sqrt[n]{\prod_{i=1}^{n} \text{Recall}_i} \label{eq:G-means}
\end{equation}

\begin{equation}
\text{F-Score} = \frac{1}{n} \sum_{i=1}^{n} F1_i \label{eq:f-score}
\end{equation}

\begin{equation}
\text{Precision} = \frac{\text{True Positive}}{\text{True Positive} + \text{False Positive}} \label{eq:Precision}
\end{equation}

\begin{equation}
\text{Recall} = \frac{\text{True Positive}}{\text{True Positive} + \text{False Negative}} \label{eq:Recall}
\end{equation}

To quantify the effectiveness of the proposed framework in overlapping reduction, the overlapping ratio metric (OR) is used \cite{Introdcution6}. Specifically, $OR_k$ represents the degree of overlap for class $k$, $OR(i,j)$ measures the overlap between the class pair $(i,j)$, and $OR_{\text{dataset}}$ denotes the overall overlap of the entire dataset. The definitions of these metrics are given in Eqs.~(\ref{eq:OR1})--(\ref{eq:OR3}).

\begin{equation}
OR_k = \frac{N_k^{Ov}}{N_k}  \label{eq:OR1}
\end{equation}

\begin{equation}
OR_{(i,j)} = \frac{1}{2} \left( \frac{N_{i,j}^{Ov}}{N_i} + \frac{N_{j,i}^{Ov}}{N_j} \right)  \label{eq:OR2}
\end{equation}

\begin{equation}
OR_{\text{dataset}} = \frac{1}{c} \sum_{k=1}^{c} OR_k  \label{eq:OR3}
\end{equation}
In this context, $N_k^{Ov}$ indicates the number of overlapping instances in class $k$; $N_{i,j}^{Ov}$ indicates the number of instances in class $i$ that overlap with class $j$. Similarly, $N_{j,i}^{Ov}$ indicates the number of instances in class $j$ that overlap with class $i$.

\subsection{Reference methods and parameters}
To demonstrate the effectiveness of the proposed framework, an extensive experimental study has been conducted against recent state-of-the-art methods. For methods with available results in their publications, we used those reported values directly; for others, where results were not available, the implementations were reproduced on our datasets based on their publicly available code. All the comparison methods were chosen based on shared key characteristics and objectives with the proposed framework (e.g., addressing class imbalance, minimizing overlapping, or using decomposition strategies), making them well-suited for comparison. In the context of multi-class imbalanced data, the selected comparison methods are divided into two groups: ad hoc methods (directly addressing multi-class problems) and OVO binary decomposition methods (applying one-vs-one or one-vs-all strategies with binary methods). Additionally, a comparative analysis has been conducted against imbalanced binary class methods. For the hyperparameter settings, we set the parameters of these methods based on the recommendations in their original literature and the default values in their public code, as shown in Table~\ref{tab:parametersConfiguration}.

\begin{table}[H]  
\centering
\caption{Overview of the parameter settings.}
\begin{tabular}{l c}
\hline
\textbf{Methods / References} & \textbf{Parameter Values}  \\
\hline
SMOTE-CDNN [9]                     & $k\_neighbors = 21$ \\
ECDNN \cite{Experimentation1}\cite{Experimentation2}                        & $k\_neighbors = 21, \; n \text{ (vote hyper-parameter)} = 2$ \\
SAMME.C2 \cite{Introdcution19}            & Number of iterations  = 200 \\
CPS-3WS \cite{Experimentation3}      & \makecell[l]{$\alpha = 0.4, \; \beta = 0.5, \; k\_neighbors = 5, \; $ \\Base classifier: decision tree }\\

\makecell[l]{Counterfactual \\SMOTE \cite{Experimentation4}}           & \makecell[l]{Strategy: flexible, Number of optimization steps = 3 \\ Number of iterations = 3 \\ Base classifier: random forest} \\

MLOS \cite{Experimentation5}                           & $\alpha = 0.4, \; k\_neighbors = 5, \; $Base classifier: SVM \\
HSCF \cite{Experimentation6}                          & Default, Base classifier: decision tree \\
GDDSAD \cite{Experimentation7}                          & $k\_neighbors = 5, \; $Number of learners = 30 \\
\hline
\end{tabular}

\label{tab:parametersConfiguration}
\end{table}

\subsection{Base classifiers}

For the prediction task, four weak base classifiers are selected to perform ensemble pruning and generate the final prediction. The k-nearest neighbor (kNN) classifier is a popular machine-learning, non-parametric algorithm. It is an instance-based method that can be applied to both classification and regression tasks \cite{Methodology11}. The number of neighbors is set to 3 to keep the classifier as weak as possible. The decision tree (DT) is also a widely used algorithm, frequently integrated into ensemble learning through the concept of a decision stump (number of trees: 1). DTs offer advantages such as interpretability, high robustness, and fast running speed\cite{Methodology11}. Gaussian Naïve Bayes and ExtraTreeClassifier are also included. For all classifiers, we use the implementations from the Scikit-learn library.  It is worth noting that the proposed framework is not limited to these classifiers and can be applied to other strong classifiers (e.g., SVM, ANN) as well. These classifiers are chosen based on the assumption that ensemble methods are typically designed for weak learners. In addition, weak classifiers generally perform only slightly better than random guessing (50\%), which makes them more likely to include low-contributing learners. This characteristic demonstrates the effectiveness of the Jaya-based ensemble pruning strategy in effectively removing such low-quality classifiers and improving performance, which relates to our earlier objectives.

\subsection{Experimental design and configuration}

This subsection describes the dataset preparation and experimental setup. To ensure result consistency and reliable generalization, the final performance scores are obtained using 5-fold stratified cross-validation (5CV). In each iteration, one fold is used for testing while the remaining folds are used for training. The final performance is reported as the aggregated results across all folds. To ensure consistency of the results, the experiment has been repeated 10 times to increase result stability, with the results aggregated across all runs for all relevant assessment metrics.

\section{Experiments results and discussions}
This section presents the experimental results of the proposed framework, including ablation studies, overlap minimization, class imbalance handling, and comparisons with state-of-the-art methods for both multi-class and binary classification. The evaluation results are discussed and interpreted in relation to the research questions. The results are presented under the following subheadings.

\begin{itemize}
\item    \textbf{RQ1:} To what extent does the proposed framework improve predictive performance and reliability on imbalanced datasets?
\item  \textbf{RQ2:} Can the framework effectively handle class overlap?

\item  \textbf{RQ3:} How does the proposed framework’s effectiveness compare to multiclass and decomposition-based methods?
\item  \textbf{RQ4:} How effective is the proposed framework in binary classification?
\item  \textbf{RQ5:} Does the proposed solution handle noisy data?

\item  \textbf{RQ6:}How does each component of the proposed framework contribute to overall performance?
\end{itemize}

\subsection{ Ablation study}
In this section, we conduct an ablation study to determine the most suitable algorithms and parameter settings for the proposed framework in addressing, first, data-level problems such as class imbalance, class overlap, and noisy data, and second, algorithm-level problems related to classification models. The paper systematically evaluates the fundamental components of the approach, as detailed in the following subsections

\subsubsection{Class-conditional region partition}
\begin{figure*}[htbp]
    \centering
    
    \subfloat[\centering Contraceptive  dataset]{{\includegraphics[width=5.9cm,height=5cm]{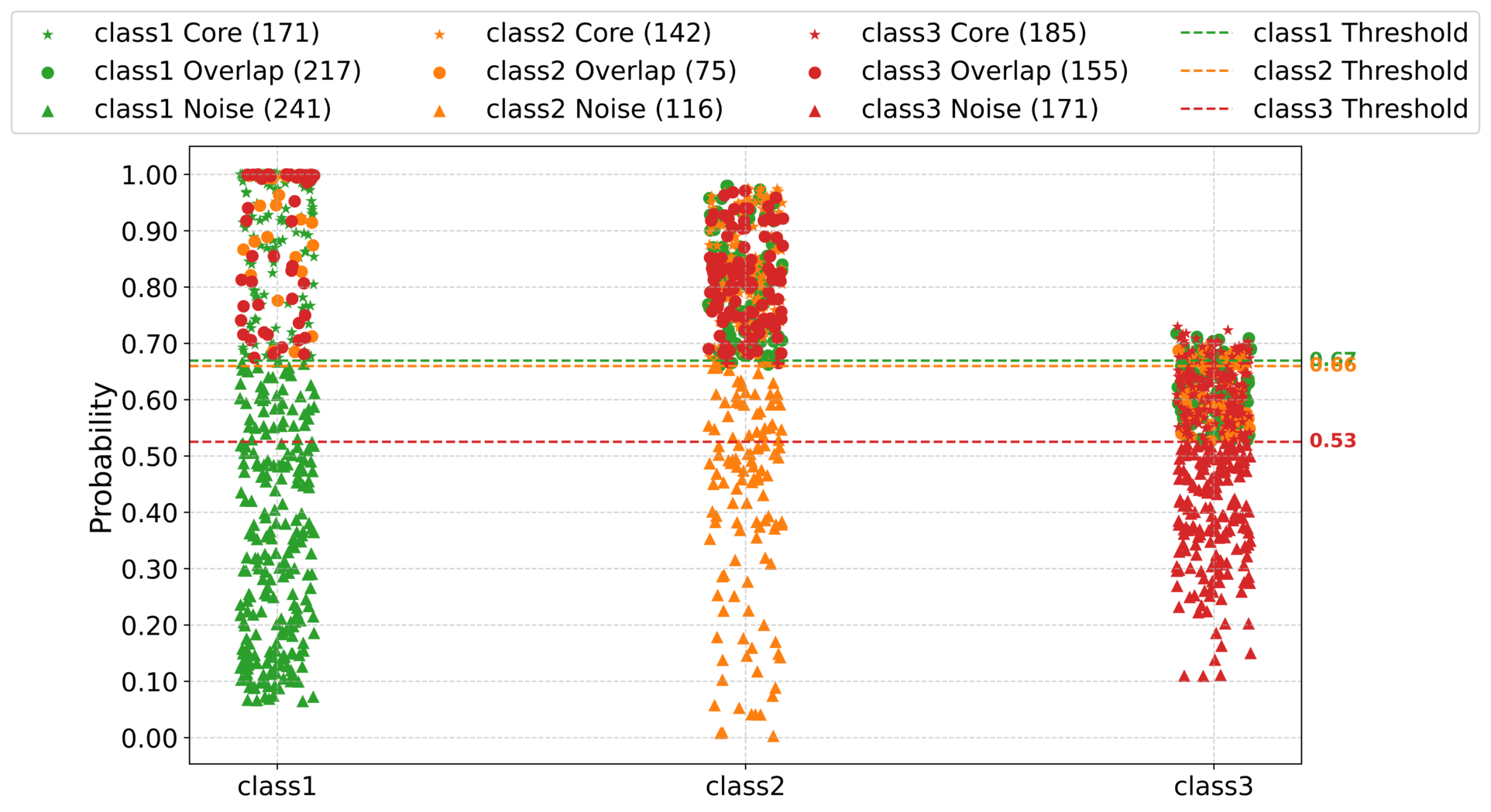} }}
    \subfloat[\centering Vehicle dataset]{{\includegraphics[width=5.9cm,height=5cm]{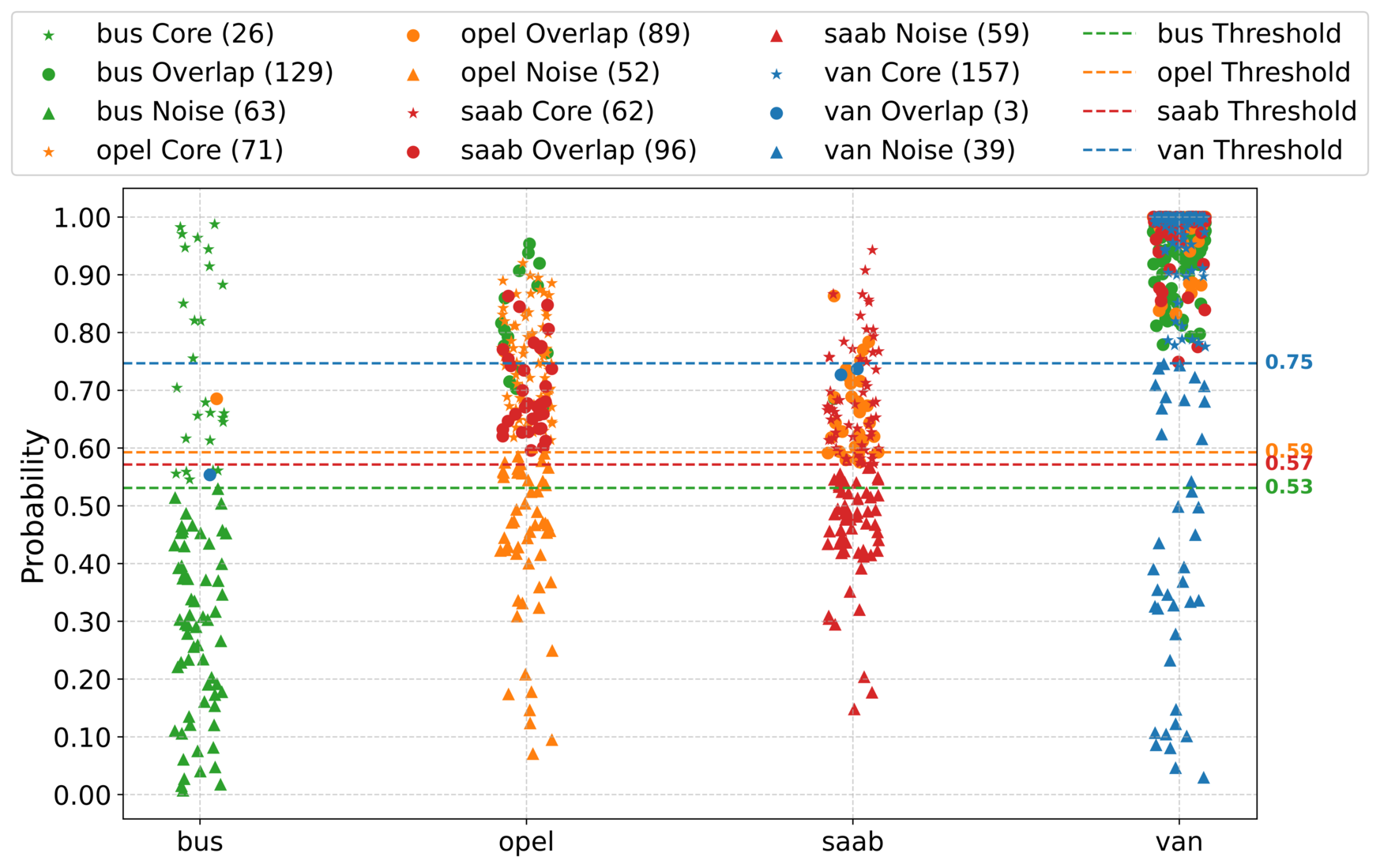} }}
    \subfloat[\centering Vertebral dataset]{{\includegraphics[width=5.9cm,height=5cm]{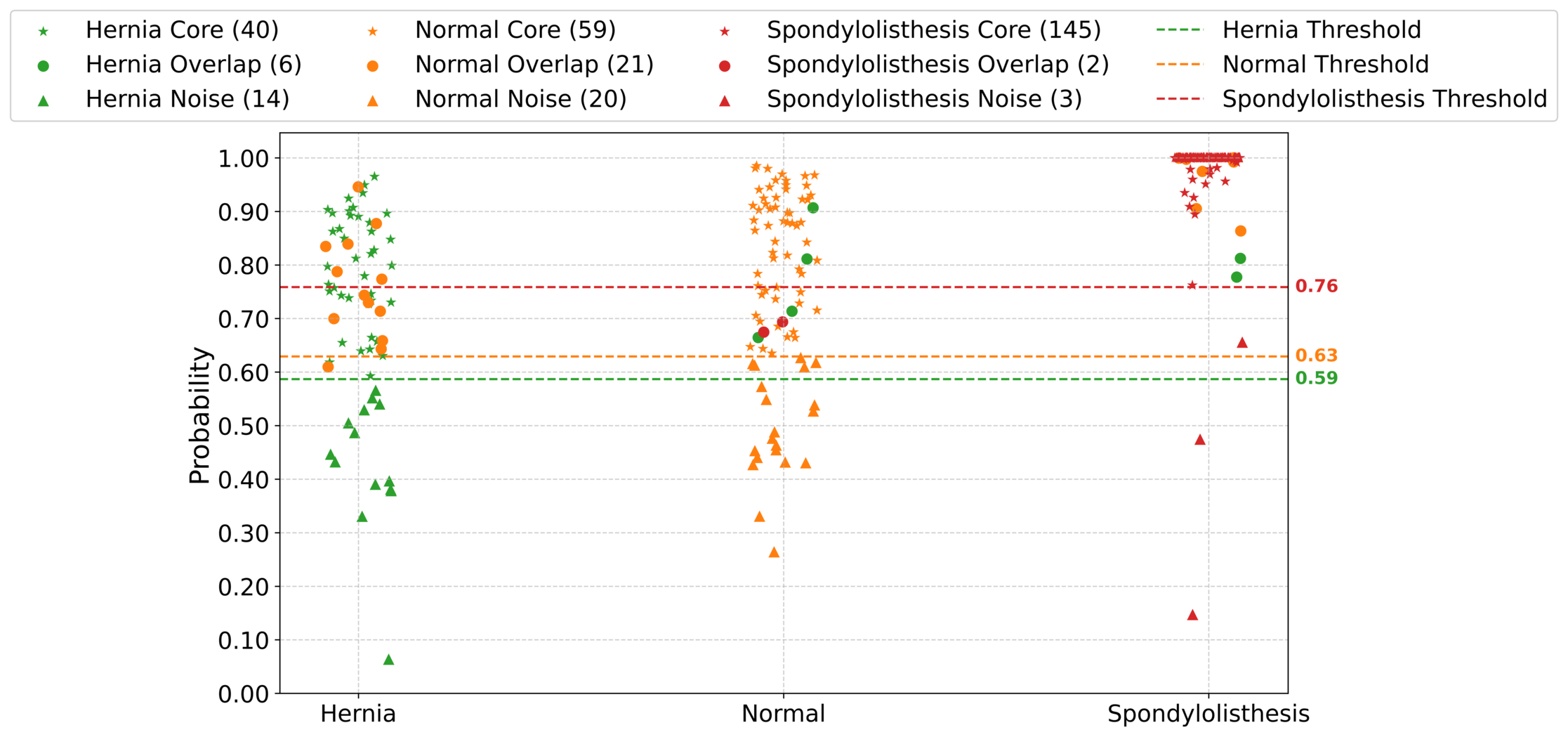} }}

    \caption{Conditional region partition  with  lower imbalanced ratio ( IR$\le$ 3 ).}
    \label{fig:Class-Conditional3}
\end{figure*}

\begin{figure*}[htbp]
    \centering
    
    \subfloat[\centering Balance  dataset]{{\includegraphics[width=5.9cm,height=5cm]{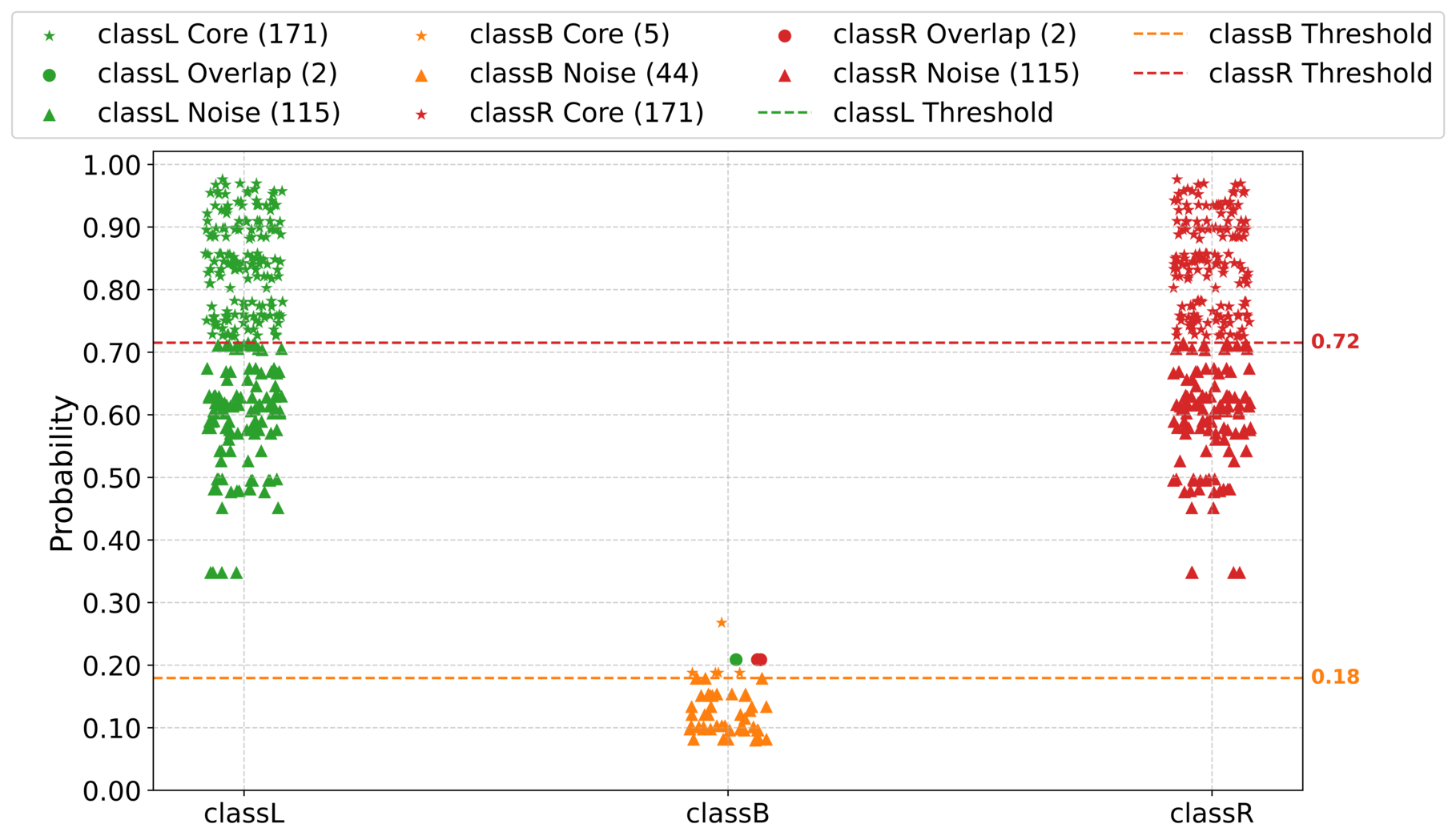} }}
    \subfloat[\centering Bioconcentration dataset]{{\includegraphics[width=5.9cm,height=5cm]{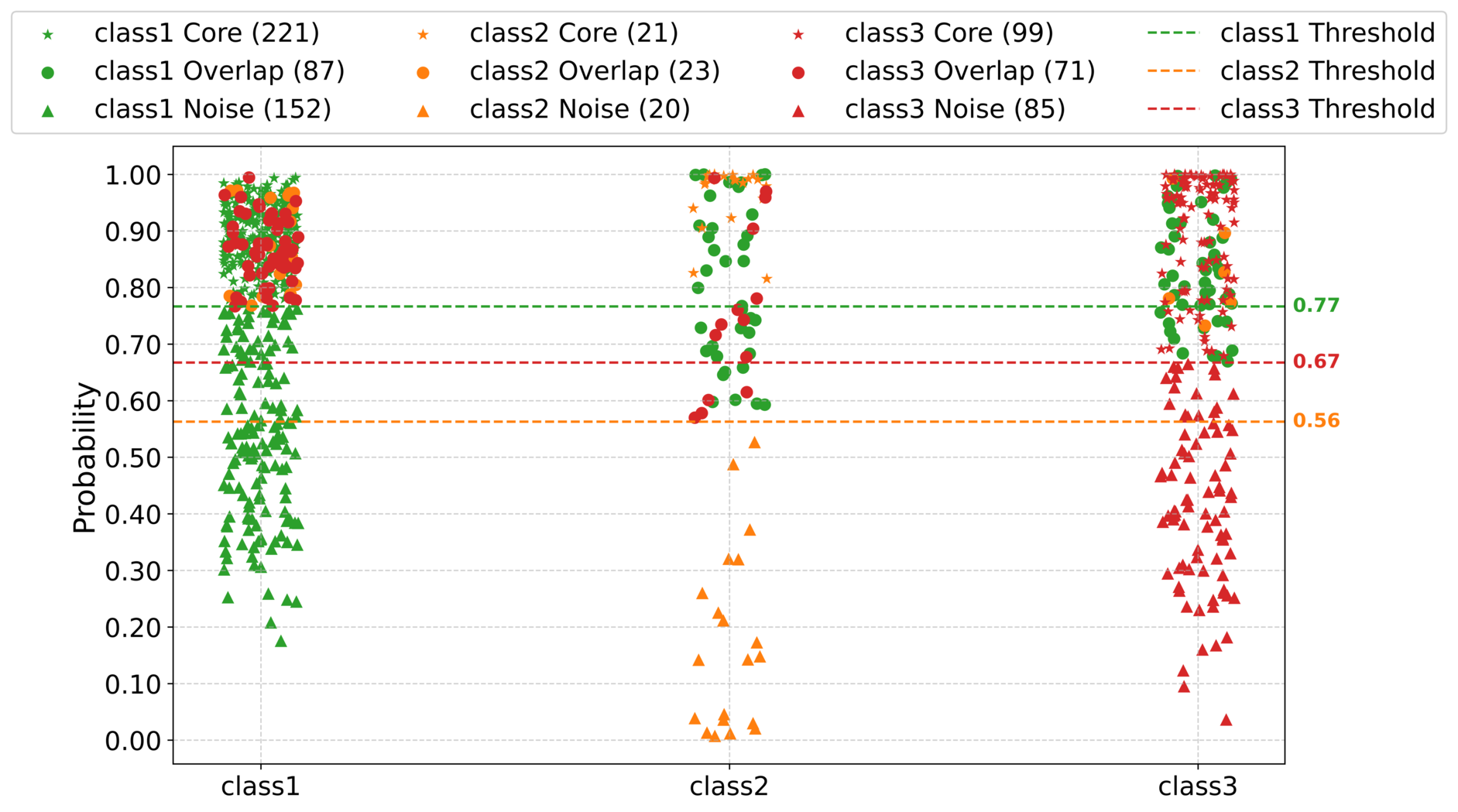} }}
    \subfloat[\centering Glass dataset]{{\includegraphics[width=5.9cm,height=5cm]{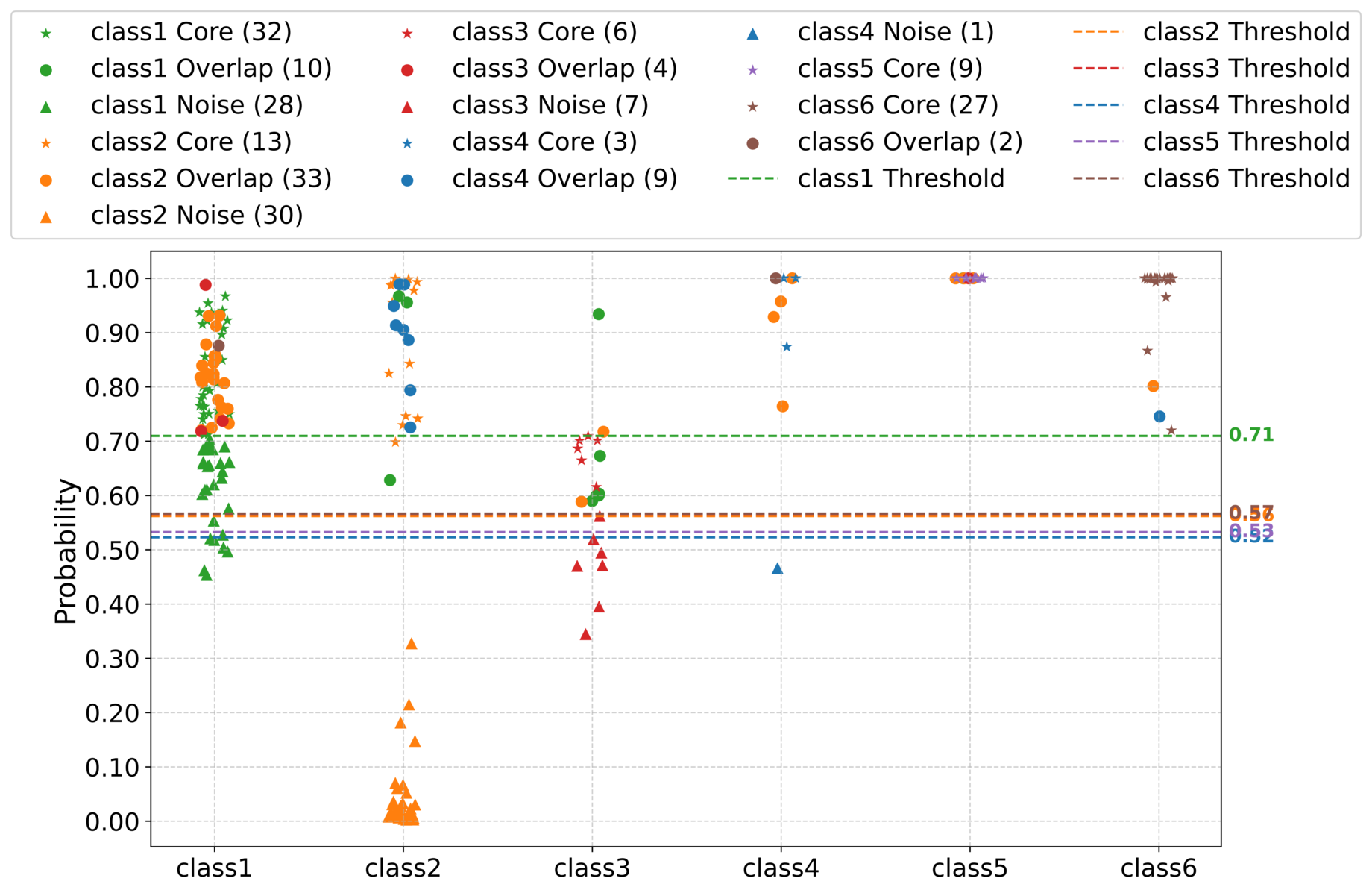} }}

    \caption{Conditional region partition with medium imbalanced ratio (3 $\le$ IR $\le$ 9).}
    \label{fig:Class-Conditional39}
\end{figure*}

\begin{figure*}[htbp]
    \centering
    
    \subfloat[\centering  Hungarian  dataset]{{\includegraphics[width=5.9cm,height=5cm]{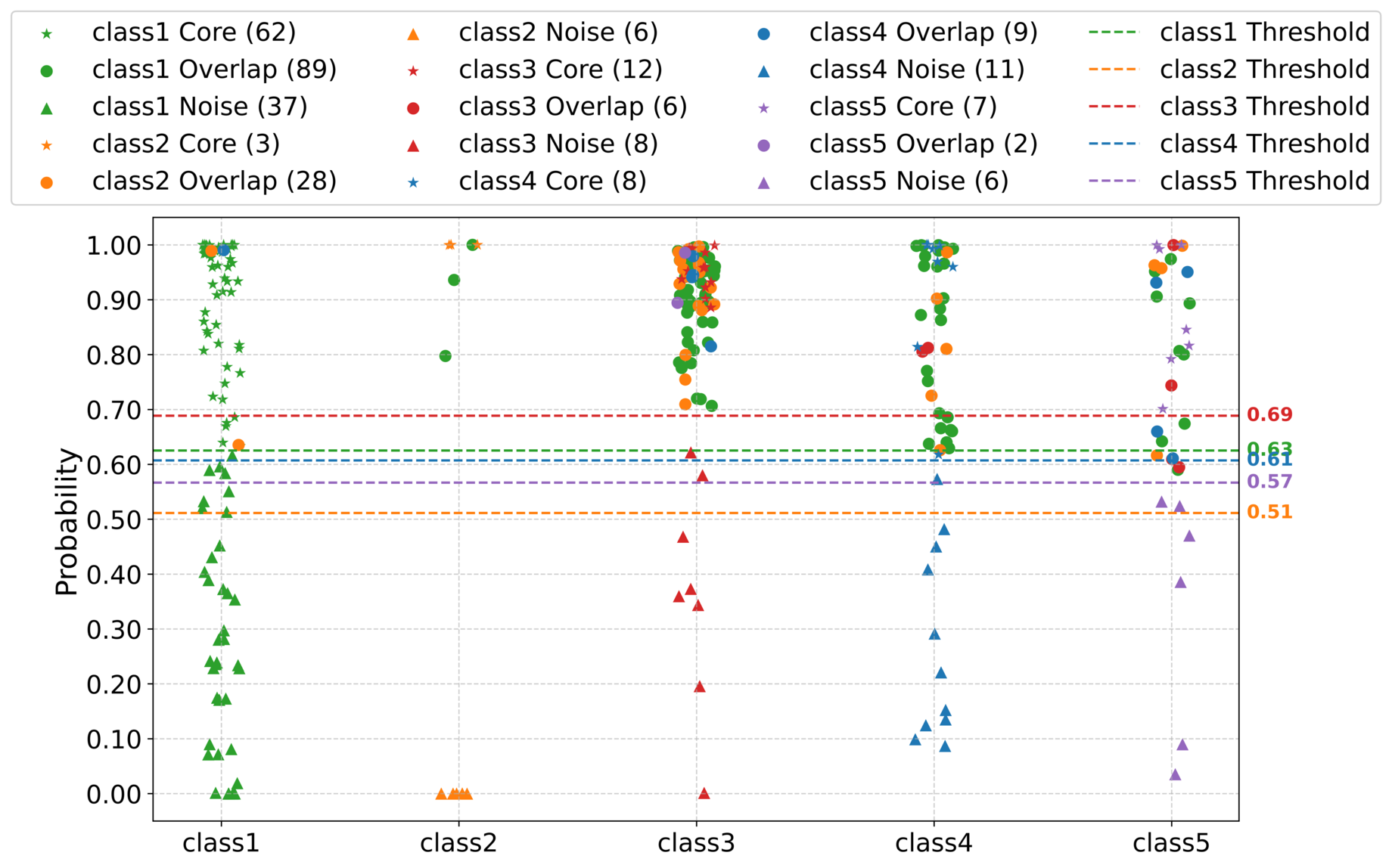} }}
    \subfloat[\centering Yeast dataset]{{\includegraphics[width=5.9cm,height=5cm]{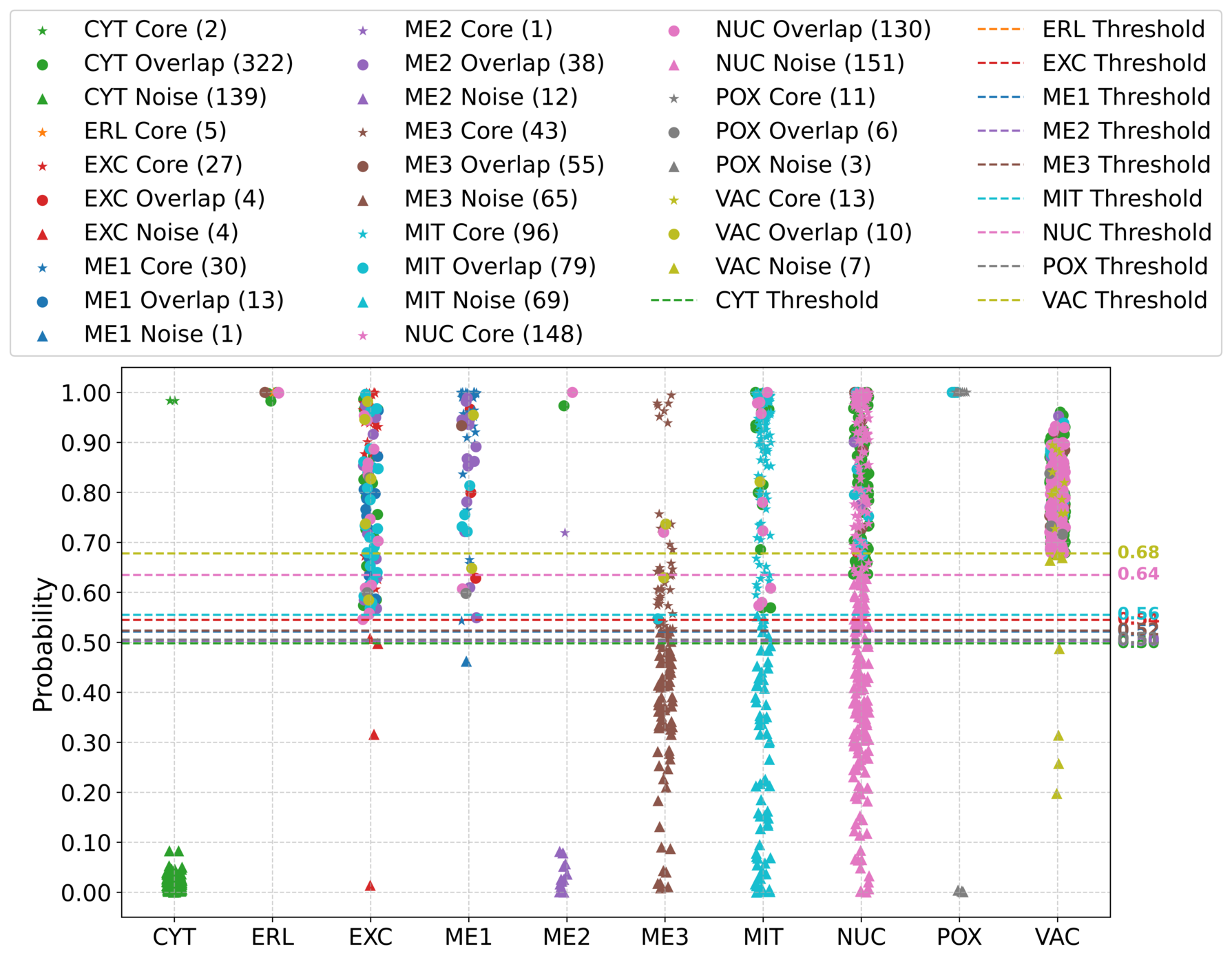} }}
    \subfloat[\centering Winequality-red dataset]{{\includegraphics[width=5.9cm,height=5cm]{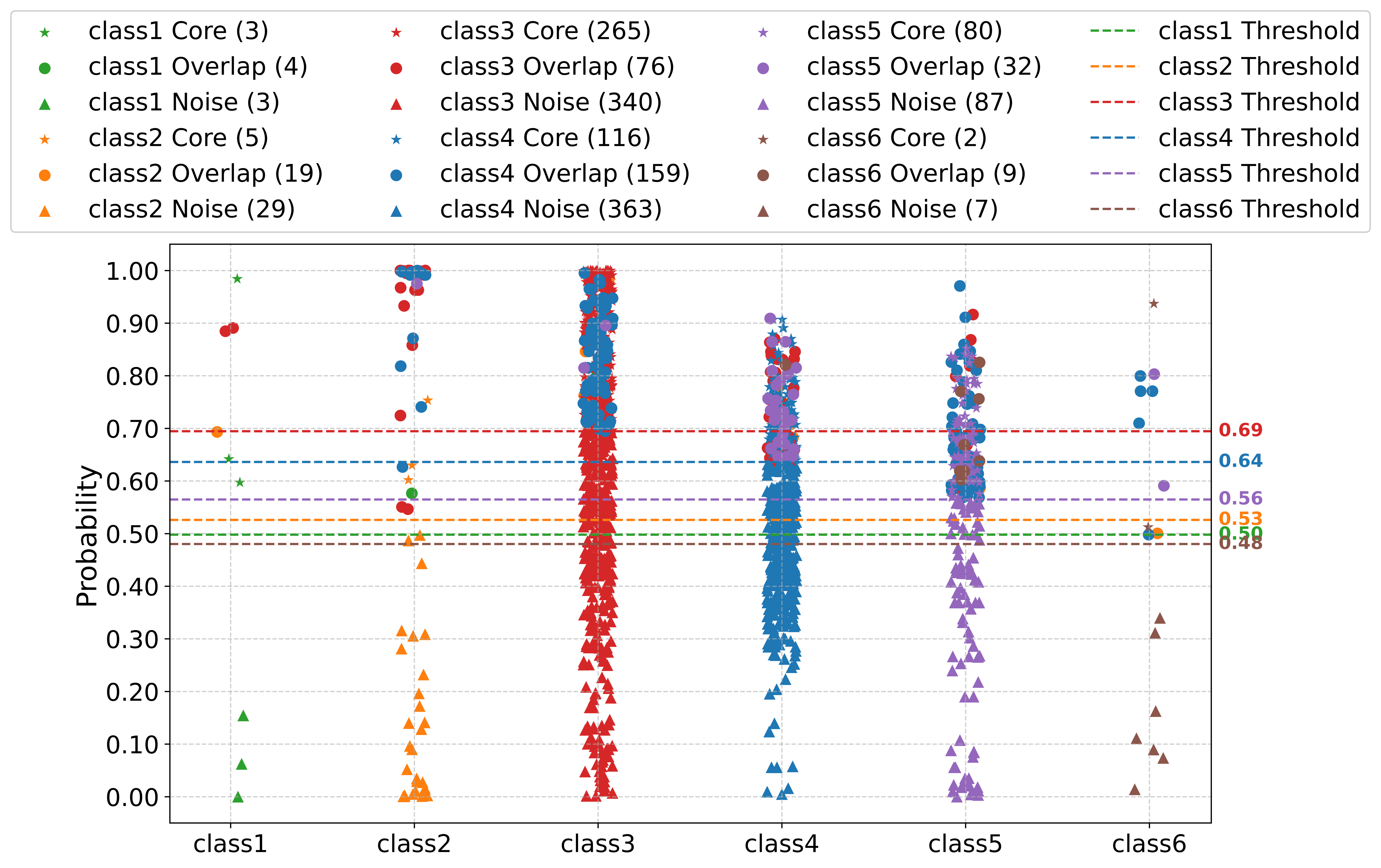} }}

    \caption{Conditional region partition with higher imbalance ratio (IR > 9).}
    \label{fig:Class-Conditional9}
\end{figure*}

An ablation study is conducted to explore two main components of the proposed framework: sample membership probability (Algorithm~\ref{alg:SMP}) and class-conditional region partition (Algorithm~\ref{alg:CCRP}). The purpose is to illustrate how the proposed framework divides the dataset into three main regions: core region, overlapping region, and noisy data region, as shown in Figs.~\ref{fig:Class-Conditional3}--\ref{fig:Class-Conditional9}. The x-axis represents the class names, while the y-axis represents the probability values. To reduce space and avoid presenting too many figures, we split the datasets according to their imbalance ratio (IR). A dataset is considered to have low imbalance if IR $\le$ 3, medium imbalance if 3 $\le$ IR $\le$ 9 and high imbalance if IR > 9 \cite{Experimentation10}. Figs.~\ref{fig:Class-Conditional3}--\ref{fig:Class-Conditional9} illustrate how the probability-based partition splits the data into the three regions based on the defined thresholds, e.g., Eq.~(\ref{eq:Coreregion})--(\ref{eq:Noisyregion}) mentioned in the methodology section.

For instance, Fig.~\ref{fig:Class-Conditional3}(a) illustrates the Contraceptive dataset, which contains three classes. For the first class, the threshold line (green dashed line) identifies 171 samples in the core region (green stars), indicating that these samples have probabilities above the Class 1 threshold and are therefore confidently assigned to their own class. Visually, these core samples do not overlap with points from other classes. Additionally, 217 samples fall within the overlapping region (green circles), meaning that their probability for at least one other class exceeds the respective class threshold, while their probability for their own class remains below its threshold. In the visualization, these overlapping points may appear in the regions corresponding to Class 2 or Class 3. Meanwhile, noisy samples (green triangles, 241 samples) refer to points with very low probabilities, meaning that the probability values for these samples do not exceed the threshold of their own class nor those of any other class. These samples are less confident and contribute minimally to the learning process and are therefore effectively disregarded. Visually, these noisy samples are located below their own class threshold. It is worth noting that some noisy points may appear above the threshold of another class in the plot; this occurs because the visualization is based on the highest probability value for plotting, rather than the minimum value. A similar explanation applies to the other datasets, which have different imbalance ratios.

\subsubsection{Analysis of class overlapping}

\begin{figure*}[htbp]
    \centering
    
    \subfloat{{\includegraphics[width=18cm,height=5cm]{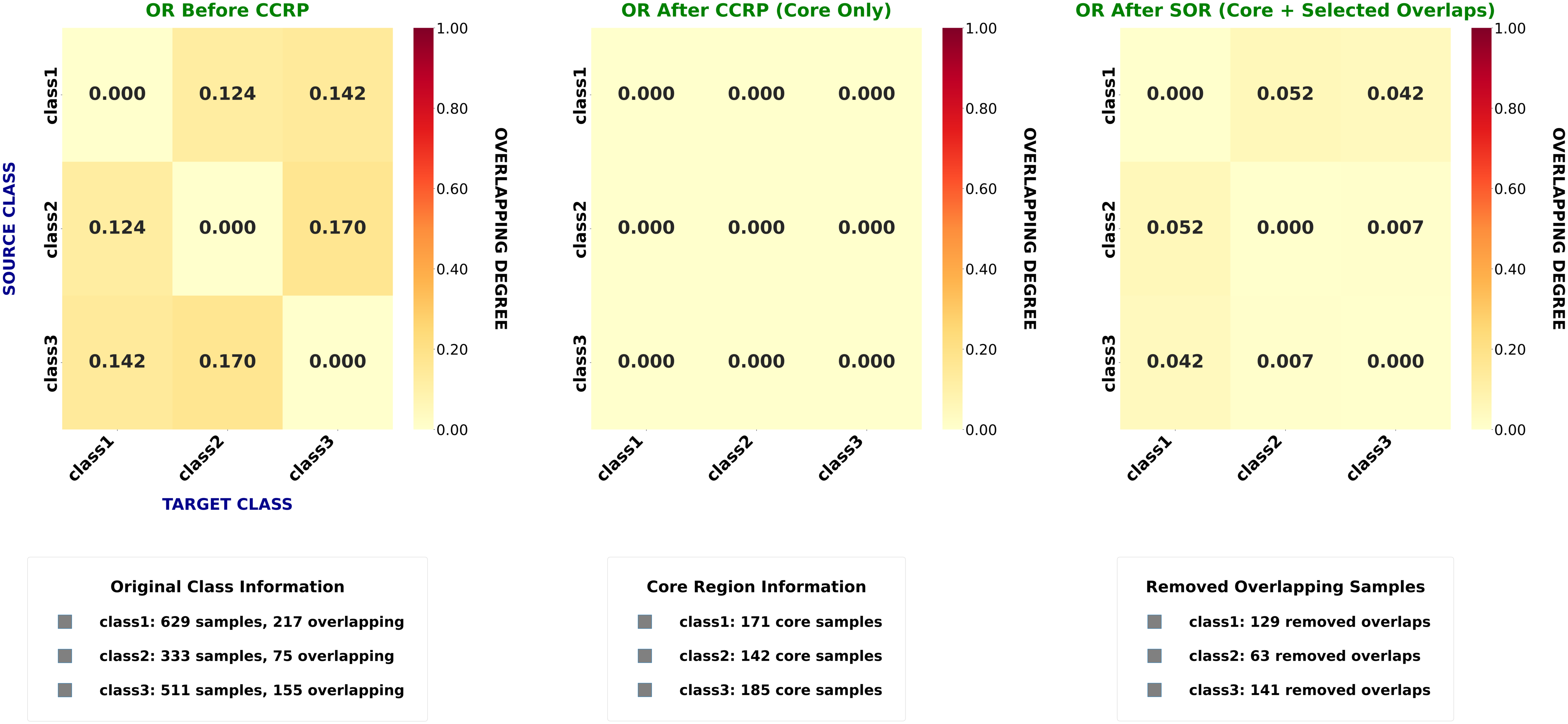} }}

    \caption{Overlapping removal on the Contraceptive dataset (IR $\le$ 3).}
    \label{fig:HeadmapOverlapping3}
\end{figure*}

\begin{figure*}[htbp]
    \centering
    
    \subfloat{{\includegraphics[width=18cm,height=5cm]{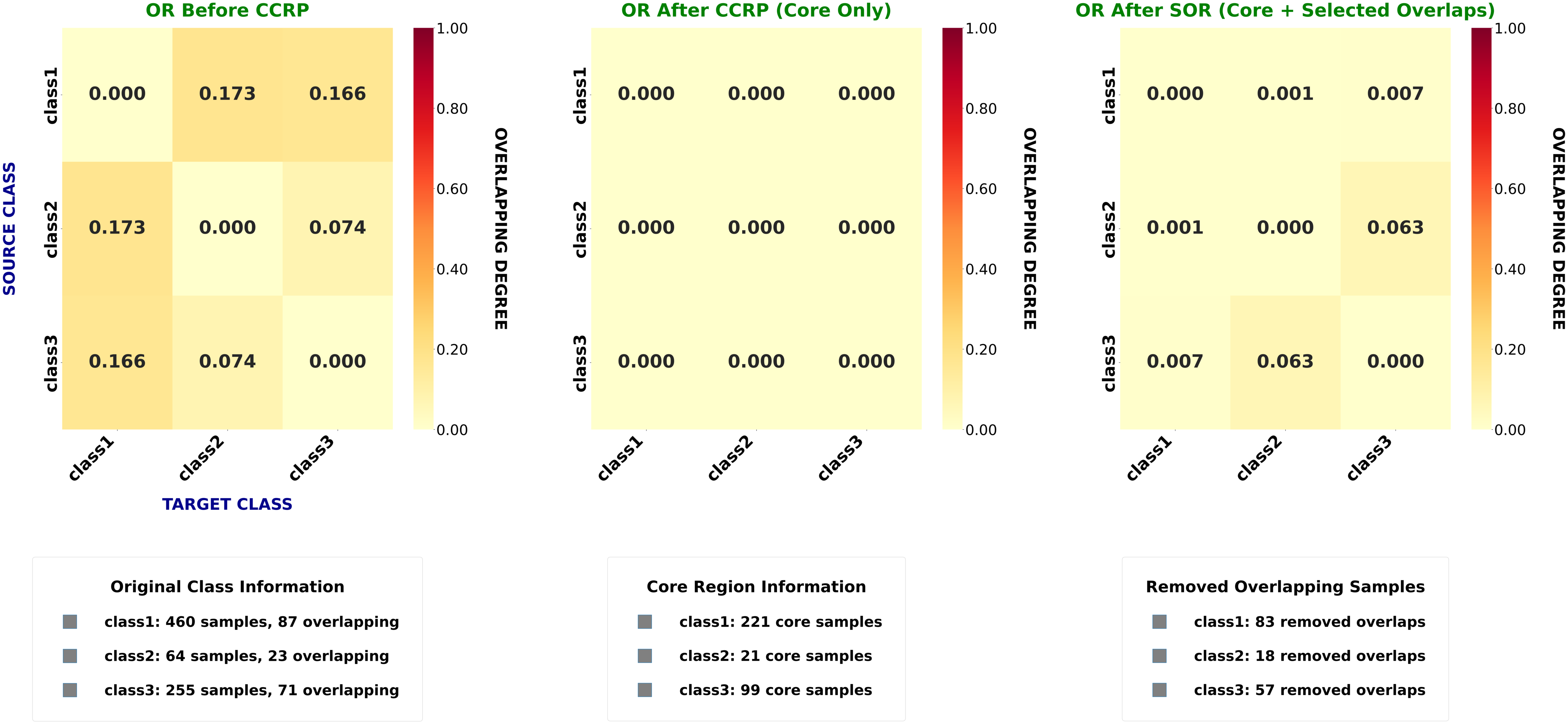} }}

    \caption{Overlapping removal on the Bioconcentration dataset (3 $\le$  IR $\le$  9).}
    \label{fig:HeadmapOverlapping39}
\end{figure*}

\begin{figure*}[htbp]
    \centering
    
    \subfloat{{\includegraphics[width=18cm,height=5cm]{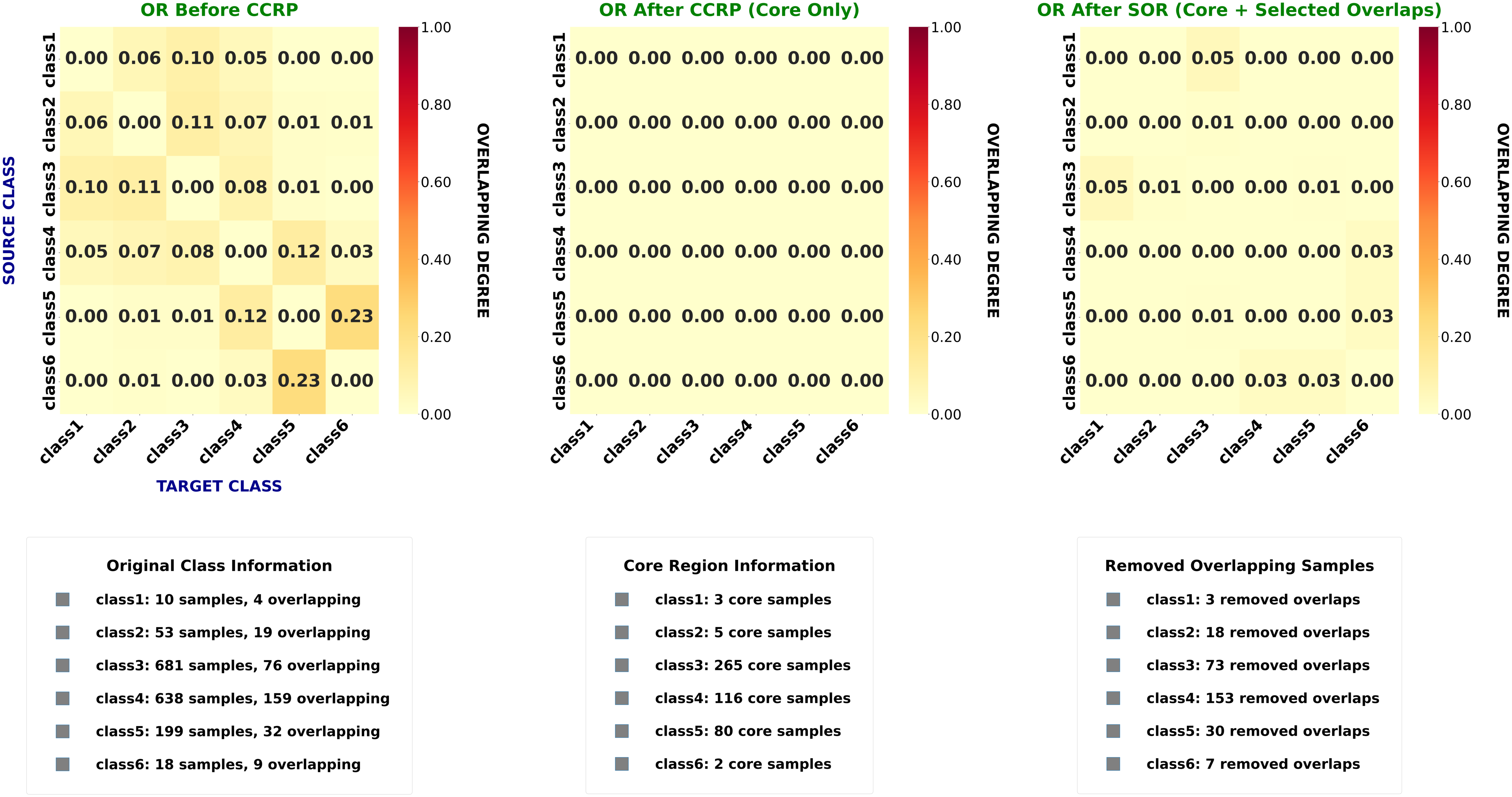} }}

    \caption{Overlapping removal on the Winequality-red dataset (IR > 9).}
    \label{fig:HeadmapOverlapping9}
\end{figure*}

To evaluate the effectiveness of the proposed framework in reducing class overlap in multi-class classification problems, we employ the Overlap Ratio (OR) metric introduced earlier, as reported in Table~\ref{tab:TableOverlappingReduction}, in conjunction with the heatmap visualisation presented in Figs.~\ref{fig:HeadmapOverlapping3}--\ref{fig:HeadmapOverlapping9}. The quantitative results in Table 4 indicate that the original datasets exhibit pronounced overlapping issues before processing, reflecting significant intersections among classes and a lack of clear separability. Such overlap hampers the classifier’s ability to accurately discriminate between samples and increases the complexity of the resulting decision boundaries. After applying the proposed framework (Algorithm~\ref{alg:SOR}), a marked reduction in class overlap is observed. As shown in Table~\ref{tab:TableOverlappingReduction}, the experimental results demonstrate a substantial decrease in class overlap with reductions ranging from 6\% to 60\%. These findings highlight the framework’s strong ability to enhance class separability, improve discriminative power, and effectively mitigate overlapping across a diverse set of multi-class datasets.

Figs.~\ref{fig:HeadmapOverlapping3}--\ref{fig:HeadmapOverlapping9} presents the heatmap visualisation, where darker colours represent a higher degree of overlap between classes. It is evident that after applying the proposed framework, the pixel intensities become noticeably lighter, indicating a reduction in class overlap. In the heatmap, each cell at the intersection of a row and a column corresponds to the overlap value between the respective classes. By comparing these intersection cells before and after applying the framework, it is clear that the overlap values have decreased, further confirming the effectiveness of the proposed framework for minimizing overlap and increasing class separability.

\begin{table}[H]
\centering
\caption{Quantitative evaluation of overlap reduction based on OR.}
\scalebox{0.85}{
\begin{tabular}{l c c c c}
\hline
\textbf{IR Group} & \textbf{Datasets} & \textbf{OR (Before)} & \textbf{OR (After)} & \textbf{Reduction} \\
\hline

\multirow{3}{*}{IR $< 3$}
& Contraceptive & 29.12 & 16.27 & 44.13 \\
& vehicle       & 36.73 & 25.14 & 31.54 \\
& vertebral     & 10.78 & 9.16  & 15.00 \\
\hline

\multirow{3}{*}{$3 < \text{IR} < 9$}
& balance          & 0.46 & 0.39 & 16.28 \\
& bioconcentration & 27.56 & 11.13 & 59.61 \\
& glass            & 26.23 & 24.59 & 6.25 \\
\hline

\multirow{3}{*}{IR $> 9$}
& heart-disease.hungarian & 38.31 & 22.71 & 40.73 \\
& Yeast                   & 34.48 & 20.50 & 40.54 \\
& winequality-red         & 29.67 & 16.69 & 43.74 \\
\hline
\end{tabular}}

\label{tab:TableOverlappingReduction}
\end{table}

\subsubsection{Algorithm visualization}

\begin{figure*}[htbp]
    \centering
    
    \subfloat{{\includegraphics[width=18cm,height=5cm]{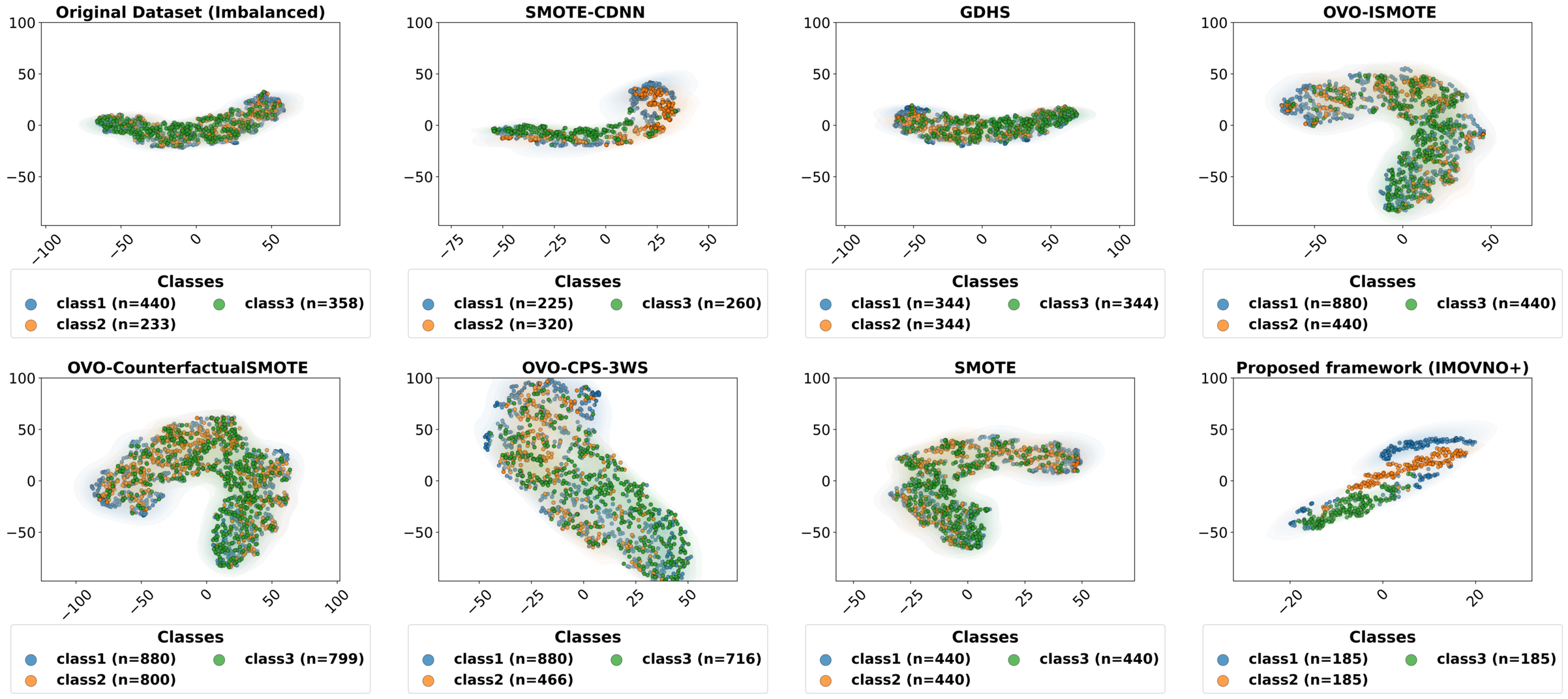} }}

    \caption{Balancing data on Contraceptive dataset(IR  $\le$ 3).}
    \label{fig:AlgorithmVisualisation3}
\end{figure*}

\begin{figure*}[htbp]
    \centering
    
    \subfloat{{\includegraphics[width=18cm,height=5cm]{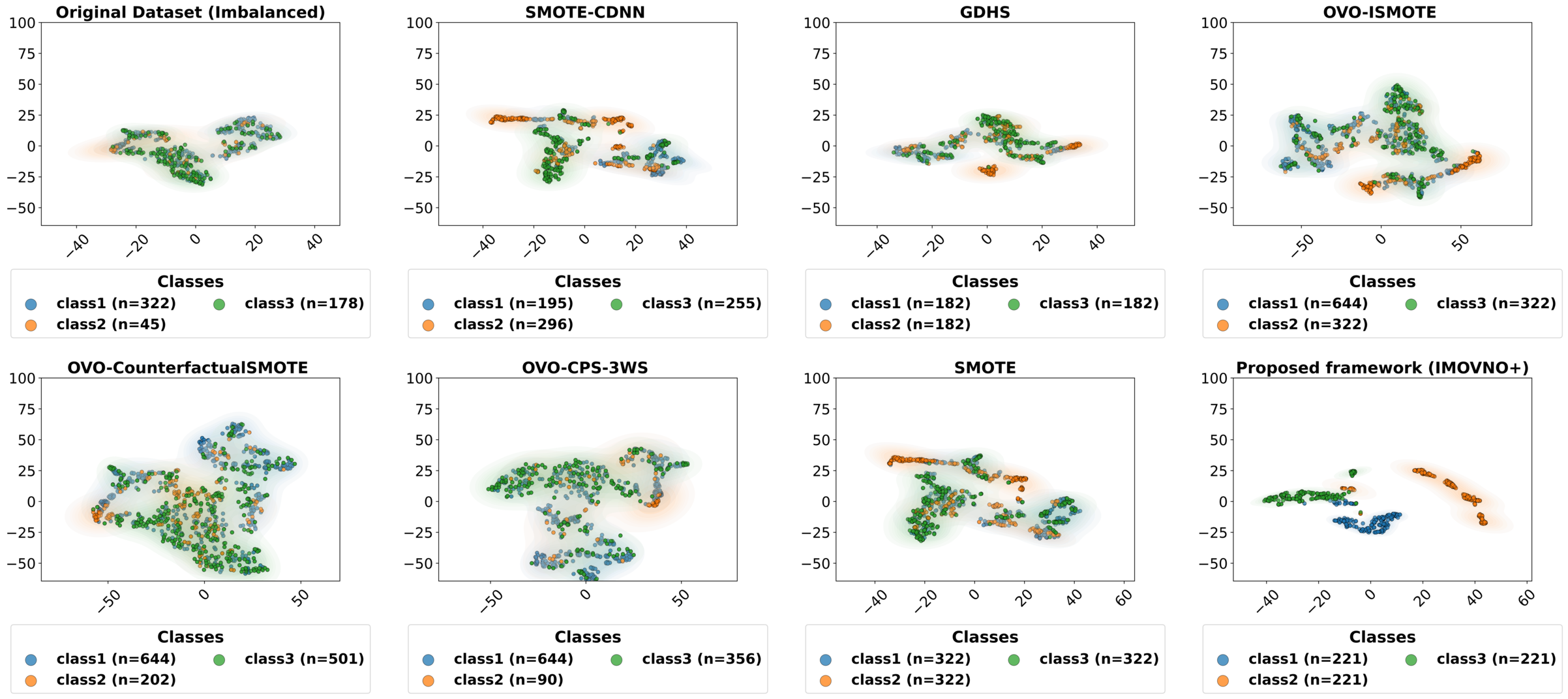} }}

    \caption{Balancing data on Bioconcentration dataset(3 $\le$  IR $\le$  9).}
    \label{fig:AlgorithmVisualisation39}
\end{figure*}

\begin{figure*}[htbp]
    \centering
    
    \subfloat{{\includegraphics[width=18cm,height=5cm]{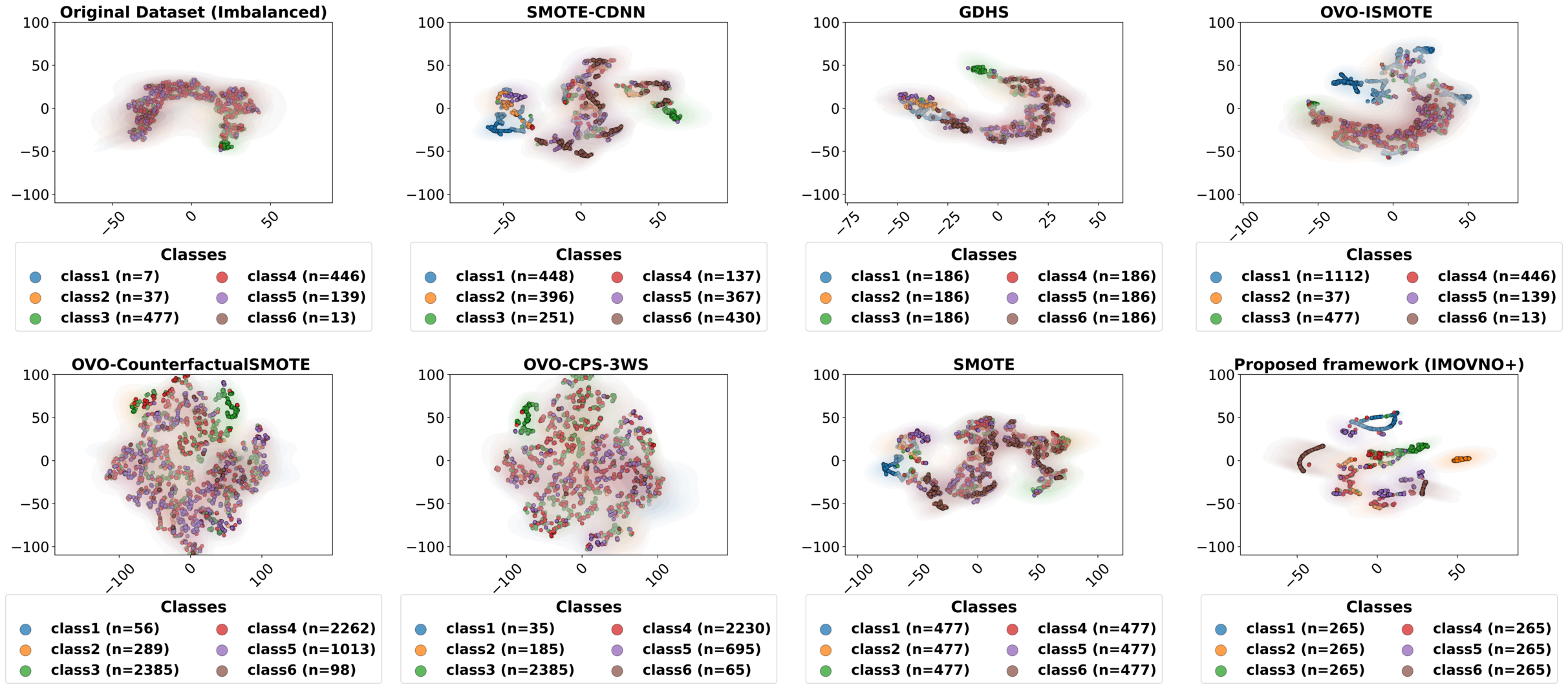} }}

    \caption{Balancing data on winequality-red dataset (IR > 9)}
    \label{fig:AlgorithmVisualisation9}
\end{figure*}

To visualize the working process of the proposed framework in addressing class imbalance and class overlapping compared to existing solutions, a visualization analysis has been conducted, as shown in Figs.~\ref{fig:AlgorithmVisualisation3}--\ref{fig:AlgorithmVisualisation9}. A set of ad hoc methods and OVO decomposition techniques have been applied to different datasets according to their imbalance ratios to illustrate the data distributions produced by these algorithms and the percentage of synthetic samples generated.

From the comparative results in Figs.~\ref{fig:AlgorithmVisualisation3}--\ref{fig:AlgorithmVisualisation9}, it can be observed that the proposed framework achieves a competitive distribution. Unlike other methods, the proposed framework generates a consistent and well-structured distribution. The data distribution produced is well separated, confirming the effective minimization of class overlapping between classes and the absence of disjunct anomalies \cite{Experimentation8}. These results are achieved through the proposed regularization penalty, which keeps the synthetic samples close to their own class (minority class) and far from other classes (majority classes), thereby creating a compact distribution.

In contrast, SMOTE-CDNN, although showing some competitive results, generates samples that still intersect, resulting in less clear decision boundaries. Similarly, with the GDHS method, the classes continue to overlap (e.g., orange points located within green points). This occurs due to the absence of a regularization penalty to control the position of newly generated samples, which may be placed within regions dominated by majority class samples. For multi-class decomposition methods, including ISMOTE, Counterfactual SMOTE, and CPS3, the classes appear more crowded. This scenario arises due to the lack of a regularization penalty. Furthermore, these methods are primarily designed for binary classification, treating multi-class problems as subsets of binary problems and thereby overlooking the relationships between classes and overlapping intersections. On the other hand, basic SMOTE still struggles with overlapping issues, as it blindly generates synthetic samples without controlling their positions \cite{Experimentation9}. Unlike other comparison methods, the proposed framework removes overlapping samples before balancing the dataset, leading to the generation of high-quality samples and minimizing overlap. The framework balances classes while maintaining the same proportion of samples for all classes. For instance, for datasets with IR $\le$ 3, all classes are balanced with 185 instances. In contrast, other methods, such as SMOTE (440 samples) and GDHS (344 samples), balance classes directly on the original data, overlapping any prior overlap cleaning. The new class size outcounts the smallest class in the original dataset (244), resulting in the creation of new overlaps between classes and noisy data, new low-quality samples, and increasing the complexity of the decision boundary and learning process.

\subsubsection{Analaysing  framework components}

We perform an ablation study to determine the most effective configuration for the proposed framework. This study explores the contribution of different core components and assesses their efficiency within the full framework. The outcomes of this ablation analysis are illustrated in Table~\ref{tab:AnalayseFrameworkComponents}. The proposed framework consists of several base components, including Class-Conditional Region Partition (CCRP), Sample Overlapping Removal (SOR), CE, and Optimized Ensemble Learning (OEL). Each component is integrated to achieve specific objectives. CCRP is used to divide the data into regions to reduce the search process. SOR is proposed to clean class overlapping. CE (Class Equalization) is an abbreviation for a two-step algorithm for data balancing and applying a multi-regularization penalty. OEL is proposed to prune the final ensemble learning.

Broadly speaking, the full proposed framework, IMOVNO+, achieved superior performance compared to certain combined components in both multi-class and binary classification contexts. For instance, for datasets with IR < 3, the components CCRP+SOR+OEL achieved lower performance than the full framework due to the absence of the balancing phase, where the final ensemble learning is trained on imbalanced data, leading to overfitting toward the majority class. Similarly, the components CCRP+SOR+CE+EL achieved lower performance due to the absence of ensemble pruning, leading to the inclusion of less-contributing classifiers in the final prediction, on which the ensemble learning is built. In addition, we performed a comparison study by replacing the proposed balancing method (Algorithm~\ref{alg:databalacing} and Algorithm~\ref{alg:OMRP}) with traditional OVO decomposition combined with basic SMOTE, e.g., CCRP+SOR+(OVO+basic-SMOTE)+OEL component.

Based on the findings, OVO+basic-SMOTE achieved lower performance across all relevant metrics compared to the full framework. This can be explained from two perspectives. For multi-class scenarios, OVO splits the multi-class problem into binary subproblems, which loses the relationships between classes, leading to degraded performance. For binary classification, SMOTE generates synthetic data within the majority class, which leads to increased overlapping without integrating a multi-regularization penalty. Therefore, from Table~\ref{tab:AnalayseFrameworkComponents}, we can conclude that class imbalance and overlapping are the main issues in the dataset. Meanwhile, an unpruned ensemble can degrade the final performance. Addressing these challenges leads to higher performance and reliability.

\begin{table*}[!t]
\centering
\caption{Component-wise ablation study of the proposed framework.}
\label{tab:AnalayseFrameworkComponents}

\renewcommand{\arraystretch}{1.3} 
\scalebox{0.55}{
\begin{tabular}{l l c c c c c l c c c c c} 
\hline
\multirow{2}{*}{Dataset (multi-class)} & \multirow{2}{*}{Framework components} 
& \multicolumn{5}{c}{\textbf{Multi-class Metrics}} 
& \multirow{2}{*}{Dataset (binary)} & \multicolumn{5}{c}{\textbf{Binary-class Metrics}} \\ 
\cline{3-7} \cline{9-13}
& & Acc & Prec & Recall & G-mean & F-score & & Acc & Prec & Recall & G-mean & F-score \\ 
\hline
\multicolumn{13}{l}{\textbf{IR < 3}} \\

\multirow{4}{*}{Contraceptive} 
& CCRP+SOR+OEL 
& 80.23 ± 0.0092 & 81.70 ± 0.0091 & 79.80 ± 0.0104 & 79.56 ± 0.0103 & 80.34 ± 0.0094 & 
\multirow{4}{*}{yeast1} & 94.58 ± 0.0042 & 97.32 ± 0.0041 & 94.23 ± 0.0050 & 95.72 ± 0.0034 & 94.69 ± 0.0046 \\

& CCRP+SOR+(OVO+basic-smote)+OEL
& 54.56 ± 0.0087 & 54.38 ± 0.0094 & 53.08 ± 0.0086 & 50.66 ± 0.0092 & 52.29 ± 0.0088 & 
& 65.81 ± 0.0104 & 50.52 ± 0.0155 & 65.54 ± 0.0421 & 64.60 ± 0.0160 & 55.94 ± 0.0202 \\

& CCRP+SOR+CE+EL
& 84.85 ± 0.0499 & 84.23 ± 0.0857 & 84.85 ± 0.0499 & 73.93 ± 0.1600 & 82.33 ± 0.0697 & 
& 90.77 ± 0.0123 & 86.39 ± 0.0200 & 97.71 ± 0.0093 & 90.35 ± 0.0137 & 91.52 ± 0.0101 \\

& CCRP+SOR+CE+OEL (IMOVNO+)
& 95.26 ± 0.0083 & 95.58 ± 0.0066 & 95.26 ± 0.0083 & 95.10 ± 0.0091 & 95.23 ± 0.0084 & 
& 95.06 ± 0.0027 & 98.12 ± 0.0031 & 91.91 ± 0.0064 & 94.99 ± 0.0028 & 94.88 ± 0.0029 \\

\cline{2-13}

\multirow{4}{*}{Vehicle} 
& CCRP+SOR+OEL 
& 75.04 ± 0.0112 & 76.70 ± 0.0253 & 68.87 ± 0.0166 & 63.15 ± 0.0274 & 69.45 ± 0.0197 & 
\multirow{4}{*}{Pima} & 81.76 ± 0.0054 & 73.58 ± 0.0132 & 66.80 ± 0.0108 & 69.69 ± 0.0089 & 76.77 ± 0.0067 \\

& CCRP+SOR+(OVO+basic-smote)+OEL
& 56.34 ± 0.0108 & 60.17 ± 0.0154 & 56.59 ± 0.0108 & 50.21 ± 0.0149 & 53.94 ± 0.0124 & 
& 79.75 ± 0.0039 & 72.31 ± 0.0169 & 58.84 ± 0.0171 & 72.26 ± 0.0069 & 64.48 ± 0.0071 \\

& CCRP+SOR+CE+EL
& 73.39 ± 0.0117 & 75.76 ± 0.0464 & 73.39 ± 0.0118 & 46.37 ± 0.1155 & 68.28 ± 0.0188 & 
& 95.17 ± 0.0036 & 98.24 ± 0.0030 & 92.02 ± 0.0069 & 95.10 ± 0.0037 & 94.98 ± 0.0038 \\

& CCRP+SOR+CE+OEL (IMOVNO+)
& 79.42 ± 0.0059 & 82.84 ± 0.0100 & 79.41 ± 0.0060 & 70.59 ± 0.0105 & 77.00 ± 0.0068 & 
& 96.21 ± 0.0041 & 97.42 ± 0.0062 & 95.02 ± 0.0094 & 96.17 ± 0.0042 & 96.15 ± 0.0044 \\

\cline{2-13}

\multirow{4}{*}{Vertebral} 
& CCRP+SOR+OEL 
& 90.38 ± 0.0042 & 86.31 ± 0.0052 & 87.86 ± 0.0074 & 87.20 ± 0.0083 & 86.48 ± 0.0062 & 
\multirow{4}{*}{biodeg} & 79.43 ± 0.0098 & 65.60 ± 0.0128 & 83.78 ± 0.0225 & 73.32 ± 0.0139 & 80.23 ± 0.0121 \\

& CCRP+SOR+(OVO+basic-smote)+OEL
& 89.89 ± 0.0064 & 86.90 ± 0.0095 & 88.21 ± 0.0099 & 87.53 ± 0.0103 & 86.76 ± 0.0085 & 
& 84.25 ± 0.0024 & 77.46 ± 0.0049 & 75.84 ± 0.0063 & 81.89 ± 0.0024 & 76.51 ± 0.0028 \\

& CCRP+SOR+CE+EL
& 78.35 ± 0.0363 & 80.42 ± 0.0562 & 78.40 ± 0.0362 & 56.77 ± 0.1123 & 73.45 ± 0.0509 & 
& 94.71 ± 0.0049 & 93.90 ± 0.0189 & 95.89 ± 0.0180 & 94.64 ± 0.0050 & 94.77 ± 0.0047 \\

& CCRP+SOR+CE+OEL (IMOVNO+)
& 91.62 ± 0.0035 & 92.58 ± 0.0036 & 91.63 ± 0.0034 & 91.08 ± 0.0036 & 91.53 ± 0.0035 & 
& 95.35 ± 0.0034 & 97.89 ± 0.0015 & 92.70 ± 0.0067 & 95.30 ± 0.0035 & 95.20 ± 0.0037 \\

\hline
\multicolumn{13}{l}{\textbf{ 3 $\leq$ IR $\leq$ 9  }} \\
\multirow{4}{*}{Balance} 
& CCRP+SOR+OEL 
& 98.85 ± 0.0000 & 88.78 ± 0.0174 & 99.23 ± 0.0001 & 99.22 ± 0.0001 & 91.81 ± 0.0092 
& \multirow{4}{*}{Ecoli1} & 86.84 ± 0.0047 & 66.16 ± 0.0107 & 91.78 ± 0.0219 & 76.60 ± 0.0098 & 88.39 ± 0.0097 \\

& CCRP+SOR+(OVO+basic-smote)+OEL
& 98.43 ± 0.0014 & 84.74 ± 0.0323 & 95.70 ± 0.0333 & 89.06 ± 0.0996 & 87.59 ± 0.0312
& & 87.03 ± 0.0038 & 65.74 ± 0.0048 & 95.88 ± 0.0171 & 89.86 ± 0.0078 & 77.77 ± 0.0078 \\

& CCRP+SOR+CE+EL
& 86.14 ± 0.0276 & 89.80 ± 0.0302 & 86.16 ± 0.0277 & 82.85 ± 0.0545 & 85.44 ± 0.0325
& & 99.20 ± 0.0029 & 99.27 ± 0.0039 & 99.15 ± 0.0032 & 99.20 ± 0.0029 & 99.20 ± 0.0029 \\

& CCRP+SOR+CE+OEL (IMOVNO+)
& 99.64 ± 0.0007 & 99.65 ± 0.0007 & 99.64 ± 0.0007 & 99.64 ± 0.0007 & 99.64 ± 0.0007
& & 99.18 ± 0.0042 & 98.61 ± 0.0068 & 99.80 ± 0.0024 & 99.17 ± 0.0042 & 99.19 ± 0.0041 \\

\cline{2-13}
\multirow{4}{*}{Bioconcentration} 
& CCRP+SOR+OEL 
& 92.69 ± 0.0059 & 90.92 ± 0.0088 & 84.94 ± 0.0063 & 83.17 ± 0.0049 & 86.80 ± 0.0044 
& \multirow{4}{*}{Segemet0} & 98.73 ± 0.0013 & 93.02 ± 0.0076 & 98.87 ± 0.0036 & 95.82 ± 0.0040 & 98.79 ± 0.0018 \\

& CCRP+SOR+(OVO+basic-smote)+OEL
& 69.27 ± 0.0065 & 71.46 ± 0.0235 & 54.94 ± 0.0078 & 46.74 ± 0.0384 & 58.19 ± 0.0109
& & 99.08 ± 0.0019 & 96.66 ± 0.0101 & 97.19 ± 0.0045 & 98.29 ± 0.0025 & 96.89 ± 0.0060 \\

& CCRP+SOR+CE+EL
& 86.17 ± 0.0632 & 86.92 ± 0.0701 & 86.16 ± 0.0633 & 71.99 ± 0.1591 & 83.20 ± 0.0821
& & 96.23 ± 0.0076 & 93.65 ± 0.0127 & 99.29 ± 0.0093 & 96.16 ± 0.0076 & 96.36 ± 0.0073 \\

& CCRP+SOR+CE+OEL (IMOVNO+)
& 96.53 ± 0.0021 & 96.77 ± 0.0020 & 96.53 ± 0.0021 & 96.41 ± 0.0022 & 96.51 ± 0.0021
& & 99.49 ± 0.0003 & 99.21 ± 0.0007 & 99.77 ± 0.0002 & 99.49 ± 0.0003 & 99.49 ± 0.0003 \\

\cline{2-13}

\multirow{4}{*}{glass} 
& CCRP+SOR+OEL 
& 71.37 ± 0.0238 & 59.36 ± 0.0610 & 57.08 ± 0.0350 & 04.10 ± 0.0628 & 55.91 ± 0.0421 
& \multirow{4}{*}{Ecoli3} & 85.11 ± 0.0140 & 42.37 ± 0.0184 & 90.57 ± 0.0287 & 57.17 ± 0.0163 & 87.10 ± 0.0133 \\

& CCRP+SOR+(OVO+basic-smote)+OEL
& 56.73 ± 0.0203 & 46.46 ± 0.0420 & 46.41 ± 0.0267 & 09.77 ± 0.0239 & 43.92 ± 0.0325
& & 91.83 ± 0.0048 & 65.93 ± 0.0389 & 55.43 ± 0.0408 & 71.75 ± 0.0294 & 57.98 ± 0.0343 \\

& CCRP+SOR+CE+EL
& 66.21 ± 0.0863 & 67.79 ± 0.0930 & 66.79 ± 0.0859 & 40.40 ± 0.1651 & 62.84 ± 0.0954
& & 97.61 ± 0.0026 & 97.55 ± 0.0082 & 97.74 ± 0.0053 & 97.60 ± 0.0027 & 97.61 ± 0.0025 \\

& CCRP+SOR+CE+OEL (IMOVNO+)
& 91.74 ± 0.0157 & 93.35 ± 0.0145 & 91.85 ± 0.0155 & 90.06 ± 0.0395 & 91.68 ± 0.0164
& & 98.23 ± 0.0036 & 98.96 ± 0.0028 & 97.49 ± 0.0076 & 98.22 ± 0.0037 & 98.20 ± 0.0038 \\

\hline

\multicolumn{13}{l}{\textbf{ IR > 9  }} \\
\multirow{4}{*}{Hungarian} 
& CCRP+SOR+OEL 
& 72.18 ± 0.0111 & 28.82 ± 0.0252 & 35.08 ± 0.0143 & 02.62 ± 0.0041 & 30.79 ± 0.0174 
& \multirow{4}{*}{$yeast-1\_vs\_7$} & 76.99 ± 0.0383 & 19.85 ± 0.0406 & 65.00 ± 0.0492 & 28.81 ± 0.0336 & 63.61 ± 0.0406 \\

& CCRP+SOR+(OVO+basic-smote)+OEL
& 65.26 ± 0.0103 & 18.52 ± 0.0304 & 22.19 ± 0.0139 & 01.21 ± 0.0071 & 19.53 ± 0.0176
& & 93.36 ± 0.0046 & 58.70 ± 0.1446 & 24.50 ± 0.0693 & 41.51 ± 0.1025 & 32.77 ± 0.0828 \\

& CCRP+SOR+CE+EL
& 80.73 ± 0.0733 & 84.32 ± 0.0720 & 80.82 ± 0.0729 & 66.37 ± 0.1584 & 78.76 ± 0.0837
& & 96.55 ± 0.0078 & 94.74 ± 0.0139 & 98.76 ± 0.0101 & 96.49 ± 0.0079 & 96.65 ± 0.0077 \\

& CCRP+SOR+CE+OEL (IMOVNO+)
& 93.21 ± 0.0066 & 93.81 ± 0.0057 & 93.21 ± 0.0066 & 92.67 ± 0.0077 & 93.08 ± 0.0069
& & 96.98 ± 0.0052 & 95.05 ± 0.0099 & 99.26 ± 0.0078 & 96.94 ± 0.0054 & 97.07 ± 0.0049 \\

\cline{2-13}
\multirow{4}{*}{Yeast} 
& CCRP+SOR+OEL 
& 69.92 ± 0.0186 & 30.50 ± 0.0291 & 31.08 ± 0.0214 & 02.24 ± 0.0053 & 29.47 ± 0.0233 
& \multirow{4}{*}{Ecoli4} & 98.20 ± 0.0053 & 91.23 ± 0.0484 & 82.50 ± 0.0335 & 85.01 ± 0.0369 & 89.98 ± 0.0203 \\

& CCRP+SOR+(OVO+basic-smote)+OEL
& 61.17 ± 0.0067 & 59.67 ± 0.0165 & 60.50 ± 0.0063 & 31.33 ± 0.0198 & 57.15 ± 0.0086
& & 98.35 ± 0.0039 & 94.73 ± 0.0449 & 80.50 ± 0.0269 & 88.84 ± 0.0157 & 85.26 ± 0.0292 \\

& CCRP+SOR+CE+EL
& 82.03 ± 0.0502 & 84.78 ± 0.0529 & 82.11 ± 0.0499 & 66.17 ± 0.1277 & 80.33 ± 0.0581
& & 98.61 ± 0.0077 & 99.68 ± 0.0019 & 97.55 ± 0.0153 & 98.58 ± 0.0079 & 98.56 ± 0.0081 \\

& CCRP+SOR+CE+OEL (IMOVNO+)
& 91.77 ± 0.0123 & 93.20 ± 0.0093 & 91.81 ± 0.0121 & 90.62 ± 0.0154 & 91.53 ± 0.0128
& & 99.93 ± 0.0016 & 99.90 ± 0.0022 & 99.96 ± 0.0011 & 99.93 ± 0.0016 & 99.93 ± 0.0016 \\

\cline{2-13}
\multirow{4}{*}{Winequality-red} 
& CCRP+SOR+OEL 
& 88.12 ± 0.0049 & 50.63 ± 0.0185 & 49.55 ± 0.0193 & 05.78 ± 0.0122 & 49.54 ± 0.0173
& \multirow{4}{*}{$ecoli0137\_vs\_26$} & 98.40 ± 0.0029 & 62.67 ± 0.0611 & 69.00 ± 0.0300 & 62.40 ± 0.0408 & 72.37 ± 0.0419 \\

& CCRP+SOR+(OVO+basic-smote)+OEL
& 61.45 ± 0.0046 & 27.98 ± 0.0264 & 27.85 ± 0.0103 & 02.19 ± 0.0057 & 26.94 ± 0.0142
& & 98.35 ± 0.0039 & 94.73 ± 0.0449 & 80.50 ± 0.0269 & 88.84 ± 0.0157 & 85.26 ± 0.0292 \\

& CCRP+SOR+CE+EL
& 57.11 ± 0.0605 & 70.64 ± 0.0770 & 57.11 ± 0.0605 & 33.07 ± 0.1073 & 52.60 ± 0.0743
& & 98.70 ± 0.0102 & 99.71 ± 0.0021 & 97.69 ± 0.0204 & 98.67 ± 0.0105 & 98.64 ± 0.0109 \\

& CCRP+SOR+CE+OEL (IMOVNO+)
& 95.24 ± 0.0029 & 95.62 ± 0.0026 & 95.24 ± 0.0029 & 95.02 ± 0.0032 & 95.21 ± 0.0030
& & 99.80 ± 0.0010 & 99.62 ± 0.0019 & 100.00 ± 0.0000 & 99.80 ± 0.0010 & 99.81 ± 0.0009 \\

\hline
\end{tabular}}
\end{table*}

\subsection{Comparative  study}

In this section, a comparative study of the proposed framework, IMOVNO+, with recent state-of-the-art methods is presented in detail. In these experiments, performance values were obtained using the mean score of a stratified five-fold cross-validation repeated ten times. To facilitate the experimental comparison, this section is organized according to the main objectives of the study. First, a comparative analysis with ad hoc multi-class imbalance methods is conducted to demonstrate the superiority of the proposed framework over recent approaches. Second, a comparison against OVO binary decomposition methods is presented. Third, an additional comparison is carried out under binary classification scenarios to illustrate the effectiveness of the proposed approach in addressing class imbalance and class overlapping in binary settings. Tables~\ref{tab:multiClassTable1}--\ref{tab:binary_Table2} present the results, with bold values indicating the best performance and the last row showing the average results across all datasets.

\subsubsection{Comparative analysis for multiclass classification}

Table~\ref{tab:multiClassTable1} shows the comparative results of the proposed framework with state-of-the-art multiclass imbalanced data methods based on SVM and decision tree classifiers across key performance metrics. The proposed framework consistently outperforms other methods across various datasets. For instance, it achieves the highest accuracy (100.00\%), highest G-Mean (100.00\%), and highest F1-score (100.00\%) on the Zoo dataset, while the Balance and New-Thyroid datasets achieve approximately the same values, with corresponding results of accuracy (99.64 ± 0.0007), G-Mean (99.64 ± 0.0007), F1-score (99.64 ± 0.0007), and accuracy (99.50 ± 0.0026), G-Mean (99.49 ± 0.0027), and F1-score (99.50 ± 0.0027), respectively, highlighting its effectiveness on multiple imbalanced multiclass datasets.

Furthermore, when comparing our proposed framework , IMOVNO+,  with the best-performing methods among those used, our framework clearly outperformed it in all performance metrics. Specifically: \\

- At a low imbalance ratio (Contraceptive, IR = 1.89), it achieved an improvement of 41.30 ± 0.043\% in G-mean, 41.33 ± 0.042\% in F-score, and 40.59 ± 0.22\% in accuracy compared to the best-performing models, DT-E-EVRS and CPSQSVM.\\

- At a low imbalance ratio (Contraceptive, IR = 1.89), it achieved an improvement of 3.19 ± 0.036\% in G-mean, 3.90 ± 0.030\% in F-score, and 3.41 ± 0.37\% in accuracy compared to the best-performing models, SVM-E-EVRS and CPSQSVM.\\

- At a medium imbalance ratio (New-Thyroid, IR = 1.89), it achieved an improvement of 3.19 ± 0.036\% in G-mean, 3.90 ± 0.030\% in F-score, and 3.41 ± 0.37\% in accuracy compared to the best-performing models, SVM-E-EVRS and CPSQSVM.\\

- At a higher imbalance ratio (Yeast, IR = 92.6), it achieved an improvement of 20.22 ± 0.042\% in G-mean, 31.73 ± 0.062\% in F-score, and 33.62 ± 0.33\% in accuracy compared to the best-performing models, DT-E-EVRS, SVM-E-EVRS, and CPSQSVM.

\begin{table*}[!t]
\centering
\caption{Comparison analysis with published state-of-the-art results on multi-class imbalanced datasets using 5-fold cross-Validation.}
\label{tab:multiClassTable1}
\renewcommand{\arraystretch}{1.3}
\scalebox{0.70}{%
\begin{tabular}{l
                r r
                r r
                r r
                r r r
                }
\toprule
\multirow{2}{*}{Dataset} 
& \multicolumn{2}{c}{DT-E-EVRS (2023) ) \cite{Introdcution18}} 
& \multicolumn{2}{c}{SVM-E-EVRS (2023) \cite{Introdcution18}} 
& \multicolumn{2}{c}{CPSQSVM (2025)  \cite{Experimentation12}} 
& \multicolumn{3}{c}{IMOVNO+ (Proposed)} 
\\
\cmidrule(lr){2-3} \cmidrule(lr){4-5} \cmidrule(lr){6-7} \cmidrule(lr){8-10}
& $F_1$-score($\%$) &G-mean($\%$)  & $F_1$-score($\%$) & G-mean($\%$) & Accuracy $\%$ & $F_1$-score($\%$) &Accuracy$\%$& $F_1$-score($\%$) & G-mean($\%$) \\
\midrule

Contraceptive 
& 53.90$\pm$0.041 & 53.80$\pm$0.042
& 53.90$\pm$0.033 & 54.50$\pm$0.033
& 54.67$\pm$0.22 & 53.71$\pm$0.20
& 95.26$\pm$0.0083 & 95.23$\pm$0.0084 & 95.10$\pm$0.0091 \\

Balance 
& 84.90$\pm$0.085 & 89.70$\pm$0.034
& 81.20$\pm$0.061 & 88.50$\pm$0.066
& 90.59$\pm$0.67 & 71.44$\pm$0.91
&  \textbf{99.64$\pm$0.0007} & \textbf{ 99.64$\pm$0.0007} &  \textbf{99.64$\pm$0.0007} \\

New-thyroid 
& 94.90$\pm$0.053 & 95.20$\pm$0.060
& 95.60$\pm$0.030 & 96.30$\pm$0.036
& 96.09$\pm$0.37 & 93.81$\pm$0.76
&  \textbf{99.50$\pm$0.0026 }&  \textbf{99.50$\pm$0.0027 }&  \textbf{99.49$\pm$0.0027 }\\

Zoo 
& 92.80$\pm$0.110 & 70.00$\pm$0.458
& 91.80$\pm$0.136 & 78.90$\pm$0.445
& 96.69$\pm$0.91 & 89.11$\pm$2.71
&  \textbf{100.00$\pm$0.0000 }& \textbf{ 100.00$\pm$0.0000 }&  \textbf{100.00$\pm$0.0000} \\

Yeast 
& 56.30$\pm$0.151 & 70.40$\pm$0.039
& 59.80$\pm$0.061 & 59.20$\pm$0.223
& 58.15$\pm$0.33 & 52.24$\pm$0.97
& 91.77$\pm$0.0123 & 91.53$\pm$0.0128 & 90.62$\pm$0.0154 \\

Vehicle 
& 77.10$\pm$0.034 & 74.50$\pm$0.051
& 77.90$\pm$0.032 & 84.50$\pm$0.040
& 83.71$\pm$0.35 & 83.70$\pm$0.34
& 79.42$\pm$0.0059 & 77.00$\pm$0.0068 & 70.59$\pm$0.0105 \\

Dermatology 
& 97.90$\pm$0.027 & 97.60$\pm$0.029
& 95.20$\pm$0.031 & 95.40$\pm$0.047
& - & -
& 98.54$\pm$0.0012 & 98.54$\pm$0.0012 & 98.50$\pm$0.0012 \\

Pageblocks 
& 70.00$\pm$0.137 & 55.80$\pm$0.447
& 70.20$\pm$0.183 & 59.20$\pm$0.466
& - & -
& 97.34$\pm$0.0026 & 97.34$\pm$0.0026 & 97.22$\pm$0.0029 \\

\bottomrule
\end{tabular}%
} 
\end{table*}

Table~\ref{tab:multiClassTable2} The proposed framework was compared with recent methods for handling multiclass imbalanced data. It consistently outperformed other approaches across all relevant metrics, except in one case: for G-Mean and F-score on the Vehicle dataset, where it ranked second after the MDGP-Forest model. However, it achieved optimal performance on a balanced dataset across four key metrics, with the following results: G-Mean (99.64 ± 0.0007\%), F-score (99.64 ± 0.0007\%), recall (99.64 ± 0.0007\%), and precision (99.65 ± 0.0007\%). In contrast, the GDHS method obtained the lowest values, with an F-score of 57.67 ± 0.0370\% and a precision of 58.19 ± 0.0368\%. The average performance reported in the last row illustrates that the proposed framework surpasses the second-best performing model (MDGP-Forest) in G-Mean and F-score by 12.82 ± 0.0401\% and 23.47 ± 0.053\%, respectively.

\begin{table*}[!t]
\centering
\caption{Comparison analysis with published state-of-the-art results on multi-class imbalanced datasets using 5-Fold Cross-Validation.}
\label{tab:multiClassTable2}
\renewcommand{\arraystretch}{1.3}
\scalebox{0.60}{%
\begin{tabular}{l
                r r
                r r r r
                r r r
                r r r r
                }
\toprule
\multirow{2}{*}{Dataset} 
& \multicolumn{2}{c}{MDGP-Forest (2025) \cite{Experimentation13}} 
& \multicolumn{4}{c}{GDHS (2025) \cite{Experimentation14}} 
& \multicolumn{3}{c}{FEHC (2024) \cite{Experimentation15}} 
& \multicolumn{4}{c}{IMOVNO+ (Proposed)} 
\\
\cmidrule(lr){2-3} \cmidrule(lr){4-7} \cmidrule(lr){8-10} \cmidrule(lr){11-14}
& $F_1$-score($\%$) &G-mean($\%$)
& Precision($\%$)& Recall($\%$) & $F_1$-score($\%$) & G-mean($\%$) 
& Precision($\%$)& Recall($\%$) & $F_1$-score($\%$)
&Precision($\%$) & Recall($\%$)& $F_1$-score($\%$) & G-mean($\%$)
\\
\midrule

Vehicle 
& 81.03$\pm$0.0261 & 87.32$\pm$0.0181
& 68.93$\pm$0.0219 & 69.12$\pm$0.0246 & 68.85$\pm$0.0227 & 65.76$\pm$0.0305
& 73.24$\pm$0.0148 & 74.54$\pm$0.0123 & 73.50$\pm$0.0139
& 82.84$\pm$0.0100 & 79.41$\pm$0.0060 & 77.00$\pm$0.0068 & 70.59$\pm$0.0105
\\

Yeast 
& 58.35$\pm$0.0614 & 73.69$\pm$0.0412
& 45.14$\pm$0.0314 & 51.60$\pm$0.0228 & 45.64$\pm$0.0260 & 38.28$\pm$0.1126
& 47.07$\pm$0.0154 & 55.86$\pm$0.0145 & 48.08$\pm$0.0147
& 93.20$\pm$0.0093 & 91.81$\pm$0.0121 & 91.53$\pm$0.0128 & 90.62$\pm$0.0154
\\

Balance 
& 93.92$\pm$0.0282 & 96.05$\pm$0.0203
& 61.58$\pm$0.0302 & 59.67$\pm$0.0531 & 59.47$\pm$0.0386 & 50.64$\pm$0.0960
& 64.40$\pm$0.0097 & 61.55$\pm$0.0199 & 60.78$\pm$0.0136
& \textbf{ 99.65$\pm$0.0007 }&  \textbf{99.64$\pm$0.0007} &  \textbf{99.64$\pm$0.0007 }& \textbf{ 99.64$\pm$0.0007}
\\

New-thyroid 
& 96.54$\pm$0.0265 & 97.12$\pm$0.0299
& 89.98$\pm$0.0333 & 93.62$\pm$0.0317 & 91.39$\pm$0.0290 & 93.33$\pm$0.0346
& 90.51$\pm$0.0163 & 95.24$\pm$0.0172 & 92.30$\pm$0.0159
&  \textbf{99.52$\pm$0.0025} &  \textbf{99.50$\pm$0.0026} &  \textbf{99.50$\pm$0.0027} & \textbf{ 99.49$\pm$0.0027}
\\

Glass 
& 76.36$\pm$0.0552 & 84.82$\pm$0.0397
& 67.62$\pm$0.0726 & 74.06$\pm$0.0435 & 66.72$\pm$0.0580 & 69.43$\pm$0.0534
& 65.73$\pm$0.0250 & 74.22$\pm$0.0336 & 65.22$\pm$0.0289
& 93.35$\pm$0.0145 & 91.85$\pm$0.0155 & 91.68$\pm$0.0164 & 90.06$\pm$0.0395
\\

Heart-disease (Hungarian) 
& 34.10$\pm$0.0778 & 54.59$\pm$0.0572
& 28.20$\pm$0.0482 & 28.55$\pm$0.0795 & 26.37$\pm$0.0570 & 09.64$\pm$0.0995
& 33.53$\pm$0.0437 & 34.71$\pm$0.0275 & 31.52$\pm$0.0261
& 93.81$\pm$0.0057 & 93.21$\pm$0.0066 & 93.08$\pm$0.0069 & 92.67$\pm$0.0077
\\

Winequality-red 
& 34.57$\pm$0.0272 & 55.87$\pm$0.0247
& 30.72$\pm$0.0187 & 39.35$\pm$0.0454 & 29.72$\pm$0.0213 & 22.35$\pm$0.1267
& 30.49$\pm$0.0092 & 40.67$\pm$0.0244 & 25.31$\pm$0.0137
& 95.62$\pm$0.0026 & 95.24$\pm$0.0029 & 95.21$\pm$0.0030 & 95.02$\pm$0.0032
\\

Bioconcentration 
& 68.23$\pm$0.1001 & 74.65$\pm$0.0694
& 53.95$\pm$0.0403 & 60.65$\pm$0.0617 & 54.61$\pm$0.0431 & 59.45$\pm$0.0662
& 59.87$\pm$0.0123 & 69.73$\pm$0.0161 & 60.37$\pm$0.0160
& 96.77$\pm$0.0020 & 96.53$\pm$0.0021 & 96.51$\pm$0.0021 & 96.41$\pm$0.0022
\\

Vertebral 
& 81.28$\pm$0.0707 & 87.06$\pm$0.0502
& 77.59$\pm$0.0346 & 77.83$\pm$0.0341 & 77.28$\pm$0.0374 & 76.74$\pm$0.0367
& 79.37$\pm$0.0126 & 79.61$\pm$0.0132 & 78.68$\pm$0.0138
& 92.58$\pm$0.0036 & 91.63$\pm$0.0034 & 91.53$\pm$0.0035 & 91.08$\pm$0.0036
\\

\midrule
Average results
& 69.38$\pm$0.0526 & 79.02$\pm$0.0390
& 58.19$\pm$0.0368 & 90.16$\pm$0.0440 & 57.67$\pm$0.0370 & 53.96$\pm$0.0729
& 60.47$\pm$0.0177 & 65.01$\pm$0.0199 & 59.53$\pm$0.0174
&  \textbf{94.04$\pm$0.0057} & \textbf{ 93.20$\pm$0.0058 }&  \textbf{92.85$\pm$0.0061 }& \textbf{ 91.84$\pm$0.0095}
\\

\bottomrule
\end{tabular}%
} 
\end{table*}

Table~\ref{tab:multiClassTable3} compares the proposed framework with advanced oversampling methods, SMOTECDNN and ECDNN, while SAMME.C2 is an improved AdaBoost classifier for multiclass classification using cost-sensitive learning. The results shown in the table indicate that the proposed framework outperforms the other methods across all relevant metrics. Based on the average results in the last row, the ECDNN method achieved the lowest average G-Mean(36.30 ± 0.0277\%) and the lowest average recall (55.99 ± 0.0145\%), while SAMME.C2 achieved the lowest average F-score (55.91 ± 0.0180\%) and the lowest average Precision (57.57 ± 0.0203\%). Compared to the second-best performing model (SMOTECDNN), IMOVNO+ achieved improvements of 37.06 ± 0.039\% in G-Mean, 36.94 ± 0.0158\% in F-score, 25.28 ± 0.019\% in recall, and 35.33 ± 0.015\% in precision.

\begin{table*}[!t]
\centering
\caption{Comparison analysis with published state-of-the-art and implemented results on multi-Class imbalanced datasets using 5-fold cross-validation.}
\label{tab:multiClassTable3}
\renewcommand{\arraystretch}{1.3}
\scalebox{0.50}{%
\begin{tabular}{l
                r r r r
                r r r r
                r r r r
                r r r r
                }
\toprule
\multirow{2}{*}{Dataset} 
& \multicolumn{4}{c}{SAMME.C2 (2024) \cite{Introdcution19}} 
& \multicolumn{4}{c}{ECDNN (2023) \cite{Experimentation1}}
& \multicolumn{4}{c}{SMOTE-CDNN (2023) \cite{Experimentation2}} 
& \multicolumn{4}{c}{IMOVNO+ (Proposed)} 
\\
\cmidrule(lr){2-5} \cmidrule(lr){6-9} \cmidrule(lr){10-13} \cmidrule(lr){14-17}
& Precision($\%$) & Recall($\%$) & $F_1$-score($\%$)& G-mean($\%$)
& Precision($\%$) & Recall($\%$) & $F_1$-score($\%$)& G-mean($\%$)
& Precision($\%$) &Recall($\%$) &$F_1$-score($\%$) & G-mean($\%$)
& Precision($\%$)& Recall($\%$)& $F_1$-score($\%$) & G-mean($\%$)
\\
\midrule

Contraceptive 
& 53.81$\pm$0.0077 & 52.69$\pm$0.0085 & 52.85$\pm$0.0083 & 51.09$\pm$0.0088
& 51.13$\pm$0.0095 & 50.32$\pm$0.0084 & 50.48$\pm$0.0088 & 49.04$\pm$0.0093
& 52.58$\pm$0.0063 & 52.65$\pm$0.0071 & 50.35$\pm$0.0071 & 51.80$\pm$0.0073
& 95.58$\pm$0.0066 & 95.26$\pm$0.0083 & 95.23$\pm$0.0084 & 95.10$\pm$0.0091
\\

Balance 
& 79.97$\pm$0.0179 & 83.42$\pm$0.0220 & 80.83$\pm$0.0189 & 81.06$\pm$0.0359
& 66.99$\pm$0.0278 & 65.90$\pm$0.0089 & 65.17$\pm$0.0153 & 31.31$\pm$0.0943
& 79.10$\pm$0.0073 & 88.71$\pm$0.0069 & 80.10$\pm$0.0113 & 88.53$\pm$0.0073
& \textbf{ 99.65$\pm$0.0007 }&  \textbf{99.64$\pm$0.0007 }&  \textbf{99.64$\pm$0.0007 }& \textbf{ 99.64$\pm$0.0007}
\\

Dermatology 
& 37.60$\pm$0.0039 & 47.65$\pm$0.0027 & 39.15$\pm$0.0039 & 03.16$\pm$0.0025
& 87.06$\pm$0.0099 & 87.73$\pm$0.0084 & 86.68$\pm$0.0085 & 86.70$\pm$0.0106
& 96.46$\pm$0.0066 & 96.35$\pm$0.0060 & 96.22$\pm$0.0064 & 96.07$\pm$0.0066
& 98.62$\pm$0.0012 & 98.55$\pm$0.0011 & 98.54$\pm$0.0012 & 98.50$\pm$0.0012
\\

Zoo 
& 65.38$\pm$0.0366 & 74.11$\pm$0.0199 & 67.54$\pm$0.0279 & 24.03$\pm$0.0636
& 78.76$\pm$0.0213 & 81.10$\pm$0.0224 & 78.79$\pm$0.0198 & 37.67$\pm$0.0637
& 83.07$\pm$0.0166 & 86.09$\pm$0.0154 & 83.38$\pm$0.0158 & 55.22$\pm$0.0406
&  \textbf{100.00$\pm$0.0000 }&  \textbf{100.00$\pm$0.0000 }&  \textbf{100.00$\pm$0.0000} & \textbf{ 100.00$\pm$0.0000}
\\

Yeast 
& 35.91$\pm$0.0201 & 35.35$\pm$0.0161 & 33.44$\pm$0.0168 & 07.35$\pm$0.0134
& 60.20$\pm$0.0097 & 56.36$\pm$0.0071 & 57.15$\pm$0.0079 & 29.72$\pm$0.0089
& 45.54$\pm$0.0129 & 56.34$\pm$0.0036 & 44.19$\pm$0.0056 & 44.24$\pm$0.0331
& 93.20$\pm$0.0093 & 91.81$\pm$0.0121 & 91.53$\pm$0.0128 & 90.62$\pm$0.0154
\\

Pageblocks 
& 65.52$\pm$0.0465 & 63.84$\pm$0.0266 & 62.54$\pm$0.0338 & 26.63$\pm$0.0640
& 48.53$\pm$0.0555 & 41.17$\pm$0.0416 & 43.25$\pm$0.0415 & 05.74$\pm$0.0371
& 59.63$\pm$0.0413 & 78.56$\pm$0.0503 & 64.30$\pm$0.0438 & 49.57$\pm$0.1593
& 97.63$\pm$0.0020 & 97.34$\pm$0.0026 & 97.34$\pm$0.0026 & 97.22$\pm$0.0029
\\

Vehicle 
& 64.48$\pm$0.0108 & 61.63$\pm$0.0106 & 62.00$\pm$0.0132 & 56.83$\pm$0.0264
& 63.41$\pm$0.0095 & 65.14$\pm$0.0081 & 63.74$\pm$0.0084 & 59.68$\pm$0.0123
& 66.75$\pm$0.0110 & 68.09$\pm$0.0081 & 65.19$\pm$0.0102 & 61.19$\pm$0.0144
& 82.84$\pm$0.0100 & 79.41$\pm$0.0060 & 77.00$\pm$0.0068 & 70.59$\pm$0.0105
\\

Glass 
& 51.67$\pm$0.0189 & 51.17$\pm$0.0191 & 49.59$\pm$0.0171 & 15.88$\pm$0.0427
& 63.10$\pm$0.0413 & 57.77$\pm$0.0322 & 57.91$\pm$0.0328 & 27.65$\pm$0.0860
& 60.41$\pm$0.0255 & 66.54$\pm$0.0295 & 57.98$\pm$0.0267 & 59.17$\pm$0.0591
& 93.35$\pm$0.0145 & 91.85$\pm$0.0155 & 91.68$\pm$0.0164 & 90.06$\pm$0.0395
\\

New-thyroid 
& 82.94$\pm$0.0266 & 88.53$\pm$0.0417 & 84.92$\pm$0.0328 & 80.71$\pm$0.0902
& 40.49$\pm$0.0236 & 35.47$\pm$0.0093 & 35.54$\pm$0.0131 & 02.61$\pm$0.0051
& 39.99$\pm$0.0089 & 55.84$\pm$0.0303 & 30.94$\pm$0.0140 & 50.32$\pm$0.0464
&  \textbf{99.52$\pm$0.0025 }& \textbf{ 99.50$\pm$0.0026 }&  \textbf{99.50$\pm$0.0027} & \textbf{ 99.49$\pm$0.0027}
\\

Hungarian
& 33.03$\pm$0.0265 & 32.38$\pm$0.0137 & 31.71$\pm$0.0176 & 10.95$\pm$0.0365
& 16.57$\pm$0.0234 & 20.07$\pm$0.0077 & 17.21$\pm$0.0112 & 00.69$\pm$0.0023
& 26.03$\pm$0.0234 & 27.49$\pm$0.0376 & 21.72$\pm$0.0257 & 13.77$\pm$0.0567
& 93.81$\pm$0.0057 & 93.21$\pm$0.0066 & 93.08$\pm$0.0069 & 92.67$\pm$0.0077
\\

Winequality-red 
& 27.69$\pm$0.0140 & 28.49$\pm$0.0103 & 27.38$\pm$0.0119 & 05.46$\pm$0.0111
& 28.07$\pm$0.0131 & 26.50$\pm$0.0055 & 26.64$\pm$0.0068 & 02.37$\pm$0.0029
& 26.60$\pm$0.0065 & 34.26$\pm$0.0181 & 18.29$\pm$0.0052 & 21.96$\pm$0.0461
& 95.62$\pm$0.0026 & 95.24$\pm$0.0029 & 95.21$\pm$0.0030 & 95.02$\pm$0.0032
\\

Bioconcentration 
& 70.18$\pm$0.0162 & 57.90$\pm$0.0097 & 60.91$\pm$0.0111 & 51.79$\pm$0.0287
& 72.04$\pm$0.0095 & 57.49$\pm$0.0078 & 61.28$\pm$0.0080 & 53.62$\pm$0.0098
& 56.19$\pm$0.0066 & 64.18$\pm$0.0070 & 53.92$\pm$0.0068 & 62.63$\pm$0.0072
& 96.77$\pm$0.0020 & 96.53$\pm$0.0021 & 96.51$\pm$0.0021 & 96.41$\pm$0.0022
\\

Vertebral 
& 75.23$\pm$0.0170 & 74.12$\pm$0.0207 & 74.04$\pm$0.0212 & 71.21$\pm$0.0274
& 80.09$\pm$0.0139 & 78.82$\pm$0.0142 & 78.88$\pm$0.0140 & 77.14$\pm$0.0173
& 80.44$\pm$0.0155 & 81.22$\pm$0.0174 & 80.03$\pm$0.0168 & 80.20$\pm$0.0192
& 92.58$\pm$0.0036 & 91.63$\pm$0.0034 & 91.53$\pm$0.0035 & 91.08$\pm$0.0036
\\

\midrule
Average results
& 57.57$\pm$0.0203 & 57.79$\pm$0.0170 & 55.91$\pm$0.0180 & 37.32$\pm$0.0347
& 58.19$\pm$0.0206 & 55.99$\pm$0.0145 & 57.15$\pm$0.0151 & 36.30$\pm$0.0277
& 59.45$\pm$0.0143 & 69.33$\pm$0.0183 & 57.43$\pm$0.0150 & 56.51$\pm$0.0387
&  \textbf{94.78$\pm$0.0047} & \textbf{ 94.61$\pm$0.0050} &  \textbf{94.37$\pm$0.0051 }&  \textbf{93.57$\pm$0.0076}
\\

\bottomrule
\end{tabular}%
} 
\end{table*}

\subsubsection{Comparative analysis with OVO binary decomposition}

Table~\ref{tab:OVOTable1} The table illustrates the comparison results of the proposed framework, IMOVNO+, with binary decomposition methods. These methods are primarily designed to address binary class imbalance problems, which can be extended to multiclass problems using the OVO technique. Based on the average results shown in the last row, the proposed framework, IMOVNO+, achieves the highest performance values across the key metrics. Meanwhile, OVO Counterfactual SMOTE achieved the lowest average performance for the three key metrics, F-score, recall, and precision, while OVO MLOS achieved the lowest average G Mean value (43.75 ± 0.0146 percent). Compared to the second-best-performing model, OVO CPS 3WS, the proposed method achieved an improvement of 45.50 ± 0.0526 percent in F1 score, 30.97 ± 0.0255 percent in recall, and 27.58 ± 0.0286 percent in precision.

Table~\ref{tab:OVOTable2} Continuing the analysis of the proposed , IMOVNO+, with recent binary methods that have been extended to multi-class problems using the OVO strategy, it is evident that  
IMOVNO+ consistently outperformed the comparison methods across all key metrics. Remarkably, as the imbalance ratio increased, the method maintained its excellent performance. For instance, on the Pageblocks and Winequality-Red datasets, where the IR values are 92.6 and 68.1, respectively, IMOVNO+ 
achieved minimum scores of 97 and 95 for all key metrics, surpassing the performance of the other methods. Notably, the OVO-GDDSAD method achieved the lowest performance across all metrics, while OVO-ISMOTE ranked as the second-best 
performing model after IMOVNO+ . From the average results in the last row, compared to the best-performing model, IMOVNO+  
obtained improvements of 37.48±0.040 in G-mean, 25.13±0.016 
in F-score, 24.58±0.017 in recall, and 25.62±0.017 in precision.

Although these are recent and powerful binary classification methods, they are not appropriate for handling multiclass problems with binary decomposition approaches, as they are primarily designed for binary class scenarios. They handle multiclass problems by decomposing them into multiple binary subproblems using OVO techniques. However, several challenges are not effectively addressed by this approach. Overlapping scenarios, where more than two classes intersect, increase inter-class confusion and complicate the decision boundary, making it difficult for the classifier to distinguish between local classes. Additionally, these methods are limited in identifying outliers or small disjoint clusters. In contrast, the proposed framework focuses on reducing overlaps between classes and applies multiple regularization penalties to ensure clear decision boundaries while generating high-quality synthetic samples, making it more effective for multiclass imbalanced datasets.

\begin{table*}[!t] 
\centering
\caption{Comparison of implemented results on multi-class datasets using binarization and 5-fold cross-validation.}
\label{tab:OVOTable1}
\renewcommand{\arraystretch}{1.3}
\scalebox{0.50}{%
\begin{tabular}{l
                r r r r
                r r r r
                r r r r
                r r r r
                }
\toprule
\multirow{2}{*}{Dataset} 
& \multicolumn{4}{c}{OVO-MLOS (2025) \cite{Experimentation5}}
& \multicolumn{4}{c}{OVO-Counterfactual SMOTE (2025) \cite{Experimentation4}} 
& \multicolumn{4}{c}{OVO-CPS-3WS (2024) \cite{Experimentation3}} 
& \multicolumn{4}{c}{IMOVNO+ (Proposed)} 
\\
\cmidrule(lr){2-5} \cmidrule(lr){6-9} \cmidrule(lr){10-13} \cmidrule(lr){14-17}
& Precision($\%$)& Recall($\%$) & $F_1$-score($\%$) & G-mean($\%$) 
& Precision($\%$) &Recall($\%$)& $F_1$-score($\%$) & G-mean($\%$) 
& Precision($\%$) & Recall($\%$) & $F_1$-score($\%$) & G-mean($\%$) 
& Precision($\%$)& Recall($\%$) &$F_1$-score($\%$) & G-mean($\%$) 
\\
\midrule

Contraceptive 
& 53.38$\pm$0.0078 & 51.99$\pm$0.0066 & 52.22$\pm$0.0071 & 50.09$\pm$0.0073
& 50.33$\pm$0.0095 & 49.38$\pm$0.0076 & 49.53$\pm$0.0082 & 47.54$\pm$0.0093
& 45.29$\pm$0.0065 & 44.92$\pm$0.0051 & 44.76$\pm$0.0057 & 42.61$\pm$0.0085
& 95.58$\pm$0.0066 & 95.26$\pm$0.0083 & 95.23$\pm$0.0084 & 95.10$\pm$0.0091
\\

Balance 
& 60.40$\pm$0.0016 & 65.46$\pm$0.0017 & 62.81$\pm$0.0017 & 09.88$\pm$0.0002
& 59.10$\pm$0.0033 & 61.87$\pm$0.0055 & 60.40$\pm$0.0042 & 09.51$\pm$0.0006
& 58.32$\pm$0.0071 & 57.46$\pm$0.0089 & 57.79$\pm$0.0079 & 05.21$\pm$0.0576
&  \textbf{99.65$\pm$0.0007 }& \textbf{ 99.64$\pm$0.0007 }&  \textbf{99.64$\pm$0.0007 }&  \textbf{99.64$\pm$0.0007}
\\

Dermatology 
& 97.32$\pm$0.0019 & 97.30$\pm$0.0021 & 97.22$\pm$0.0021 & 97.11$\pm$0.0025
& 97.92$\pm$0.0036 & 97.61$\pm$0.0043 & 97.64$\pm$0.0043 & 97.42$\pm$0.0050
& 94.92$\pm$0.0077 & 94.37$\pm$0.0158 & 94.38$\pm$0.0120 & 94.01$\pm$0.0179
& 98.62$\pm$0.0012 & 98.55$\pm$0.0011 & 98.54$\pm$0.0012 & 98.50$\pm$0.0012
\\

Zoo 
& 84.82$\pm$0.0261 & 86.81$\pm$0.0169 & 84.97$\pm$0.0216 & 57.73$\pm$0.0786
& 50.33$\pm$0.0095 & 49.38$\pm$0.0076 & 49.53$\pm$0.0082 & 47.54$\pm$0.0093
& 80.46$\pm$0.0361 & 82.27$\pm$0.0445 & 80.91$\pm$0.0370 & 49.83$\pm$0.1329
&  \textbf{100.00$\pm$0.0000 }& \textbf{ 100.00$\pm$0.0000} &  \textbf{100.00$\pm$0.0000} & \textbf{ 100.00$\pm$0.0000}
\\

Yeast 
& 59.75$\pm$0.0122 & 54.67$\pm$0.0060 & 55.88$\pm$0.0064 & 27.61$\pm$0.0113
& 55.79$\pm$0.0155 & 51.38$\pm$0.0078 & 52.07$\pm$0.0085 & 69.67$\pm$0.0055
& 61.79$\pm$0.0601 & 65.57$\pm$0.0512 & 62.35$\pm$0.0542 & 52.25$\pm$0.0984
& 93.20$\pm$0.0093 & 91.81$\pm$0.0121 & 91.53$\pm$0.0128 & 90.62$\pm$0.0154
\\

Pageblocks 
& 35.14$\pm$0.0133 & 36.16$\pm$0.0142 & 35.40$\pm$0.0138 & 02.08$\pm$0.0026
& 69.87$\pm$0.0429 & 69.82$\pm$0.0416 & 69.40$\pm$0.0409 & 31.39$\pm$0.1123
& 73.59$\pm$0.0302 & 74.86$\pm$0.0302 & 73.25$\pm$0.0298 & 40.69$\pm$0.0811
& 97.63$\pm$0.0020 & 97.34$\pm$0.0026 & 97.34$\pm$0.0026 & 97.22$\pm$0.0029
\\

Vehicle 
& 76.31$\pm$0.0073 & 77.16$\pm$0.0063 & 76.37$\pm$0.0067 & 73.92$\pm$0.0083
& 69.80$\pm$0.0105 & 68.64$\pm$0.0092 & 69.01$\pm$0.0092 & 65.44$\pm$0.0100
& 64.75$\pm$0.0135 & 66.40$\pm$0.0111 & 65.37$\pm$0.0117 & 63.46$\pm$0.0134
& 82.84$\pm$0.0100 & 79.41$\pm$0.0060 & 77.00$\pm$0.0068 & 70.59$\pm$0.0105
\\

Glass 
& 64.53$\pm$0.0224 & 59.88$\pm$0.0134 & 60.02$\pm$0.0143 & 19.83$\pm$0.0200
& 60.59$\pm$0.0624 & 59.57$\pm$0.0238 & 57.80$\pm$0.0301 & 28.35$\pm$0.0847
& 62.04$\pm$0.0743 & 62.20$\pm$0.0523 & 60.83$\pm$0.0579 & 47.23$\pm$0.1295
& 93.35$\pm$0.0145 & 91.85$\pm$0.0155 & 91.68$\pm$0.0164 & 90.06$\pm$0.0395
\\

New-thyroid 
& 97.07$\pm$0.0090 & 91.23$\pm$0.0121 & 93.40$\pm$0.0097 & 90.38$\pm$0.0137
& 97.03$\pm$0.0103 & 93.40$\pm$0.0174 & 94.77$\pm$0.0124 & 92.90$\pm$0.0210
& 94.08$\pm$0.0163 & 91.59$\pm$0.0296 & 92.18$\pm$0.0233 & 90.81$\pm$0.0351
& \textbf{ 99.52$\pm$0.0025 }&  \textbf{99.50$\pm$0.0026 }&  \textbf{99.50$\pm$0.0027
}& \textbf{ 99.49$\pm$0.0027}
\\

Hungarian
& 33.17$\pm$0.0292 & 31.23$\pm$0.0105 & 30.30$\pm$0.0170 & 05.66$\pm$0.0226
& 27.45$\pm$0.0360 & 20.04$\pm$0.0302 & 22.08$\pm$0.0322 & 11.12$\pm$0.0279
& 17.80$\pm$0.0431 & 16.89$\pm$0.0252 & 16.48$\pm$0.0322 & 05.78$\pm$0.0275
& 93.81$\pm$0.0057 & 93.21$\pm$0.0066 & 93.08$\pm$0.0069 & 92.67$\pm$0.0077
\\

Winequality-red 
& 31.06$\pm$0.0056 & 28.55$\pm$0.0026 & 28.70$\pm$0.0030 & 02.29$\pm$0.0002
& 41.16$\pm$0.0314 & 35.47$\pm$0.0125 & 36.44$\pm$0.0162 & 04.64$\pm$0.0120
& 39.31$\pm$0.0290 & 35.83$\pm$0.0185 & 36.55$\pm$0.0181 & 01.82$\pm$0.0364
& 95.62$\pm$0.0026 & 95.24$\pm$0.0029 & 95.21$\pm$0.0030 & 95.02$\pm$0.0032
\\

Bioconcentration 
& 83.94$\pm$0.0037 & 60.26$\pm$0.0067 & 64.69$\pm$0.0077 & 53.49$\pm$0.0130
& 78.25$\pm$0.0261 & 63.01$\pm$0.0075 & 66.63$\pm$0.0108 & 58.08$\pm$0.0111
& 65.66$\pm$0.0204 & 60.79$\pm$0.0115 & 62.35$\pm$0.0136 & 58.08$\pm$0.0125
& 96.77$\pm$0.0020 & 96.53$\pm$0.0021 & 96.51$\pm$0.0021 & 96.41$\pm$0.0022
\\

Vertebral 
& 80.55$\pm$0.0070 & 79.97$\pm$0.0039 & 79.72$\pm$0.0059 & 78.64$\pm$0.0089
& 57.13$\pm$0.0074 & 57.80$\pm$0.0106 & 57.32$\pm$0.0079 & 09.03$\pm$0.0013
& 75.53$\pm$0.0227 & 75.02$\pm$0.0206 & 74.76$\pm$0.0226 & 72.80$\pm$0.0256
& 92.58$\pm$0.0036 & 91.63$\pm$0.0034 & 91.53$\pm$0.0035 & 91.08$\pm$0.0036
\\

\midrule
Average results
& 66.34$\pm$0.0113 & 63.13$\pm$0.0079 & 63.21$\pm$0.0090 & 43.75$\pm$0.0146
& 62.67$\pm$0.0206 & 59.80$\pm$0.0143 & 60.20$\pm$0.0148 & 44.05$\pm$0.0239
& 67.20$\pm$0.0282 & 63.64$\pm$0.0250 & 63.15$\pm$0.0251 & 48.07$\pm$0.0520
&  \textbf{94.78$\pm$0.0047 }&  \textbf{94.61$\pm$0.0050 }&  \textbf{94.37$\pm$0.0051} &  \textbf{93.57$\pm$0.0076}
\\

\bottomrule
\end{tabular}%
} 
\end{table*}

\begin{table*}[!t]
\centering
\caption{Comparison of implemented results on multi-class datasets using binarization and 5-fold cross-validation.}
\label{tab:OVOTable2}
\renewcommand{\arraystretch}{1.3}
\scalebox{0.50}{%
\begin{tabular}{l
                r r r r
                r r r r
                r r r r
                r r r r
                }
\toprule
\multirow{2}{*}{Dataset} 
& \multicolumn{4}{c}{OVO-HSCF (2025) \cite{Experimentation6}}
& \multicolumn{4}{c}{OVO-GDDSAD (2025) \cite{Experimentation7}} 
& \multicolumn{4}{c}{OVO-ISMOTE (2025)  \cite{Experimentation16}} 
& \multicolumn{4}{c}{IMOVNO+ (Proposed)} 
\\
\cmidrule(lr){2-5} \cmidrule(lr){6-9} \cmidrule(lr){10-13} \cmidrule(lr){14-17}
&Precision($\%$) & Recall($\%$) & $F_1$-score($\%$)  & G-mean($\%$) 
& Precision($\%$) & Recall($\%$) &$F_1$-score($\%$)  & G-mean($\%$) 
& Precision($\%$) &  Recall($\%$) & $F_1$-score($\%$)  & G-mean($\%$) 
& Precision($\%$)&  Recall($\%$) & $F_1$-score($\%$)  & G-mean($\%$) 
\\
\midrule

Contraceptive 
& 45.29$\pm$0.0065 & 44.92$\pm$0.0051 & 44.76$\pm$0.0057 & 42.61$\pm$0.0085
& 47.01$\pm$0.0519 & 43.38$\pm$0.0108 & 34.65$\pm$0.0258 & 24.18$\pm$0.0708
& 52.81$\pm$0.0118 & 52.48$\pm$0.0129 & 52.46$\pm$0.0123 & 51.549$\pm$0.0145
& 95.58$\pm$0.0066 & 95.26$\pm$0.0083 & 95.23$\pm$0.0084 & 95.10$\pm$0.0091
\\

Balance 
& 58.32$\pm$0.0071 & 57.46$\pm$0.0089 & 57.79$\pm$0.0079 & 13.14$\pm$0.0455
& 78.15$\pm$0.0104 & 86.86$\pm$0.0164 & 78.93$\pm$0.0127 & 86.64$\pm$0.0166
& 59.84$\pm$0.0042 & 60.43$\pm$0.0042 & 60.07$\pm$0.0037 & 15.40$\pm$0.0560
& \textbf{ 99.65$\pm$0.0007 }& \textbf{ 99.64$\pm$0.0007 }&  \textbf{99.64$\pm$0.0007 }& \textbf{ 99.64$\pm$0.0007}
\\

Dermatology 
& 95.25$\pm$0.0098 & 94.19$\pm$0.0138 & 94.32$\pm$0.0139 & 93.56$\pm$0.0184
& 94.88$\pm$0.0079 & 93.71$\pm$0.0110 & 93.87$\pm$0.0099 & 93.17$\pm$0.0120
& 97.82$\pm$0.0061 & 97.59$\pm$0.0060 & 97.57$\pm$0.0064 & 97.40$\pm$0.0067
& 98.62$\pm$0.0012 & 98.55$\pm$0.0011 & 98.54$\pm$0.0012 & 98.50$\pm$0.0012
\\

Zoo 
& 83.13$\pm$0.0347 & 85.55$\pm$0.0287 & 83.29$\pm$0.0328 & 52.18$\pm$0.0852
& 78.08$\pm$0.0382 & 78.71$\pm$0.0424 & 75.23$\pm$0.0410 & 50.97$\pm$0.1025
& 86.60$\pm$0.0162 & 90.62$\pm$0.0111 & 89.33$\pm$0.0130 & 67.55$\pm$0.0493
&  \textbf{100.00$\pm$0.0000 }&  \textbf{100.00$\pm$0.0000} &  \textbf{100.00$\pm$0.0000 }&  \textbf{100.00$\pm$0.0000}
\\

Yeast 
& 50.88$\pm$0.0250 & 49.84$\pm$0.0205 & 49.16$\pm$0.0208 & 24.66$\pm$0.0350
& 45.55$\pm$0.0187 & 48.96$\pm$0.0181 & 45.48$\pm$0.0160 & 27.84$\pm$0.0386
& 57.71$\pm$0.0120 & 59.35$\pm$0.0118 & 57.78$\pm$0.0107 & 36.88$\pm$0.0306
& 93.20$\pm$0.0093 & 91.81$\pm$0.0121 & 91.53$\pm$0.0128 & 90.62$\pm$0.0154
\\

Pageblocks 
& 75.31$\pm$0.0559 & 72.41$\pm$0.0287 & 72.01$\pm$0.0419 & 45.88$\pm$0.0908
& 34.67$\pm$0.0232 & 52.49$\pm$0.0403 & 36.44$\pm$0.0262 & 25.36$\pm$0.0817
& 70.46$\pm$0.0431 & 78.64$\pm$0.0492 & 76.01$\pm$0.0445 & 57.13$\pm$0.1278
& 97.63$\pm$0.0020 & 97.34$\pm$0.0026 & 97.34$\pm$0.0026 & 97.22$\pm$0.0029
\\

Vehicle 
& 70.90$\pm$0.0111 & 70.40$\pm$0.0118 & 70.41$\pm$0.0112 & 67.66$\pm$0.0135
& 42.77$\pm$0.0214 & 44.88$\pm$0.0128 & 42.17$\pm$0.0126 & 34.51$\pm$0.0400
& 74.80$\pm$0.0058 & 75.58$\pm$0.0057 & 74.94$\pm$0.0058 & 72.06$\pm$0.0079
& 82.84$\pm$0.0100 & 79.41$\pm$0.0060 & 77.00$\pm$0.0068 & 70.59$\pm$0.0105
\\

Glass 
& 73.55$\pm$0.0409 & 69.83$\pm$0.0277 & 69.59$\pm$0.0297 & 53.80$\pm$0.0953
& 17.53$\pm$0.0000 & 19.78$\pm$0.0000 & 16.25$\pm$0.0000 & 02.41$\pm$0.0000
& 78.12$\pm$0.0297 & 77.99$\pm$0.0162 & 75.96$\pm$0.0168 & 65.91$\pm$0.0762
& 93.35$\pm$0.0145 & 91.85$\pm$0.0155 & 91.68$\pm$0.0164 & 90.06$\pm$0.0395
\\

New-thyroid 
& 94.08$\pm$0.0163 & 91.59$\pm$0.0296 & 92.18$\pm$0.0233 & 90.81$\pm$0.0351
& 73.97$\pm$0.0097 & 78.29$\pm$0.0128 & 74.71$\pm$0.0112 & 77.48$\pm$0.0153
& 95.19$\pm$0.0131 & 95.32$\pm$0.0189 & 94.93$\pm$0.0165 & 95.03$\pm$0.0209
&  \textbf{99.52$\pm$0.0025 }&  \textbf{99.50$\pm$0.0026 }&  \textbf{99.50$\pm$0.0027} & \textbf{ 99.49$\pm$0.0027}
\\

Hungarian
& 29.01$\pm$0.0430 & 28.89$\pm$0.0250 & 28.22$\pm$0.0317 & 06.08$\pm$0.0210
& 20.18$\pm$0.0132 & 20.43$\pm$0.0301 & 16.61$\pm$0.0166 & 11.01$\pm$0.0449
& 36.37$\pm$0.0367 & 35.35$\pm$0.0241 & 34.75$\pm$0.0255 & 15.27$\pm$0.0548
& 93.81$\pm$0.0057 & 93.21$\pm$0.0066 & 93.08$\pm$0.0069 & 92.67$\pm$0.0077
\\

Winequality-red 
& 39.31$\pm$0.0290 & 35.83$\pm$0.0185 & 36.55$\pm$0.0181 & 10.44$\pm$0.0355
& 21.95$\pm$0.0064 & 26.02$\pm$0.0293 & 19.68$\pm$0.0103 & 06.07$\pm$0.0269
& 38.01$\pm$0.0124 & 40.20$\pm$0.0157 & 38.60$\pm$0.0127 & 12.04$\pm$0.0274
& 95.62$\pm$0.0026 & 95.24$\pm$0.0029 & 95.21$\pm$0.0030 & 95.02$\pm$0.0032
\\

Bioconcentration 
& 65.66$\pm$0.0204 & 60.79$\pm$0.0115 & 62.35$\pm$0.0136 & 58.08$\pm$0.0125
& 47.36$\pm$0.0144 & 50.15$\pm$0.0105 & 45.69$\pm$0.0119 & 48.40$\pm$0.0225
& 70.95$\pm$0.0118 & 66.43$\pm$0.0141 & 67.85$\pm$0.0131 & 64.16$\pm$0.0200
& 96.77$\pm$0.0020 & 96.53$\pm$0.0021 & 96.51$\pm$0.0021 & 96.41$\pm$0.0022
\\

Vertebral 
& 75.53$\pm$0.0227 & 75.02$\pm$0.0206 & 74.76$\pm$0.0226 & 72.80$\pm$0.0256
& 77.05$\pm$0.0095 & 77.94$\pm$0.0097 & 76.07$\pm$0.0105 & 76.17$\pm$0.0115
& 80.36$\pm$0.0154 & 80.40$\pm$0.0143 & 79.88$\pm$0.0127 & 78.81$\pm$0.0147
& 92.58$\pm$0.0036 & 91.63$\pm$0.0034 & 91.53$\pm$0.0035 & 91.08$\pm$0.0036
\\

\midrule
Average results
& 65.86$\pm$0.0248 & 64.36$\pm$0.0193 & 65.89$\pm$0.0223 & 48.60$\pm$0.040
& 52.24$\pm$0.0173 & 55.51$\pm$0.0188 & 50.44$\pm$0.0157 & 43.40$\pm$0.0372
& 69.16$\pm$0.0167 & 70.03$\pm$0.0157 & 69.24$\pm$0.0147 & 56.09$\pm$0.0390
&  \textbf{94.78$\pm$0.0047 }&  \textbf{94.61$\pm$0.0050} &  \textbf{94.37$\pm$0.0051} &  \textbf{93.57$\pm$0.0076}
\\

\bottomrule
\end{tabular}%
} 
\end{table*}

\subsubsection{Comparative analysis for binary classification}

To evaluate the proposed framework, IMOVNO+,for handling class imbalance in a binary classification context, a comparison with recent state-of-the-art binary classification methods is presented in Tables~\ref{tab:binary_Table1} and 
\ref{tab:binary_Table2}. Based on the findings in Table~\ref{tab:binary_Table1}, the proposed framework, IMOVNO+,outperforms other methods across all key metrics. It achieved a peak precision value of 100.00 ± 0.0000\% on the Yeast 2vs8 dataset. Additionally, it achieved approximately similar results for the four key metrics on the Ecoli 1, Yeast 4, and Yeast 6 datasets.
The results in Tables~\ref{tab:binary_Table2} show that the proposed framework, IMOVNO+, consistently surpasses other approaches across various binary class imbalance scenarios. As shown by the average results in the last row, DT-Boosting achieved the lowest values in G-Mean (71.02 ± 0.0372\%) and F-score (64.27 ± 0.0367\%). Compared to the second-best performing model (LR-DC-Glow) in the G-Mean metric, IMOVNO+ achieved an improvement of 13.77\%, while it outperformed the second-best performing model (MLCC-IBC) in F-score by 20.92\%.
\begin{table*}[!t]
\centering
\caption{Comparison analysis with published state-of-the-art and implemented results on binary imbalanced datasets using 5-fold cross-validation.}
\label{tab:binary_Table1}
\renewcommand{\arraystretch}{1.3}
\scalebox{0.58}{%
\begin{tabular}{l r r r r r r r r r r r r r r r}
\toprule
\multirow{2}{*}{Dataset} 
& \multicolumn{3}{c}{SIMPOR (2023) \cite{Experimentation17}} 
& \multicolumn{2}{c}{\makecell[l]{Adaptive SV-Borderline \\SMOTE-SVM (2024) \cite{Experimentation18}} }
& \multicolumn{2}{c}{SVM-OMOS (2024) \cite{Experimentation19}} 
& \multicolumn{3}{c}{UBMD (2024) \cite{Experimentation20}} 
& \multicolumn{5}{c}{IMOVNO+ (Proposed)} 
\\
\cmidrule(lr){2-4} 
\cmidrule(lr){5-6} 
\cmidrule(lr){7-8} 
\cmidrule(lr){9-11} 
\cmidrule(lr){12-16}
& Precision($\%$)& Recall($\%$) & $F_1$-score($\%$)
& $F_1$-score($\%$)& G-mean($\%$)
& $F_1$-score($\%$) & G-mean($\%$)
& Recall($\%$) & $F_1$-score($\%$) & G-mean($\%$) 
& Accuracy($\%$) & Precision($\%$) & Recall($\%$) & $F_1$-score($\%$) & G-mean($\%$)
\\
\midrule

yeast1
& 72.70 & 70.40 & 71.50
& 79.00 & 68.30
& 58.9$\pm$0.012 & 71.10$\pm$0.010
& 78.84 & 63.32 & 75.00
& 95.06$\pm$0.0027 & 98.12$\pm$0.0031 & 91.91$\pm$0.0064 & 94.88$\pm$0.0029 & 94.99$\pm$0.0028 
\\

Ecoli1
& 81.00 & 85.40 & 83.10
& 88.60 & 87.90
& 83.9$\pm$0.022 & 89.9$\pm$0.016
& 92.80 & 80.29 & 88.10
&\textbf{ 99.18$\pm$0.0042} & 98.61$\pm$0.0068 & \textbf{99.80$\pm$0.0024 }& \textbf{99.17$\pm$0.0042} & \textbf{99.19$\pm$0.0041 }
\\

yeast-1\_vs\_7
& 77.90 & 65.50 & 71.00
& 50.20 & 69.70
& 30.8$\pm$0.069 & 7.72$\pm$0.077
& 78.67 & 26.61 & 73.92
& 96.98$\pm$0.0052 & 95.05$\pm$0.0099 & \textbf{99.26$\pm$0.0078 }& 97.07$\pm$0.0049 & 96.94$\pm$0.0054
\\

Yeast2\_vs\_8
& 91.30 & 87.20 & 88.40
& 98.70 & 74.30
& 71.4$\pm$0.099 & 93.90$\pm$0.064
& 90.00 & 28.77 & 83.52
& 98.80$\pm$0.0047 &\textbf{ 100.00$\pm$0.0000 }& 97.62$\pm$0.0094 & 98.77$\pm$0.0050 & 98.79$\pm$0.0048
\\

glass2
& 73.10 & 74.70 & 73.70
& 53.00 & 78.10
& 45.30$\pm$0.106 & 83.50$\pm$0.071
& 76.33 & 25.84 & 68.18
& 92.99$\pm$0.0049 & 88.36$\pm$0.0066 & 99.31$\pm$0.0025 & 93.46$\pm$0.0042 & 92.74$\pm$0.0053 
\\

Abalone9\_vs\_18
& 80.60 & 75.10 & 77.70
& - & -
& 83.90$\pm$0.022 & 89.90$\pm$0.016
& 68.44 & 22.32 & 69.39
& 98.98$\pm$0.0000 & 98.99$\pm$0.0000 & 98.99$\pm$0.0000 & 98.98$\pm$0.0000 & 98.98$\pm$0.0000 
\\

pima
& 77.60 & 77.80 & 77.70
& - & -
& 69.00$\pm$0.041 & 76.00$\pm$0.033
& 83.20 & 70.37 & 77.26
& 96.21$\pm$0.0041 & 97.42$\pm$0.0062 & 95.02$\pm$0.0094 & 96.15$\pm$0.0044 & 96.17$\pm$0.0042 
\\

Yeast4
& 80.00 & 79.00 & 79.30
& 88.60 & 82.70
& - & -
& 86.69 & 25.45 & 83.87
& \textbf{99.19$\pm$0.0005 }& \textbf{99.68$\pm$0.0006} & 98.70$\pm$0.0011 &\textbf{ 99.19$\pm$0.0005} &\textbf{ 99.19$\pm$0.0005} 
\\

Yeast6
& 78.50 & 71.40 & 74.50
& 93.30 & 88.40
& - & -
& 86.29 & 44.92 & 86.22
& \textbf{99.66$\pm$0.0005 }& \textbf{99.34$\pm$0.0010} & \textbf{100.00$\pm$0.0000} & \textbf{99.67$\pm$0.0005} & \textbf{99.66$\pm$0.0005}
\\

\bottomrule
\end{tabular}%
}
\end{table*}

\begin{table*}[!t] 
\centering
\caption{Comparison analysis on binary imbalanced datasets using 5-fold cross-validation.}
\label{tab:binary_Table2}
\renewcommand{\arraystretch}{1.3}
\scalebox{0.67}{%
\begin{tabular}{l
                r r r r
                r r
                r r
                r r r r
                }
\toprule
\multirow{2}{*}{Dataset} 
& \multicolumn{4}{c}{DT-Boosting} 
& \multicolumn{2}{c}{LR-DC-Glow (2023) \cite{Experimentation21}} 
& \multicolumn{2}{c}{MLCC-IBC (2023) \cite{Experimentation22}} 
& \multicolumn{4}{c}{IMOVNO+ (Proposed)} 
\\
\cmidrule(lr){2-5} \cmidrule(lr){6-7} \cmidrule(lr){8-9} \cmidrule(lr){10-13}
& Precision($\%$) & Recall($\%$) & $F_1$-score($\%$) & G-mean($\%$)
& $F_1$-score($\%$) & G-mean($\%$)
& $F_1$-score($\%$) & G-mean($\%$)
& Precision($\%$) & Recall($\%$) & $F_1$-score($\%$) & G-mean($\%$)
\\
\midrule

messidor\_features
& 68.33$\pm$0.0143 & 63.19$\pm$0.0296 & 65.40$\pm$0.0128 & 64.42$\pm$0.0091
& 75.56 & 74.85
& 75.07 & 67.92
& 98.20$\pm$0.0061 & 87.24$\pm$0.0171 & 92.29$\pm$0.0070 & 92.56$\pm$0.0064
\\

Pima
& 66.66$\pm$0.0209 & 56.70$\pm$0.0201 & 60.92$\pm$0.0094 & 69.02$\pm$0.0081
& 67.88 & 75.22
& 69.94 & 75.53
& 97.42$\pm$0.0062 & 95.02$\pm$0.0094 & 96.15$\pm$0.0044 & 96.17$\pm$0.0042
\\

biodeg
& 74.84$\pm$0.0146 & 73.51$\pm$0.0250 & 73.95$\pm$0.0118 & 79.96$\pm$0.0107
& 78.72 & 83.87
& 84.07 & 88.61
& 97.89$\pm$0.0015 & 92.70$\pm$0.0067 & 95.20$\pm$0.0037 & 95.30$\pm$0.0035
\\

vehicle3
& 60.66$\pm$0.0349 & 29.91$\pm$0.0175 & 39.44$\pm$0.0204 & 52.33$\pm$0.0164
& 71.32 & 81.97
& 68.80 & 82.05
& 95.00$\pm$0.0078 & 94.74$\pm$0.0068 & 94.84$\pm$0.0057 & 94.82$\pm$0.0057
\\

Ecoli1
& 78.64$\pm$0.0330 & 76.07$\pm$0.0465 & 76.51$\pm$0.0352 & 83.97$\pm$0.0283
& 75.28 & 86.15
& 79.12 & 90.59
& 98.61$\pm$0.0068 &  \textbf{99.80$\pm$0.0024} &  \textbf{99.17$\pm$0.0042 }&  \textbf{99.19$\pm$0.0041}
\\

new-thyroid1
& 93.98$\pm$0.0334 & 92.00$\pm$0.0420 & 92.44$\pm$0.0268 & 95.11$\pm$0.0219
& 94.00 & 96.34
& 98.46 & 98.52
& 98.76$\pm$0.0023 &  \textbf{100.00$\pm$0.0000 }&  \textbf{99.37$\pm$0.0012 }& \textbf{ 99.35$\pm$0.0012}
\\

segment0
& 98.42$\pm$0.0056 & 98.33$\pm$0.0039 & 98.36$\pm$0.0029 & 99.02$\pm$0.0019
& 99.10 & 99.34
& 98.69 & 98.83
& \textbf{ 99.21$\pm$0.0007 }&  \textbf{99.77$\pm$0.0002 }&  \textbf{99.49$\pm$0.0003} & \textbf{ 99.49$\pm$0.0003}
\\

yeast3
& 77.99$\pm$0.0171 & 77.94$\pm$0.0242 & 77.66$\pm$0.0151 & 86.90$\pm$0.0139
& 67.65 & 89.60
& 80.25 & 90.16
& 96.19$\pm$0.0012 & 95.55$\pm$0.0005 & 95.86$\pm$0.0007 & 95.87$\pm$0.0007
\\

ecoli3
& 58.96$\pm$0.0488 & 44.86$\pm$0.0462 & 48.29$\pm$0.0313 & 64.25$\pm$0.0322
& 63.59 & 88.24
& 73.09 & 93.11
& 98.96$\pm$0.0028 & 97.49$\pm$0.0076 & 98.20$\pm$0.0038 & 98.22$\pm$0.0037
\\

yeast-2\_vs\_4
& 80.06$\pm$0.0274 & 68.84$\pm$0.0350 & 73.07$\pm$0.0278 & 81.71$\pm$0.0217
& 62.35 & 83.08
& 86.45 & 91.57
&  \textbf{99.38$\pm$0.0040 }& 96.07$\pm$0.0217 & 97.63$\pm$0.0108 & 97.67$\pm$0.0104
\\

ecoli-0-1-4-7\_vs\_5-6
& 84.10$\pm$0.0338 & 66.40$\pm$0.0512 & 72.07$\pm$0.0411 & 79.72$\pm$0.0370
& 80.00 & 92.42
& 85.49 & 92.84
&  \textbf{99.66$\pm$0.0031} & \textbf{ 99.86$\pm$0.0017} &  \textbf{99.75$\pm$0.0018} &  \textbf{99.75$\pm$0.0018}
\\

ecoli4
& 80.11$\pm$0.0548 & 69.00$\pm$0.0583 & 71.01$\pm$0.0432 & 81.16$\pm$0.0402
& 75.65 & 90.79
& 87.09 & 94.12
&  \textbf{99.90$\pm$0.0022 }&  \textbf{99.96$\pm$0.0011 }&  \textbf{99.93$\pm$0.0016} & \textbf{ 99.93$\pm$0.0016}
\\

abalone9-18
& 60.70$\pm$0.1327 & 16.89$\pm$0.0464 & 24.91$\pm$0.0596 & 35.56$\pm$0.0754
& 54.30 & 87.89
& 69.39 & 79.59
& 98.99$\pm$0.0000 & 98.99$\pm$0.0000 & 98.98$\pm$0.0000 & 98.98$\pm$0.0000
\\

Wilt
& 92.56$\pm$0.0148 & 50.54$\pm$0.0277 & 65.05$\pm$0.0219 & 70.82$\pm$0.0191
& 83.21 & 97.17
& 89.19 & 93.51
& 97.87$\pm$0.0007 & 99.85$\pm$0.0029 & 98.85$\pm$0.0014 & 98.83$\pm$0.0014
\\

ecoli-0-1-3-7\_vs\_2-6
& 44.67$\pm$0.2023 & 41.00$\pm$0.1758 & 41.00$\pm$0.1862 & 45.43$\pm$0.1861
& 51.67 & 86.86
& 90.00 & 91.21
&  \textbf{99.62$\pm$0.0019 }& \textbf{ 100.00$\pm$0.0000 }&  \textbf{99.80$\pm$0.0010} & \textbf{ 99.81$\pm$0.0009}
\\

Yeast6
& 62.15$\pm$0.0690 & 39.14$\pm$0.0384 & 45.54$\pm$0.0338 & 59.48$\pm$0.0437
& 31.01 & 88.06
& 56.30 & 79.66
& \textbf{ 99.34$\pm$0.0010} & \textbf{ 100.00$\pm$0.0000} &  \textbf{99.67$\pm$0.0005} &  \textbf{99.66$\pm$0.0005}
\\

abalone-20\_vs\_8-9-10
& 26.00$\pm$0.1248 & 9.33$\pm$0.0343 & 12.87$\pm$0.0458 & 18.95$\pm$0.0660
& 28.83 & 84.54
& 57.93 & 69.64
&  \textbf{99.86$\pm$0.0000 }&  \textbf{100.00$\pm$0.0000 }&  \textbf{99.93$\pm$0.0000} &  \textbf{99.93$\pm$0.0000}
\\

\midrule
Average results
& 71.02$\pm$0.0498 & 58.39$\pm$0.0413 & 64.27$\pm$0.0367 & 71.02$\pm$0.0372
& 63.54 & 84.20
& 76.97 & 83.32
& \textbf{98.40$\pm$0.0030} &  \textbf{97.61$\pm$0.0040 }&  \textbf{97.89$\pm$0.0028 }&  \textbf{97.97$\pm$0.0028}
\\

\bottomrule
\end{tabular}%
}
\end{table*}

\begin{figure*}[htbp]
    \centering
    
    \subfloat[\centering Accuracy scores ]{{\includegraphics[width=9cm,height=6cm]{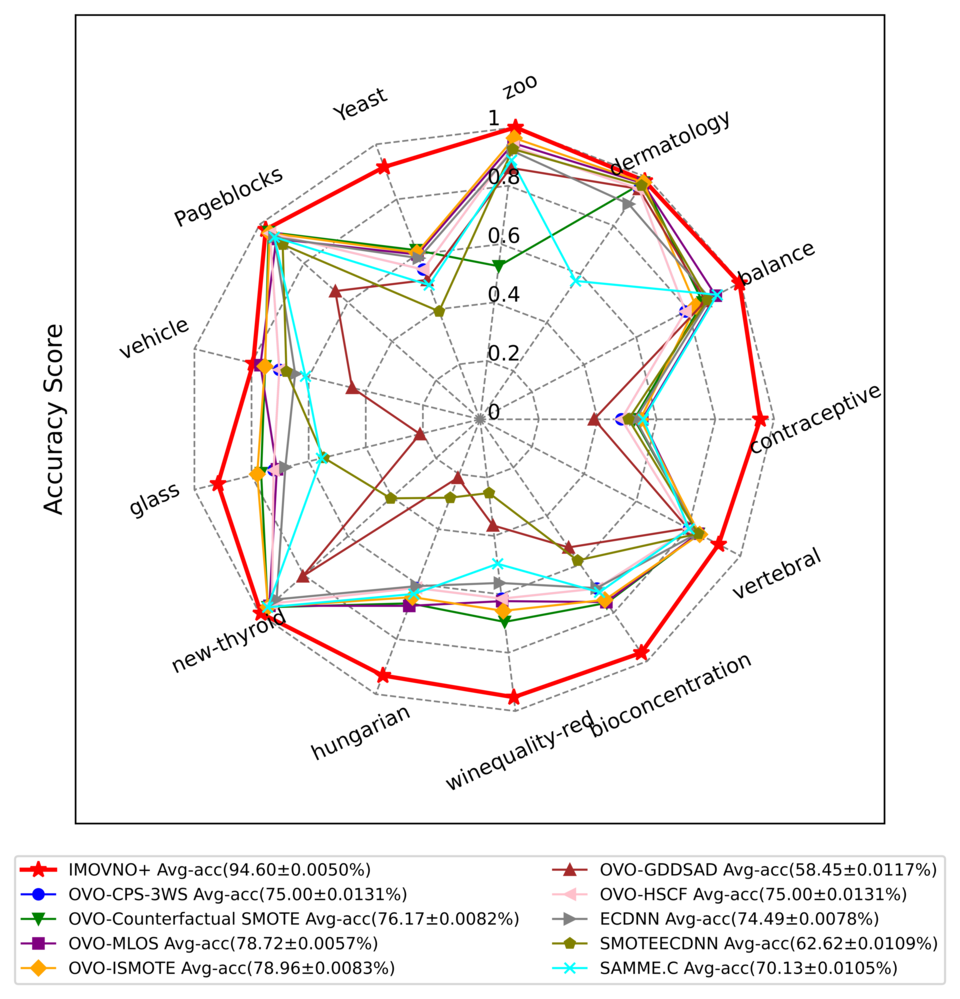} }}
    \subfloat[\centering  AUC scores ]{{\includegraphics[width=9cm,height=6cm]{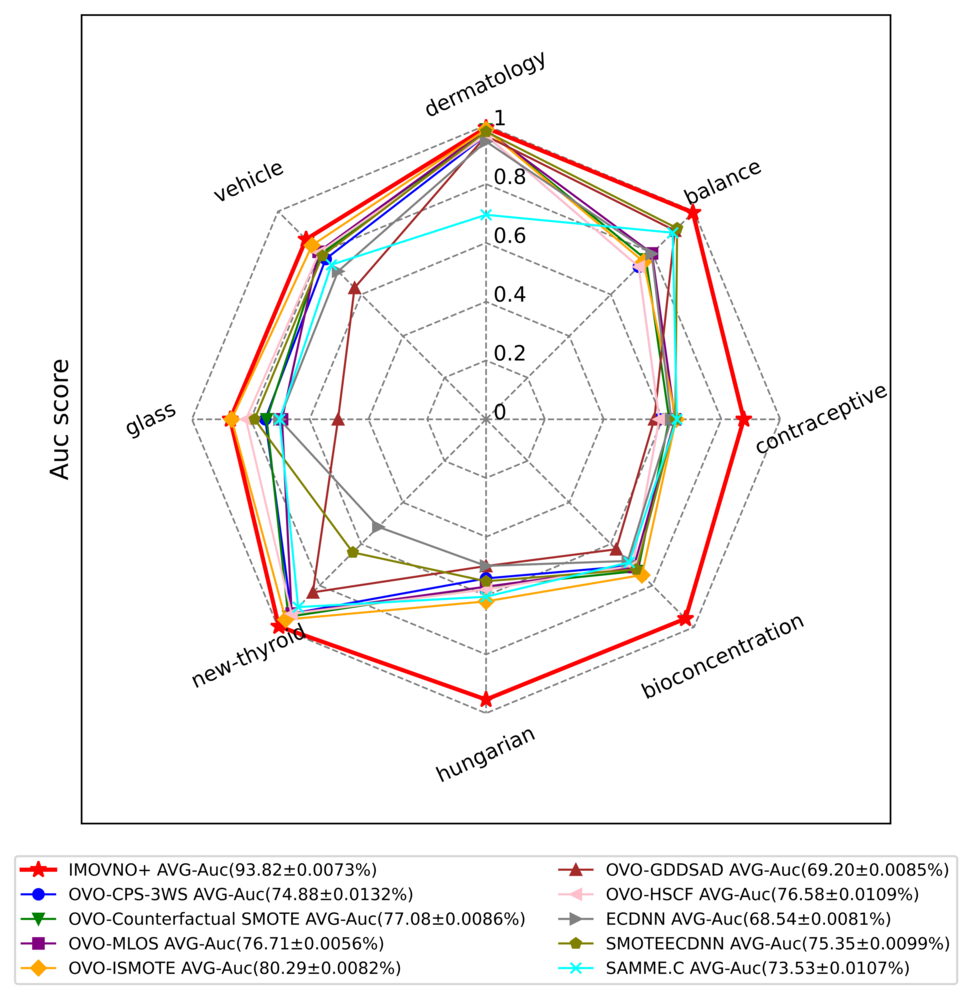} }}
    
    \caption{Comparative analysis of accuracy and AUC scores across various multi-class imbalanced datasets.}
    \label{fig:AccuracyAucRader}
\end{figure*}

In Figs.~\ref{fig:AccuracyAucRader}(a), A radar chart displays the comparison of accuracy performance of the 10 imbalanced methods on 13 multi-class imbalanced datasets. Each dataset is represented by an equiangular spoke in the radar chart, and the performance score of each method is shown as a 13-sided shape. The evaluation performance ranges between 0 and 1. For example, 0 indicates that all model predictions are incorrect, while 1 indicates the optimal value where all predicted samples are correct. The proposed framework’s performance across all datasets is represented by the red line. It is evident that the IMOVNO+ line covers all other broken lines, demonstrating its superior performance across all datasets. It achieved optimal or near-optimal performance (100\%) on the Zoo, Dermatology, Balance, and New-Thyroid datasets. In contrast, the OVO-GDDSAD method achieved lower accuracy scores, with its line located below all other methods and an average accuracy of 58.45 ± 0.0117\%. Unlike the traditional assumption that higher accuracy combined with lower performance on other metrics may indicate overfitting toward the majority class, our model is trained on balanced datasets where higher accuracy is a strong indicator of the effectiveness of the proposed framework. Additionally, considering other metrics such as F-score and precision from previous experiments (Tables), all key metrics achieved the highest performance, supporting that IMOVNO+ mitigates overfitting by balancing overall performance. Compared to the second-best performing model (OVO-ISMOTE), IMOVNO+ achieved an improvement of 15.64 ± 0.010\%.

Similarly, a radar chart is used to display the comparison of AUC performance of the 10 imbalanced methods on 8 datasets in Figs.~\ref{fig:AccuracyAucRader}(b). Higher AUC values indicate the ability of the model to distinguish between classes \cite{Experimentation11}. The chart shows the remarkable performance of the proposed framework, IMOVNO+, with its red line located above all other methods. The average AUC results show that IMOVNO+  method improved by 25.28 ± 0.011\% over the worst-performing model (ECDNN) and by 13.53 ± 0.011\% over the best-performing model (OVO-ISMOTE).

\subsubsection{Analysis  the impact of noise reduction}
\begin{figure*}[htbp]
    \centering
    
    \subfloat[\centering Contraceptive  dataset]{{\includegraphics[width=6cm,height=6cm]{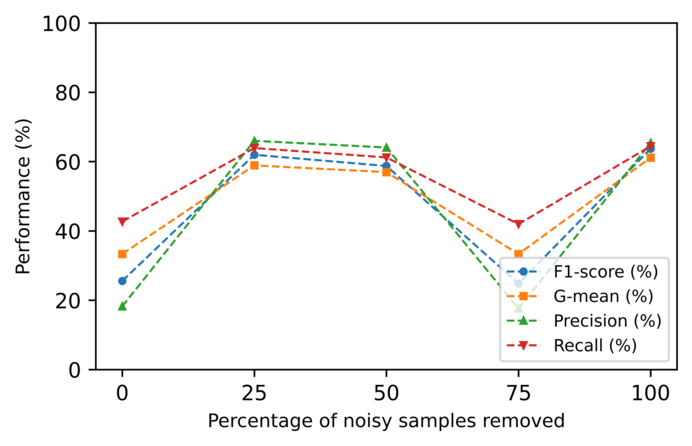} }}
    \subfloat[\centering Vehicle dataset]{{\includegraphics[width=6cm,height=6cm]{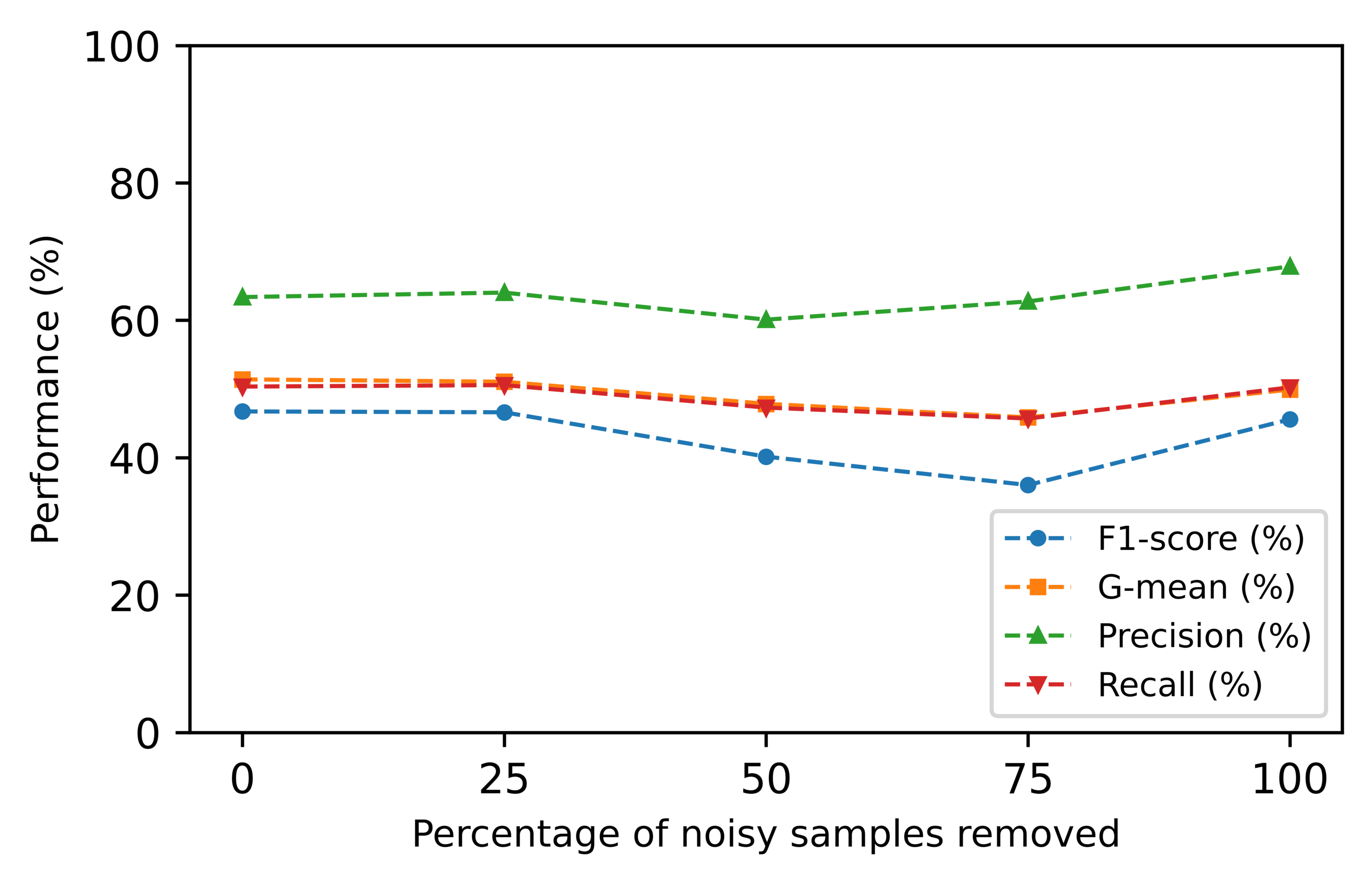} }}
    \subfloat[\centering Vertebral dataset]{{\includegraphics[width=6cm,height=6cm]{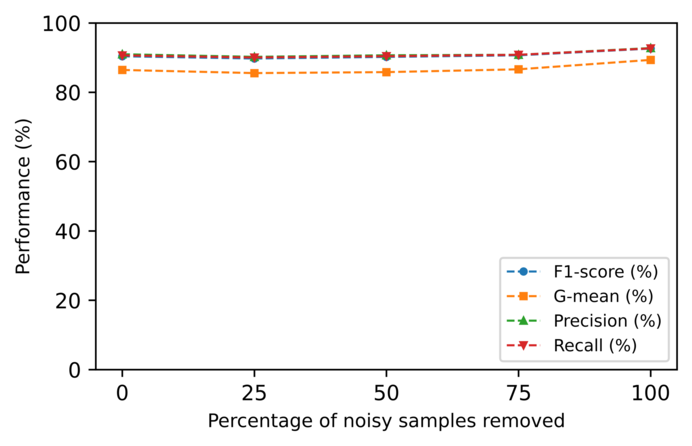} }}

    \caption{Experimental results of noise reduction for different noise thresholds (IR $\le$ 3).}
    \label{fig:NoiseReduction3}
\end{figure*}

\begin{figure*}[htbp]
    \centering
    
    \subfloat[\centering Contraceptive  dataset]{{\includegraphics[width=6cm,height=6cm]{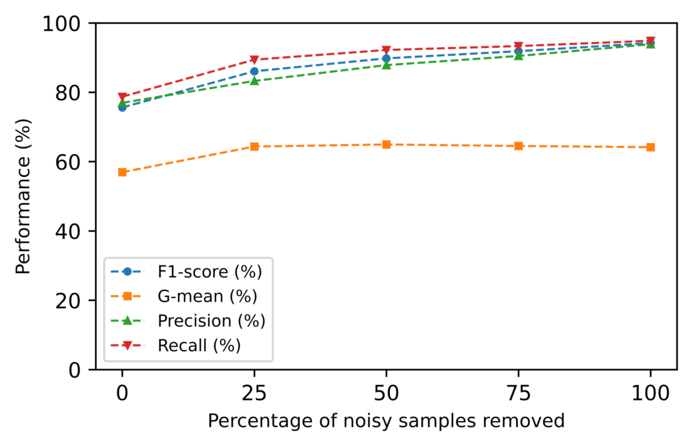} }}
    \subfloat[\centering Vehicle dataset]{{\includegraphics[width=6cm,height=6cm]{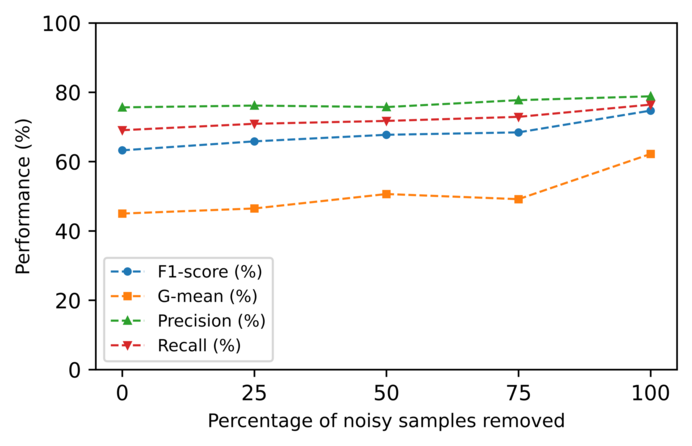} }}
    \subfloat[\centering Vertebral dataset]{{\includegraphics[width=6cm,height=6cm]{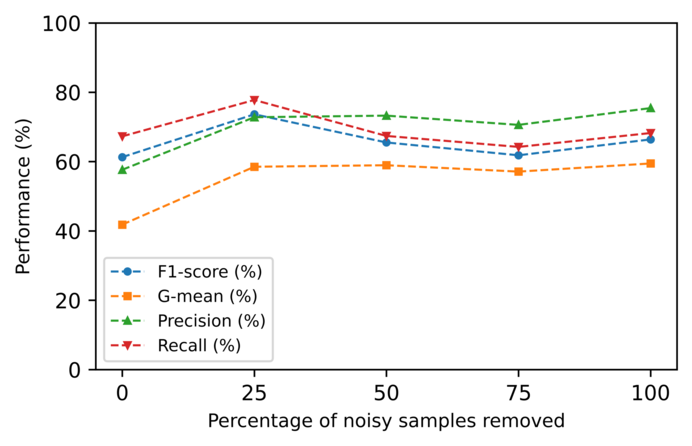} }}

    \caption{Experimental results of noise reduction for different noise thresholds (3 $\le$  IR $\le$ 9).}
    \label{fig:NoiseReduction39}
\end{figure*}

\begin{figure*}[htbp]
    \centering
    
    \subfloat[\centering Contraceptive  dataset]{{\includegraphics[width=6cm,height=6cm]{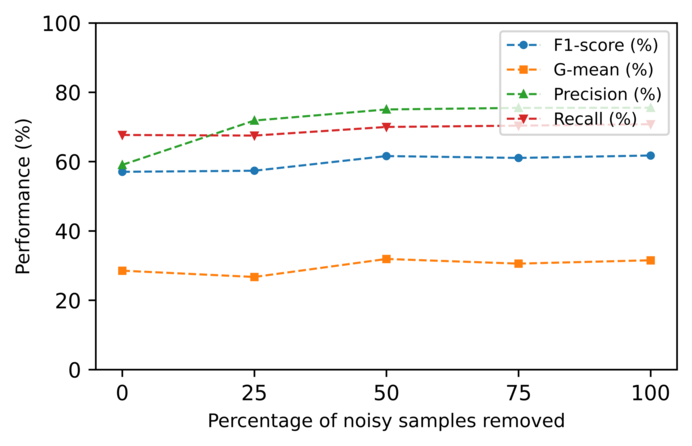} }}
    \subfloat[\centering Vehicle dataset]{{\includegraphics[width=6cm,height=6cm]{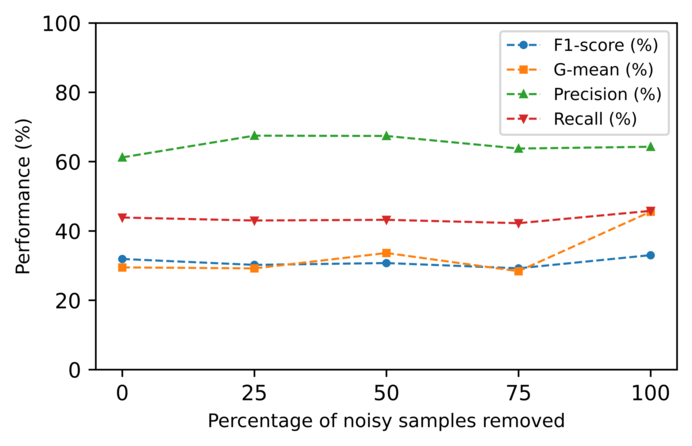} }}
    \subfloat[\centering Vertebral dataset]{{\includegraphics[width=6cm,height=6cm]{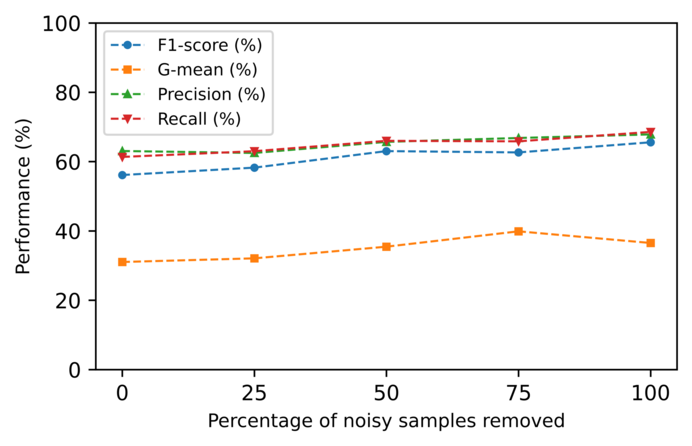} }}

    \caption{Experimental results of noise reduction for different noise thresholds (IR > 9).}
    \label{fig:NoiseReduction9}
\end{figure*}

To evaluate the impact of noisy data on performance and the ability of the proposed framework to address this issue, we tested it under different noise thresholds (0\%, 25\%, 50\%, 75\%, and 100\%).  The threshold variation was applied only to the noisy region identified by the Class-Conditional Region Partition (Algorithm~\ref{alg:SMP} and Algorithm~\ref{alg:CCRP}) to isolate the impact of noise from that of other regions, such as overlapping areas. This experiment aims to assess the degree of noise reduction and its effect on the model’s performance. The experiments were carried out on various datasets with different imbalance ratios, as shown in Figs.~\ref{fig:NoiseReduction3}--\ref{fig:NoiseReduction9}. The abscissa represents the performance scores corresponding to key metrics such as F-score, precision, and recall, while the ordinate indicates the percentage of noisy data removed at each threshold. A value of 0\% represents no noisy data removed, whereas 100\% indicates that all noisy samples have been removed (reaching the final threshold). For most datasets, the proposed framework, IMOVNO+, consistently improves its performance across different thresholds for all key metrics. For instance, on the contraceptive dataset, when no noisy samples are removed, all key metric scores range between 10\% and 50\%, while the performance progressively increases as more noisy samples are removed, reaching approximately 70\% for all relevant metrics when all noisy samples are removed. Similar results were achieved on other datasets with different imbalance ratios.

\subsubsection{Analysis of the strong classifiers ensemble}

In this subsection, we further evaluate the effectiveness of the proposed framework when used with strong classifiers. A strong  classifier is defined as a model that can achieve performance close to the optimal. In this experiment, several strong classifiers were used to replace the earlier weak classifiers to demonstrate the robustness of the proposed framework. These classifiers include logistic regression, random forest, MLP classifier, and support vector machine.

\begin{figure*}[htbp]
    \centering
    
    \subfloat[\centering G-mean]{{\includegraphics[width=9cm,height=6cm]{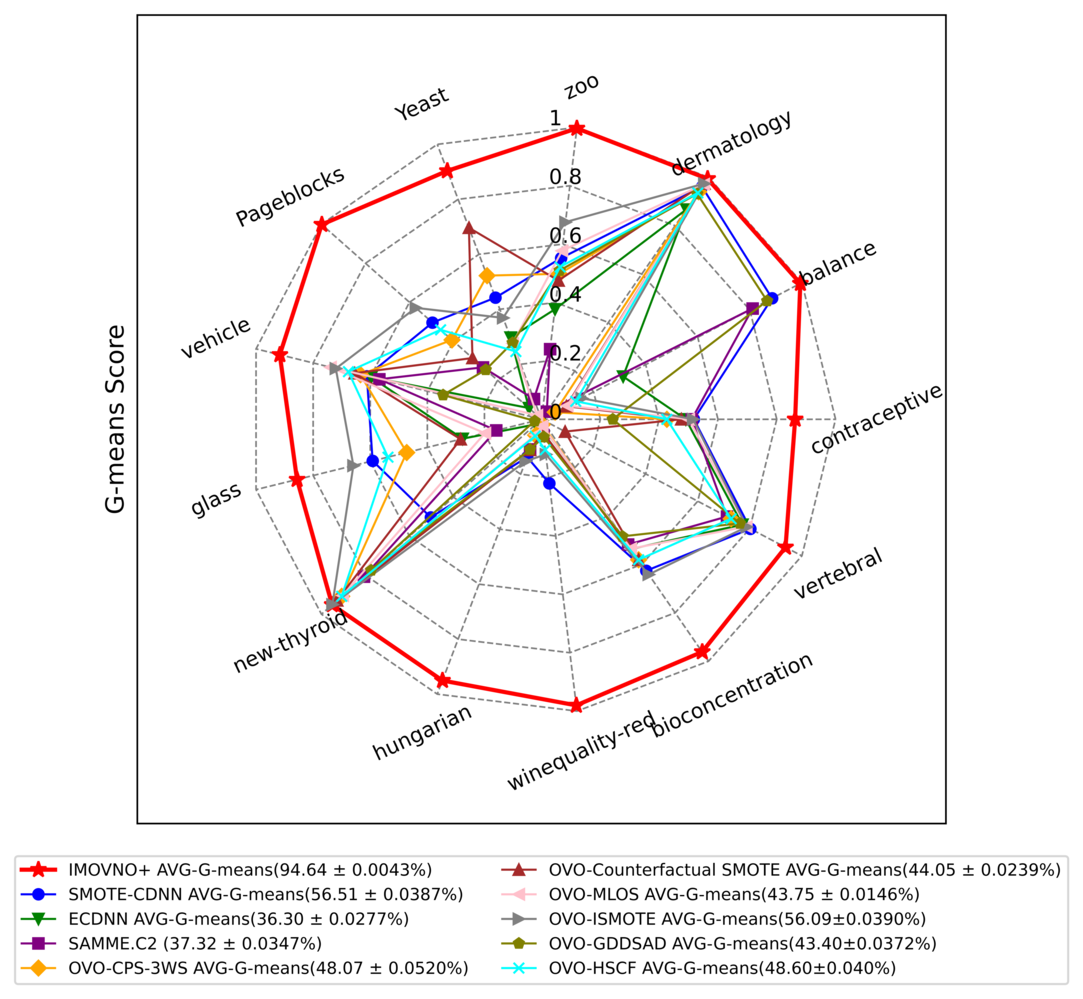} }}
    \subfloat[\centering F-score]{{\includegraphics[width=9cm,height=6cm]{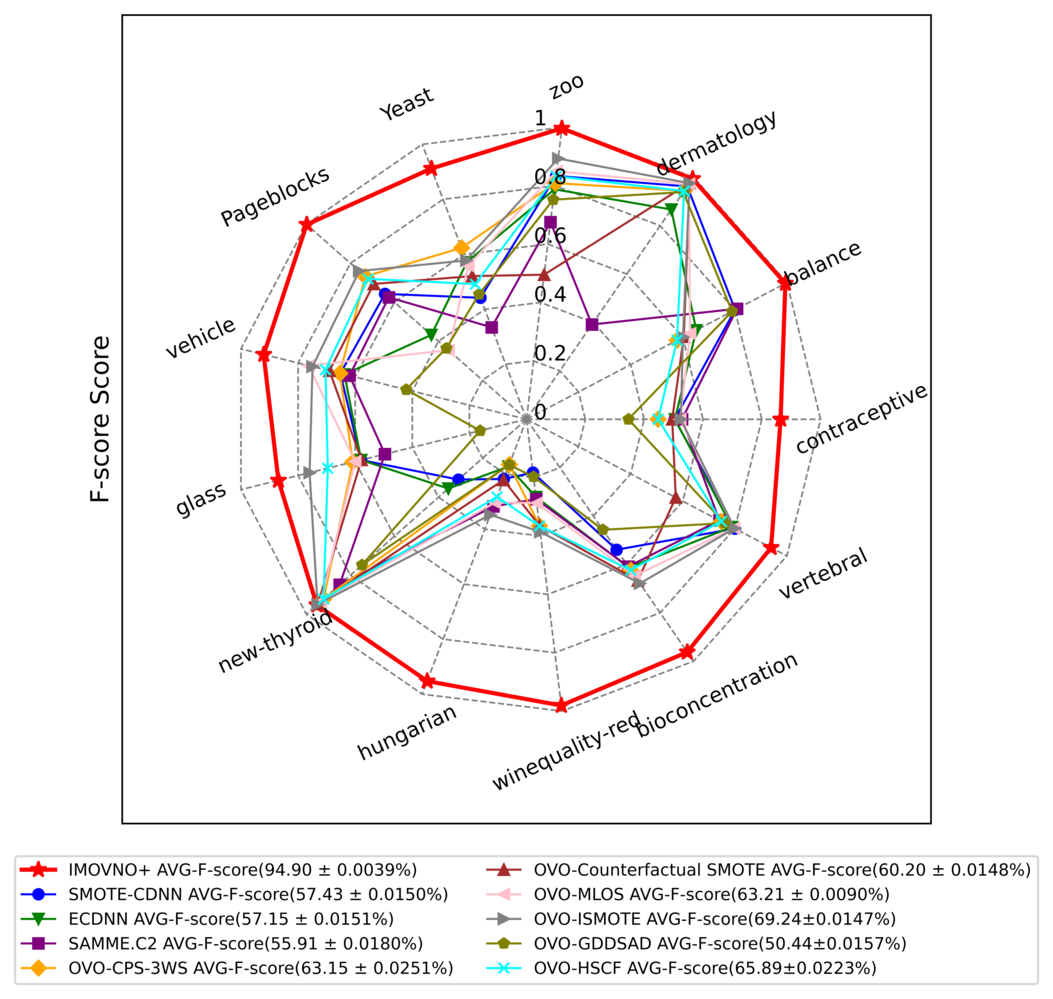} }}\\

    \subfloat[\centering  Recall]{{\includegraphics[width=9cm,height=6cm]{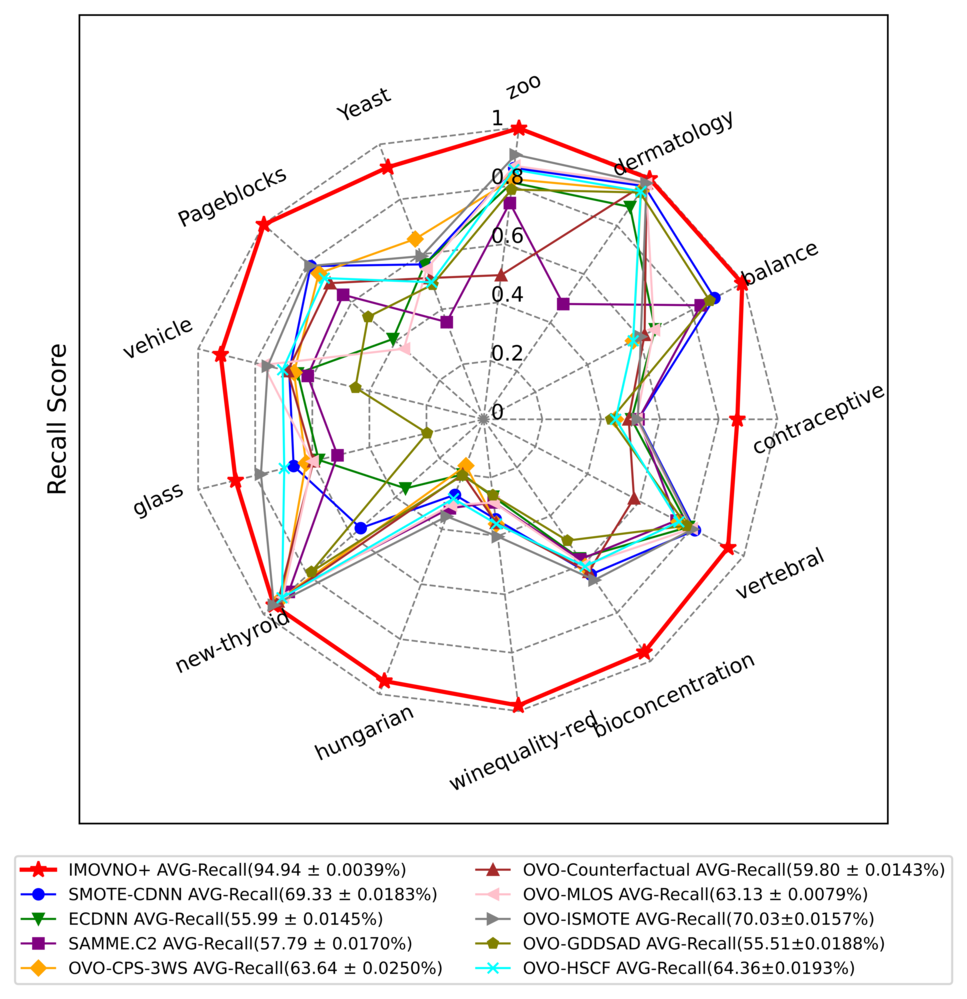} }} 
     \subfloat[\centering Precision]{{\includegraphics[width=9cm,height=6cm]{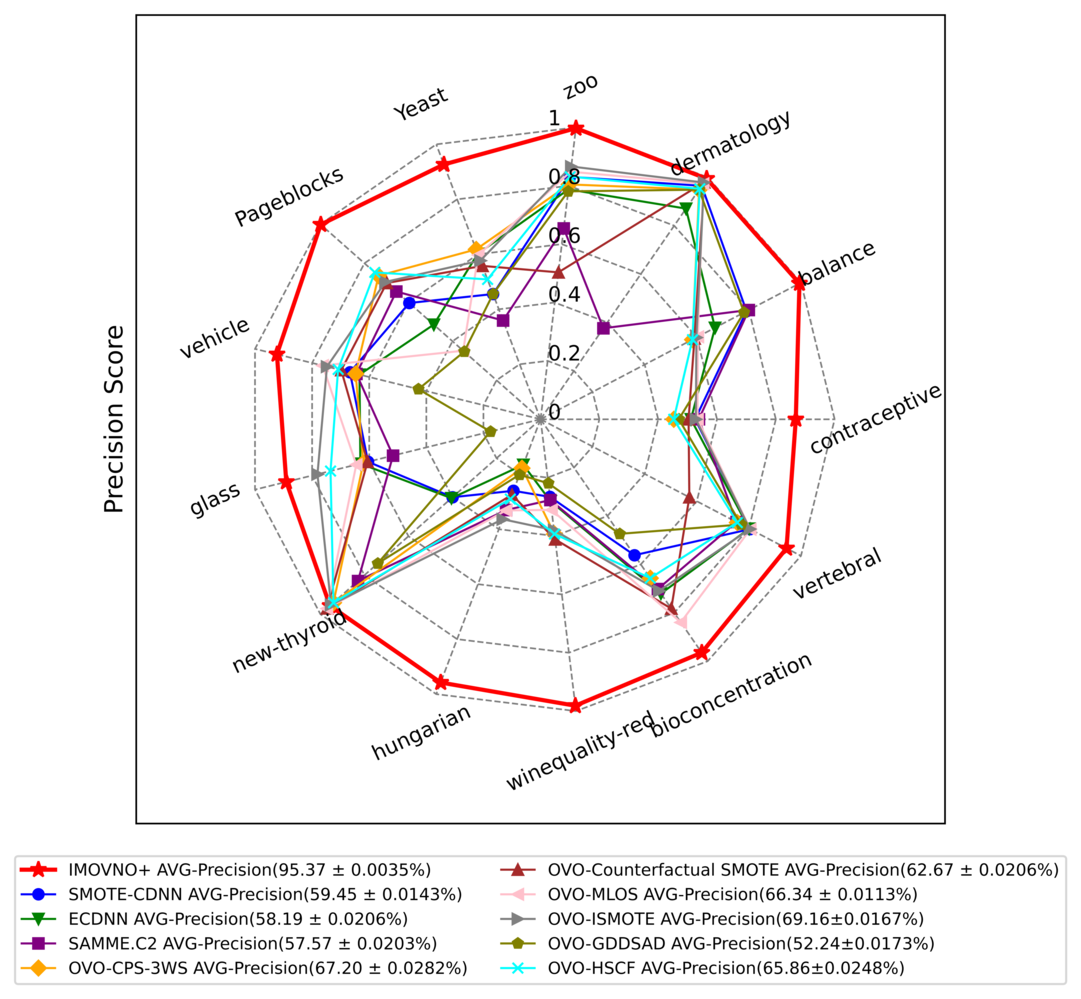} }}

    \caption{Comparative analysis using the strong classifiers.}
    \label{fig:AnalaysStrongClassifiers}
\end{figure*}

As shown in Fig.~\ref{fig:AnalaysStrongClassifiers}, the comparative results across four key metrics are presented using a radar chart for ten methods evaluated on thirteen imbalanced datasets. The comparison includes both ad hoc multiclass imbalanced learning methods and OVO decomposition-based approaches. Overall, the proposed framework, shown as the red line, consistently surpasses all competing methods across all relevant metrics. Remarkably, the proposed framework, IMOVNO+, achieved optimal performance on the Dermatology, Zoo, Balance, and Pageblocks datasets for all metrics. It is also worth noting that OVO-ISMOTE achieved similar performance on only one dataset, New-Thyroid. Additionally, the average results across all thirteen datasets show that the IMOVNO+ outperforms all other methods.

\subsubsection{Statistical significance}

\begin{figure*}[htbp]
    \centering
    
    \subfloat[\centering F-score ]{{\includegraphics[width=9cm,height=5cm]{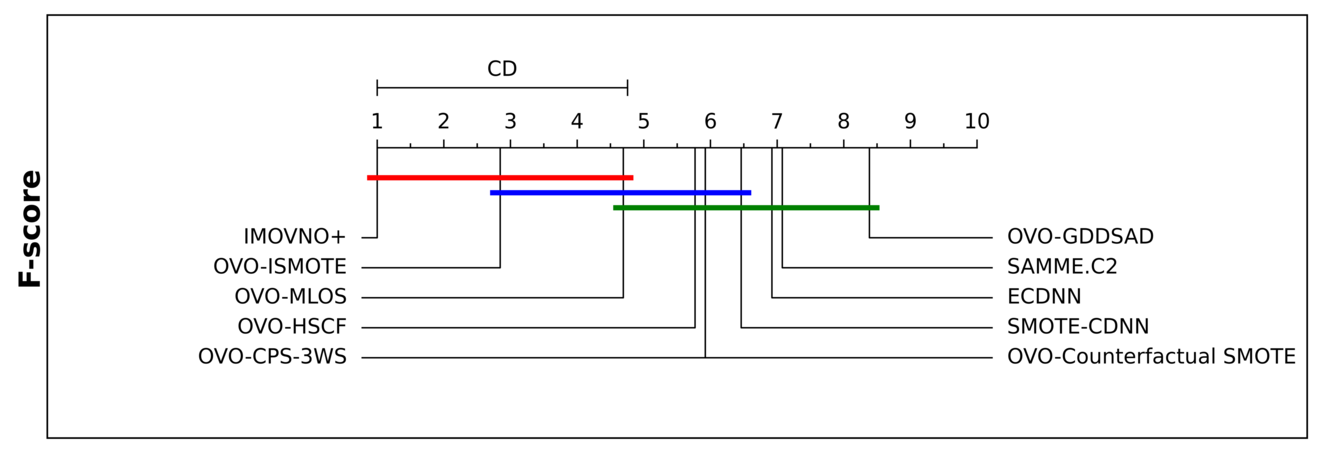} }}
    \subfloat[\centering G-mean]{{\includegraphics[width=9cm,height=5cm]{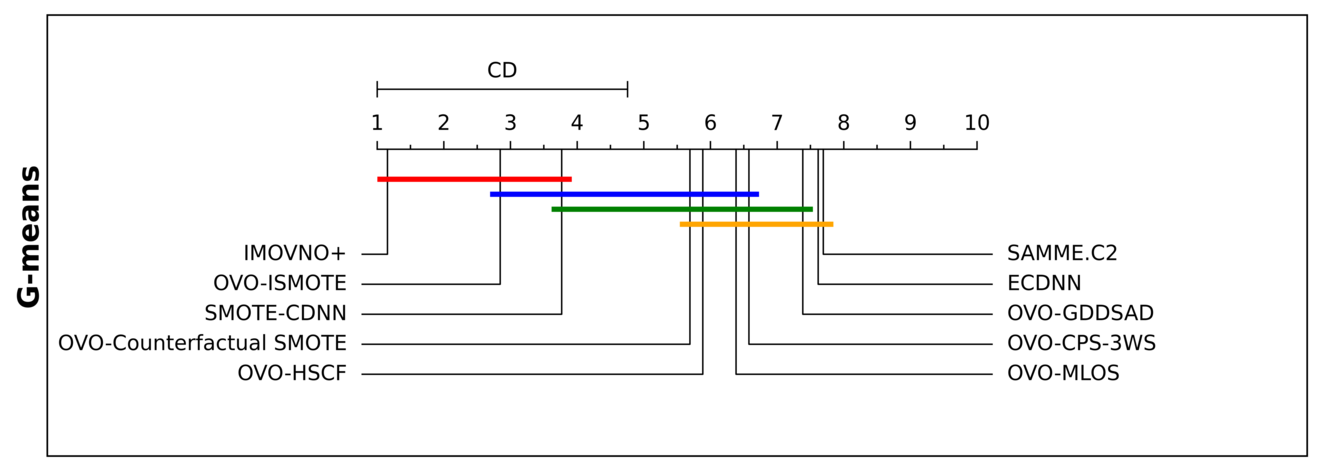} }}\\

    \subfloat[\centering Precision]{{\includegraphics[width=9cm,height=5cm]{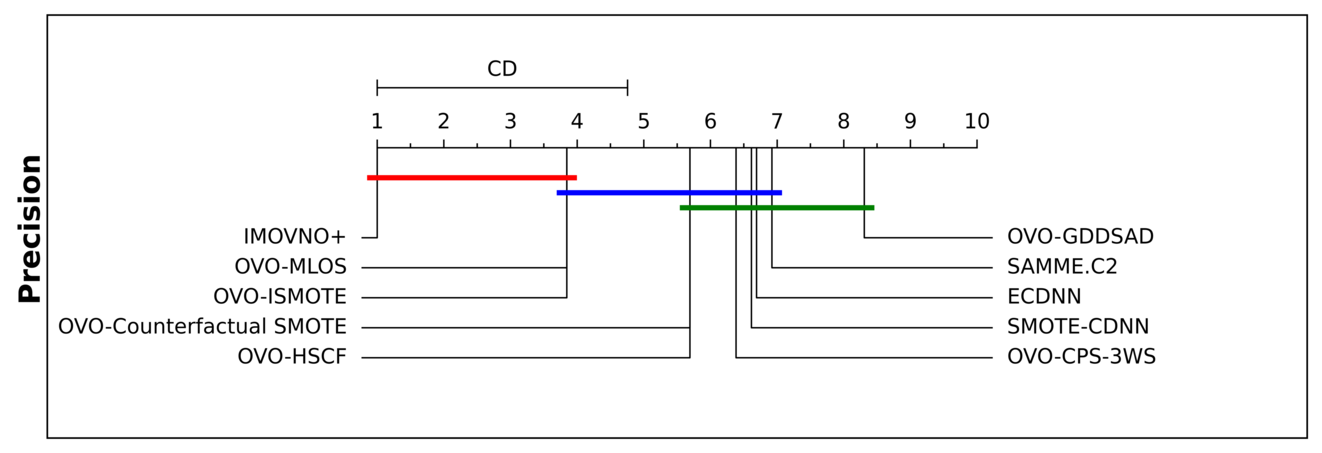} }} 
     \subfloat[\centering Recall]{{\includegraphics[width=9cm,height=5cm]{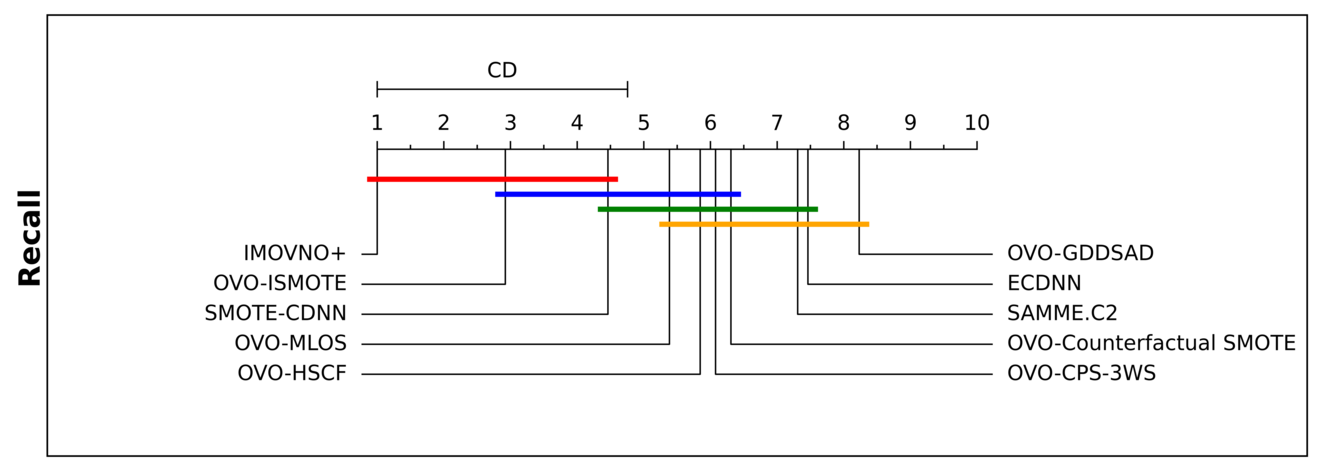} }}

    \caption{Critical difference diagram from the Nemenyi test comparing 10 rebalancing approaches across 13 datasets.}
    \label{fig:CriticalDifferencesGraph}
\end{figure*}

To systematically explore the statistical significance of all compared methods, a nonparametric critical difference (CD) diagram based on the post-hoc Nemenyi test was used to visualize the significance of differences between the algorithms. The results are presented in Fig.~\ref{fig:CriticalDifferencesGraph} with respect to G-mean, F-score, recall, and precision metrics at a significance level of 0.05 across thirteen imbalanced datasets. The coordinate axis illustrates the average ranking of each model, where the left-hand side represents the lowest (best) average rank, and the horizontal lines indicate groups of models with no statistically significant difference in classification performance.

It can be seen from Fig.~\ref{fig:CriticalDifferencesGraph} that the proposed framework, IMOVNO+, obtains the best mean rank, along with OVO-ISMOTE, with respect to all key metrics. For instance, for the G-mean metric in Fig.~\ref{fig:CriticalDifferencesGraph}(a), the proposed framework, IMOVNO+, has the best mean ranking, followed by OVO-ISMOTE as the second best. Moreover, IMOVNO+ performs statistically significantly better than all other methods, except for SMOTE-CDNN, which shows similar performance; both are associated with the same horizontal line in the CD diagram. Similar trends can be observed for the F-score, recall, and precision metrics.

\subsection{General dicussion}

Conventional learning models typically assume balanced data, but real-world applications frequently face challenges related to class imbalance. Although these challenges are significant, issues such as class overlap, noisy data, and associated anomalies can have an even greater impact on model performance and generalisation. Multiclass problems exacerbate these difficulties. Unlike binary classification, multiclass problems are more complex due to the association among classes. Existing one-versus-one binarization solutions fail to fully address these challenges because splitting the problem into subgroups can result in the loss of important characteristics of the original data. Meanwhile, ad hoc solutions have been proposed as alternatives; however, this research area remains largely unexplored, with only a few methods developed so far.
Furthermore, the farthest distances are often considered noisy data and discarded, but this strategy can lead to the loss of informative data due to conventional evaluation functions, such as probability-based measures. At the algorithmic level, ensemble learning faces additional challenges in effectively integrating the contributions of weak classifiers, which often results in limited robustness and suboptimal predictive performance.
To address these challenges, this study proposes a new framework, IMOVNO+, for both binary and multiclass scenarios. IMOVNO+  divides the dataset into three regions: core, overlapping, and noisy, using conditional probability evaluation. Noisy samples are removed based on their low probability, as they are uninformative, while overlapping regions are processed to remove very close samples that may cause class overlap. Next, a balancing process is applied to the minority class using a modified SMOTE with multi-regularisation penalties. Finally, to address the low contribution of ensemble learners, a pruning procedure is applied using a modified direct Jaya algorithm to select the most contributive learners.

Experimental findings illustrate that the IMOVNO+ improves both reliability and performance. In terms of accuracy, it achieved the highest value of 100.00 ± 0.0000\%  on the Zoo dataset, demonstrating its effectiveness and making it a strong candidate for applications in sensitive domains such as healthcare. However, accuracy alone is not sufficient to evaluate model robustness. Therefore, additional metrics such as G-Mean, F-score, precision, and recall were employed for broader validation.  The results show that the IMOVNO+ outperforms multiclass methods, OVO decomposition, and binary classification approaches across all evaluated metrics as reported in the tables. It achieved optimal values for all four key metrics, reaching 100.00 ± 0.0000\% on the Zoo dataset. Unlike other methods, the findings reveal that the IMOVNO+ successfully balances performance by achieving both high recall and high precision, supporting the study’s objective of correctly detecting the minority class or rare samples. This balanced performance also indicates the model’s ability to reduce overfitting. Additionally, the IMOVNO+ achieved the lowest standard deviation. A lower standard deviation indicates higher model reliability, meaning the results are stable across multiple runs with more consistent and confident predictions. The findings of the AUC values in Figs.~\ref{fig:AccuracyAucRader}(b), presented using a radar chart, show that the IMOVNO+ achieves optimal AUC values compared to the other methods. These results indicate the model’s ability to distinguish between classes while reducing overfitting, due to the proposed multi-regularization penalty and overlap-cleaning mechanism.
The heatmap results quantify the degree of overlap between classes before and after applying the proposed method. The results indicate a lower degree of class overlap after processing. This finding demonstrates that the proposed method effectively reduces intersecting samples within inter-class regions through the cleaning process (Algorithm~\ref{alg:SOR}) and through the proposed multi-regularization penalty, which preserves the clarity of decision boundaries after the oversampling process (Algorithm~\ref{alg:OMRP}). Clearer decision boundaries indicate an improved ability of the classifier to learn all classes more effectively. On the other hand, the reduction of noisy data shown in Figure 8 indicates that several data instances contribute very little to the final performance. Cleaning these confusing samples increases the strength of the learner and improves its ability to accurately detect high-certainty samples and performance.

In summary, the proposed framework, IMOVNO+,  has practical significance, particularly in sensitive domains such as healthcare, where class imbalance is a major issue. It improves prediction performance and reduces the risk of mortality. By addressing data issues, the IMOVNO+ enhances classification for new patient cases, minimizes false negatives, and has the potential to save lives. Low standard deviation increases the model’s confidence and stability, which in turn improves prediction reliability in real-world applications. Theoretically, this research proposes new ideas to handle class imbalance in multiclass contexts, an area that remains largely unexplored, opening the door for future research to focus on this issue.

\section{Conclusion}

This paper presents a comprehensive and cohesive framework, IMOVNO+, aimed at enhancing data quality and model performance by concurrently addressing intrinsic data-level issues, such as class imbalance, class overlap, and noisy data, alongside algorithm-level challenges within classification models in both multiclass and binary contexts.

At the data level, the proposed framework adopts a probabilistic approach grounded in conditional probability to evaluate the informational contribution of each sample to the learning process. This framework assesses sample relevance without altering the underlying classifier, thereby maintaining the original data distribution while enhancing data quality. Specifically, data samples are categorized into three semantically significant regions based on their contribution to learning: core, overlapping, and noisy regions. Samples within the core region demonstrate high posterior certainty and strong class representativeness, rendering them highly informative for model training. Conversely, samples in the overlapping region reside in ambiguous zones shared by multiple classes, resulting in complex decision boundaries and increased inter-class uncertainty. Noisy samples are characterized by low conditional probability and minimal learning utility; thus, they are identified and removed to further elevate data quality.

At the algorithm level, our proposed framework diverges from conventional methods that typically operate either solely at the algorithm level (e.g., cost-sensitive learning or ensemble methods) or depend on distance-based heuristics. By integrating both data and algorithm-level strategies, the IMOVNO+ offers a robust alternative to traditional decomposition and binarization techniques, effectively preserving the intricate inter-class relationships inherent in multiclass problems.

A thorough experimental evaluation conducted on several real-world imbalanced datasets illustrates that the IMOVNO+ consistently outperforms state-of-the-art benchmark methods, resulting in enhanced classification performance, greater reliability, and a notable reduction in class overlap. Experimental findings on 35 datasets using G-mean, recall, F-score, and precision demonstrate that IMOVNO+ consistently outperforms state-of-the-art methods for class imbalance, achieving improvements approaching 100\% in several cases. For multi-class datasets, IMOVNO+ achieves gains of 37.06–57.27\% in G-mean, 25.13–43.93\% in F1-score, 24.58–39.10\% in precision, and 25.62–42.54\% in recall. In binary classification, the framework consistently attains near-perfect performance across all evaluation metrics, with improvements ranging from 13.77\% to 39.22\%.

Despite its promising results, the current framework is constrained to numerical and textual data. Future research directions encompass extending the proposed framework to address additional data quality concerns, such as small disjuncts, and adapting it to image data through advanced data augmentation strategies and deep learning techniques. Furthermore, exploring the interaction between our framework and dimensionality reduction methods, such as Principal Component Analysis (PCA), presents another promising avenue for enhancing performance in high-dimensional contexts.

\bibliographystyle{ieeetr}
\bibliography{references}

\end{document}